\documentclass[conference]{IEEEtran}
\usepackage{times}
  
\usepackage[numbers]{natbib}
\usepackage{multicol}
\usepackage[bookmarks=true]{hyperref}
\usepackage{xcolor}
\usepackage{algorithm}
\usepackage{algpseudocode}
\usepackage{algorithmicx}
\usepackage{multirow}
\usepackage{amsmath, amssymb, amsthm} % For math
\usepackage{graphicx}                % For including figures
\usepackage{float}                   % For forcing figure placement
\usepackage{booktabs}                % For professional tables
\usepackage{caption}                 % For sub-captions
\usepackage{subcaption}
\let\labelindent\relax
\usepackage{enumitem}
\usepackage{longtable}
\usepackage{wrapfig}

\usepackage{eso-pic}
\usepackage[dvipsnames]{xcolor}

\setlist*[itemize]{leftmargin=1em, itemindent=0pt}
\setlist*[enumerate]{leftmargin=1em, itemindent=0pt}

\DeclareMathOperator*{\argmax}{arg\,max}

\begin{document}

\AddToShipoutPictureBG*{
  \AtPageUpperLeft{%
    \put(\LenToUnit{0.5\paperwidth},\LenToUnit{-1.2cm}){%
      \makebox[0pt][c]{\textcolor{BrickRed}{\textbf{To appear in the Proceedings of Robotics: Science and Systems (RSS) 2026.}}}
    }%
  }%
}

\title{Adaptive Smooth Tchebycheff Attention\\for Multi-Objective Policy Optimization}

\author{
\authorblockN{Alejandro Murillo-González, Mahmoud Ali and Lantao Liu}
\authorblockA{Indiana University--Bloomington\\
\texttt{\{almuri, alimaa, lantao\}@iu.edu}}
\vspace{-30pt}
}
   
\maketitle

\begin{abstract}
Multi-objective reinforcement learning in robotic domains requires balancing complex, non-convex trade-offs between conflicting objectives. While linear scalarization methods provide stability, they are theoretically incapable of recovering solutions within non-convex regions of the Pareto front. Conversely, static non-linear scalarizations (e.g., Tchebycheff) can theoretically access these regions but often suffer from severe gradient variance and optimization instability in deep RL. In this work, we propose an \textbf{Adaptive Smooth Tchebycheff} framework that resolves this tension by dynamically modulating the curvature of the optimization landscape. We introduce a novel \textit{conflict-driven controller} that regulates the optimization smoothness based on real-time gradient interference. This allows the agent to anneal toward precise, non-convex scalarization when objectives align, while elastically reverting to stable, smooth approximations when destructive gradient conflicts emerge. We validate our approach on a challenging robotic stealth visual search task---a proxy for monitoring of protected/fragile ecosystems---where an agent must balance search, exposure/interference minimization and exploration speed. Extensive ablations confirm that our conflict-aware adaptation enables the robust discovery of Pareto-optimal policies in non-convex regions inaccessible to linear baselines and unstable for static non-linear methods.\\ \\ \textit{Website:}~\url{https://alejandromllo.github.io/research/pasta/}.\\
\end{abstract}


\IEEEpeerreviewmaketitle

\vspace{-6pt}
\section{Introduction} \label{sec:intro}

Deep reinforcement learning (RL) has demonstrated remarkable efficacy in difficult continuous robot control tasks, including quadrupedal locomotion, unstructured terrain navigation, and dexterous manipulation~\cite{kober2013reinforcement, hwangbo2019learning, murillo2025situationally, openai2019rubiks}. Standard approaches typically rely on a scalar reward signal to guide policy optimization. In this paradigm, the environment provides observations and a single scalar reward at each timestep, aggregating various performance metrics—such as energy efficiency, safety, and task completion—into a monolithic objective. This scalar reward often obscures the underlying trade-offs inherent in complex robotic settings and how the preference for each component varies.

Multi-Objective Reinforcement Learning (MORL) addresses scenarios where the environment returns a vector of conflicting rewards~\cite{roijers2013survey, hayes2021practical}. This formulation is critical in robotics, where agents must balance disparate goals—for instance, maximizing speed while minimizing mechanical wear and battery consumption, or navigating to a target while maintaining safety boundaries. The goal of MORL is to find solutions in the Pareto front, the set of policies where no objective can be improved without degrading another.

A common strategy to solve MORL problems is Linear Scalarization (LS), which reduces the reward vector to a scalar via a convex combination of weights. While computationally efficient, LS is theoretically limited: it can only recover solutions on the convex hull of the Pareto front~\cite{das1997closer, noghin2015linear}. In domains with non-convex trade-off surfaces—common in complex dynamics or sparse reward settings—LS fails to discover optimal policies in the concave regions of the front, potentially leading to suboptimal or unsafe behaviors that fail to satisfy mission requirements and constraints.

To overcome the geometric limitations of LS, Tchebycheff (TCH) scalarization is frequently employed in the multi-objective optimization (MOO) literature~\cite{bowman1976relationship, gunantara2018review}. TCH minimizes the maximum weighted deviation from an ideal ``utopia'' point, theoretically allowing it to recover all Pareto-optimal solutions regardless of convexity. However, the TCH formulation relies on a {\em non-differentiable} $\max$ operator. In the context of deep RL, where policy updates rely on backpropagation of gradients, this non-smoothness introduces significant instability \cite{lin2024smooth}. It results in sparse or undefined gradients (subgradients) that hinder the training dynamics \cite{fliege2000steepest, sener2018multi, lin2024few}. 

Recently, \citet{lin2024smooth} introduced Smooth Tchebycheff (STCH) scalarization for gradient-based MOO, replacing the non-differentiable $\max$ operator with a Log-Sum-Exp approximation. This formulation is controlled by a strictly positive smoothness parameter, $\mu$. 
%As $\mu \to 0$, 
As $\mu$ goes to zero, STCH approaches the true Tchebycheff function; 
however, this creates a bottleneck for gradient flow, directing 
feedback almost exclusively from the dominant objective (i.e., the 
one furthest from the ``utopia'' point). Conversely, as $\mu$ moves away from zero, the function behaves increasingly like a linear weighted sum, ensuring smooth gradient flow. This introduces a fundamental trade-off: while the smooth gradient flow mitigates the vanishing gradient problem common in deep learning \cite{pezeshki2021gradient, he2016deep}, it comes at the cost of TCH approximation accuracy. Excessive smoothing degrades the TCH precision, effectively reverting the method to a linear scalarization that is incapable of recovering solutions in non-convex regions of the Pareto front.
 
Furthermore, applying STCH to RL introduces a unique challenge: parameter sensitivity. A fixed $\mu$ is difficult to tune because the ``dataset'' in RL (the agent's experience) is non-stationary. As the policy evolves, the scale of advantages and the proximity to the varying utopia point shift, effectively changing the optimization landscape. A $\mu$ that is too small may cause vanishing gradients, while a large $\mu$ fails to exploit the Tchebycheff properties.

In this work, we propose \textbf{PASTA} ({\em \underline{P}olicy-optimization via \underline{A}daptive \underline{S}mooth \underline{T}chebycheff \underline{A}ttention}), a novel multi-objective reinforcement learning algorithm specifically designed to address the fundamental tension between optimization stability and geometric precision in model-free RL. At the core of PASTA is our {\em Adaptive Smooth Tchebycheff} (\textbf{ASTCH}) scalarization framework, which dynamically modulates the curvature of the optimization landscape via $\mu$. This integration enables agents to stably recover Pareto-optimal solutions within non-convex concavities---regions theoretically unreachable by linear scalarization---while avoiding the catastrophic gradient variance and feature starvation that plague static non-linear approximations. Within ASTCH, we introduce a novel {\em conflict-driven controller} that adjusts the smoothing parameter $\mu$ based on real-time gradient interference. Our method directly monitors the \textit{conflict ratio} between objective gradients; the controller anneals $\mu$ to sharpen the scalarization for precision as training progresses, while employing a {\em dynamic braking mechanism} to momentarily relax smoothness whenever destructive gradient conflicts threaten the policy update. This ``elastic'' adaptation prevents gradient starvation and catastrophic forgetting, promoting robust convergence across the entire Pareto front.

\noindent Overall, our contributions are summarized as follows:
\begin{itemize}
    \item We propose the ASTCH scalarization framework. At its core is a novel Adaptive Smoothness Controller that dynamically adjusts the STCH smoothness parameter $\mu$ based on an online estimate of inter-objective gradient conflict, effectively eliminating the need for manual tuning and preventing gradient starvation.
    \item We develop PASTA, an algorithm that integrates the ASTCH framework into an on-policy policy-gradient method. Our formulation explicitly accounts for the non-stationarity of utopia points and value estimates arising in multi-objective RL, and introduces design choices that further improve performance.
    \item We validate our approach on challenging real-world multi-objective exploration tasks where a ground robot must navigate and visually search an environment while minimizing exposure---a proxy for applications such as stealth monitoring in protected ecosystems or search-and-rescue in contaminated zones. We further demonstrate cross-platform real-world applicability by deploying PASTA on quadrotors tasked with safe traversal among stochastic dynamic obstacles. Finally, we establish broader baseline comparisons using multi-objective MuJoCo benchmarks \cite{felten_toolkit_2023}.
\end{itemize}

\section{Related Work}
\label{sec:related_work}

{Multi-objective planning and control in robotics} involves conflicting objectives, such as energy efficiency versus speed, searching while minimizing interference, or tracking precision versus safety constraints, among many others. We have seen this in diverse problems across RL \cite{van2024revisiting, yang2024robot, yu2024adaptive, afm-RSS-25, 
ibrahim2024comprehensive, causal-structure-dist-learning, jeon2023benchmarking} and control settings \cite{jain2021optimal, jardali2025zero, howard2007optimal, carron2019data, wilde2024scalarizing, tran2020optimized}. In this work, we are interested in RL problems with multiple objectives, for which we want to find an optimal policy for a specific trade-off, instead of learning the whole Pareto front.

{MORL} extends the standard Markov Decision Process (MDP) framework to vector-valued rewards, necessitating a trade-off between conflicting objectives \cite{felten2024multi}. Coverage-based methods use evolutionary algorithms or envelope Q-learning to learn the entire Pareto front \cite{yang2019generalized, van2014multi, reymond2022pareto, chen2020combining}. On the other hand, preference-based approaches like ours have relied on scalarization, surrogate utility functions, uncertainty modeling or intricate optimizations methods \cite{mu2025preference, yang2025preference, wirth2017survey, xu2022preference}.

Closer to our work leveraging TCH/STCH scalarization, \citet{van2013scalarized} proposed a TCH action selection strategy that was shown to outperform linear-based approaches. They also designed a multi-objective Q-learning framework that can leverage both linear and non-linear scalarization methods. Conversely, we directly learn a policy by optimizing an objective function that involves a value learned using STCH-prioritized components.

Recently, \citet{qiu2024traversing} reformulated TCH into a min-max-max optimization problem. This structure enables a preference-free framework based on an Upper Confidence Bound (UCB) that learns Pareto optimal policies with provable sample efficiency. They further extend their method to use STCH. For multi-objective multi-agent RL (MO-MARL), \citet{hu2024tvdo} proposed a factorized value decomposition method inspired by TCH to minimize the maximum deviation in action-value between the team's global action and the local optimal action of each individual agent. To address the limitations of static local scalarization in MO-MARL, \citet{van2014novel} propose a hybrid adaptive weight algorithm combining the layered exploration of RA-TPLS \cite{dubois2011improving} in early iterations with the adaptive norm-based search intensification of AN-TPLS \cite{dubois2011improving}. This approach removes the reliance on rigid dichotomic schemes and biased seeds, resulting in improved system-wide Pareto coverage. In contrast to the aforementioned approaches, our work focuses on tightly integrating STCH into well-known on-policy RL algorithms to learn preference-optimal policies.

Also on the adaptive side, \citet{abels2019dynamic} focus on the setting where scalarization parameters vary over time, and propose a multi-objective Q-network architecture that takes the relative importance of objectives as an input to output weight-dependent Q-values. To mitigate the non-stationarity inherent in changing preferences and to improve sample efficiency in high-dimensional environments, they introduce Diverse Experience Replay (DER), which ensures a diverse range of weight configurations are retained to reduce replay buffer bias. Similar adaptive scalarization parameter schemes have been explored for robotic arm control \cite{shianifar2025adaptive}. Rather than selecting scalarization preferences, we focus on balancing optimization smoothness, STCH approximation precision and solution quality.

\vspace{-2mm}
\section{Preliminaries} \label{sec:preliminaries}

\subsection{Multi-Objective Optimization (MOO)}

Many tasks involve optimizing multiple, often conflicting, objectives simultaneously. This gives rise to a Multi-objective Optimization Problem (MOP), which is formally defined as:
\begin{equation}
    \min_{\mathbf{x} \in \mathcal{X}} \mathbf{F}(\mathbf{x}) = \min_{\mathbf{x} \in \mathcal{X}} [f_1(\mathbf{x}), f_2(\mathbf{x}), \dots, f_m(\mathbf{x})]^\top
    \label{eq:mop_def}
\end{equation}
where $\mathbf{x}$ is the decision vector (e.g., policy parameters $\theta$) from the feasible set $\mathcal{X}$, and $\mathbf{F}: \mathcal{X} \to \mathbb{R}^m$ is the vector-valued objective function mapping $\mathbf{x}$ to an $m$-dimensional objective.

\subsubsection{Pareto Optimality}

MOPs do not typically have a single optimal solution. Instead, we seek a set of ``best" trade-offs, defined by the concept of dominance.

\textbf{Dominance:} A solution $\mathbf{x}_A$ \textbf{dominates} $\mathbf{x}_B$ (denoted $\mathbf{x}_A \prec \mathbf{x}_B$) if its objective vector $\mathbf{u} = \mathbf{F}(\mathbf{x}_A)$ is no worse in any objective and strictly better in at least one:
\begin{equation*}
    \forall i \in \{1,\dots,m\}:  u_i \leq v_i ~~ \land ~~ \exists j \in \{1, \dots, m\}: u_j < v_j
\end{equation*}
where $\mathbf{v} = \mathbf{F}(\mathbf{x}_B)$.

The concept of dominance lets us define additional key components of the MOP. A solution $\mathbf{x}^* \in \mathcal{X}$ is \textbf{Pareto optimal} if no other solution $\mathbf{x} \in \mathcal{X}$ dominates it. The set of all Pareto optimal solutions in the decision space is the \textbf{Pareto Set}, $\mathcal{PS}$. Its image in the objective space, $\mathcal{PF} = \{ \mathbf{F}(\mathbf{x}) \mid \mathbf{x} \in \mathcal{PS} \}$, is the \textbf{Pareto Front}. Finally, the goal of MOO is to find a set of solutions that approximates the $\mathcal{PF}$.
 
\subsubsection{Solution Methods}

We focus on \textit{gradient-based} MOO solution methods, given their compatibility with deep learning based RL. Most gradient-based MOO methods can be categorized as \textit{scalarization} or \textit{adaptive gradient} approaches~\cite{lin2024smooth}. The latter seek to simultaneously improve all objectives every iteration by discovering an appropriate gradient direction~\cite{schaffler2002stochastic, desideri2012mutiple}. Scalarization-based approaches convert the MOP into a parameterized Single-Objective Problem (SOP). 

\textbf{Linear Scalarization:} The simplest scalarization method is the weighted sum: $L(\mathbf{x}, \mathbf{w}) = \sum_{i=1}^m w_i f_i(\mathbf{x})$, for a weight vector $\mathbf{w}$ with $w_i \ge 0, \sum w_i = 1$. While simple, this method can \emph{only} find solutions on the \textit{convex hull} of the Pareto Front~\cite{das1997closer}. It fails to find solutions in non-convex regions, which are necessary and common in practice \cite{ehrgott2005multicriteria}.

\textbf{Tchebycheff Scalarization:} To address this, Tchebycheff scalarization is widely used \cite{bowman1976relationship}. It works by minimizing the maximum weighted distance to an ideal ``utopia" point $\mathbf{z}^* = [z_1^*, \dots, z_m^*]^\top$.
The utopia point consists of components $z_i^*$ that represent the theoretical minimum (or slightly below) of each objective $f_i$ individually. Crucially, $\mathbf{z}^*$ is strictly infeasible; achieving all ideal values simultaneously is impossible due to the conflicting nature of the objectives. Conceptually, this acts like a ``carrot on a stick'' for the optimizer: the utopia point is the carrot suspended just beyond the feasible region. The optimizer is driven to minimize the gap to this unreachable target, which effectively pulls the solution as far as possible towards the Pareto frontier.
The function to minimize is:
\begin{equation}
    L_T(\mathbf{x}, \mathbf{w}, \mathbf{z}^*) = \max_{i \in \{1,\dots,m\}} \{ w_i (f_i(\mathbf{x}) - z_i^*) \}.
    \label{eq:tchebycheff}
\end{equation}
The Tchebycheff method is guaranteed to find \emph{any} Pareto optimal solution \cite{steuer1983interactive, lin2024smooth}, regardless of the front's shape. 

\textbf{Gradient-Based Tchebycheff Optimization:} The non-linear $\max$ function in Eq. \eqref{eq:tchebycheff} poses a challenge for gradient-based methods. In practice, the subgradient is used \cite{lin2024few}. The gradient of $L_T$ with respect to $\mathbf{x}$ (or policy parameters $\theta$) is the gradient of the \emph{single objective} that is currently the ``worst-performer" (the one that determines the $\max$):
\begin{equation}
    j = \argmax_{i} \{ w_i (f_i(\mathbf{x}) - z_i^*) \} \implies \nabla_{\mathbf{x}} L_T \approx \nabla_{\mathbf{x}} \big( w_j f_j(\mathbf{x}) \big) \nonumber
\end{equation}
This approach introduces its own challenges. The subgradient can ``jump" abruptly as the $\argmax$ switches between objectives, leading to unstable training (see Fig. 2 of \cite{lin2024smooth} for an example). It also requires a well-defined utopia point $\mathbf{z}^*$, which may not be known a priori and would need to be estimated. This process is formalized in Alg. \ref{alg:tchebycheff_step} (Appendix \ref{sec:appendix-tch-pseudocode}). 

\subsection{Multi-Objective Reinforcement Learning (MORL)}
\label{subsec:morl_prelims}

MORL models the problem as a Multi-Objective Markov Decision Process (MOMDP), defined by the tuple $(\mathcal{S}, \mathcal{A}, P, \mathbf{r}, \gamma)$. Here, $\mathcal{S}$ is the state space, $\mathcal{A}$ is the action space, and $P: \mathcal{S} \times \mathcal{A} \times \mathcal{S} \to [0, 1]$ represents the state transition dynamics. Unlike standard MDPs, the reward function $\mathbf{r}: \mathcal{S} \times \mathcal{A} \to \mathbb{R}^m$ returns a vector of $m$ distinct objective signals $\mathbf{r}(s,a) = [r_1(s,a), \dots, r_m(s,a)]^\top$. 

The goal of the agent is to learn a policy $\pi(a|s)$ that optimizes the expected discounted vector return: $\mathbf{J}(\pi) = \mathbb{E}_{\pi} \left[ \sum_{t=0}^{\infty} \gamma^t \mathbf{r}(s_t, a_t) \right]$
where $\gamma \in [0, 1)$ is the discount factor. 

Since objectives are often conflicting, no single set of \textit{policy parameters} maximizes all simultaneously. We therefore seek policies that are \textit{Pareto-optimal} \cite{hayes2021practical}. A policy $\pi$ is said to \textit{dominate} another policy $\pi'$ (denoted $\mathbf{J}(\pi) \succ \mathbf{J}(\pi')$) if its \textbf{performance} is superior or equal in all objectives and strictly superior in at least one:
\begin{equation}
    \forall i \in \{1,\dots,m\}, J_i(\pi) \ge J_i(\pi') ~~ \land ~~ \exists j, J_j(\pi) > J_j(\pi').
\end{equation}
The set of non-dominated policies constitutes the \textit{Pareto Front}.

\subsection{Policy Optimization Algorithms} \label{subsec:prelim-ppo}

This section draws from \cite{SpinningUp2018, sutton1999policy, schulman2015trust, schulman2017proximal} to trace the evolution of modern policy optimization algorithms.

\subsubsection{Policy Gradient Methods (PG) \cite{sutton1999policy}}

directly optimize the parameters $\theta$ of a policy $\pi_\theta$ to maximize the expected return. This is typically achieved via stochastic gradient ascent on the policy's performance $J(\pi_{\theta_k})$:
\begin{equation}
    \theta_{k+1} = \theta_k + \alpha \nabla_{\theta_k} J(\pi_{\theta_k}),
\end{equation}
where $\alpha$ is the learning rate and the policy gradient is:
\begin{equation}
    \nabla_{\theta_k} J(\pi_{\theta_k}) = \underset{s, a \sim \pi_{\theta_k}}{\mathbb{E}} \Bigg{[} \sum_{t=0}^T \nabla_{\theta_k} \log \pi_{\theta_k} (a \mid s) A^{\pi_{\theta_k}} (s, a) \Bigg{]}.
\end{equation}
Here, $A^{\pi_{\theta_k}}(s, a)$ is the \textit{advantage function}, often derived from a simultaneously learned value function $V_{\phi_k}$.

However, PG methods are sensitive to the learning rate $\alpha$. A single step that is too large can drastically change the policy, leading to a ``performance collapse". 

\subsubsection{Trust Region Policy Optimization (TRPO) \cite{schulman2015trust}}
tries to address the instability of PG by taking the largest possible step that improves performance while explicitly constraining how much the new policy $\pi_\theta$ can diverge from the old one $\pi_{\theta_k}$. This ``trust region" is enforced using the Kullback-Leibler (KL) divergence. The theoretical TRPO update is a constrained optimization problem:
\begin{equation}
    \theta_{k+1} = \argmax_{\theta} L (\theta, {\theta_k}) \texttt{ s.t. } \bar{D}_{KL} (\theta \mid \mid \theta_k) \leq \psi \label{eq:trpo_formulation}
\end{equation}
where $\psi$ is the step size (the ``size" of the trust region), $L(\theta, {\theta_k})$ is the \textit{surrogate advantage}:
\begin{equation}
    L(\theta, \theta_k) = \underset{s, a \sim \pi_{\theta_k}}{\mathbb{E}} \Bigg{[} \frac{\pi_{\theta} (a \mid s) }{\pi_{\theta_k} (a \mid s) } A^{\pi_{\theta_k}}(s, a) \Bigg{]},
\end{equation}
and $\bar{D}_{KL} (\theta \mid \mid \theta_k)$ is the average KL-divergence between policies, both computed using states visited and corresponding actions sampled by the old policy. In practice, the authors solve an approximate optimization problem derived from a Taylor expansion of Eq. \eqref{eq:trpo_formulation}.

\subsubsection{Proximal Policy Optimization (PPO) \cite{schulman2017proximal}}
emerged as an alternative that retains the stability of TRPO but with a simpler, first-order optimization approach. PPO's key insight is to reformulate the objective function to penalize large policy changes directly, rather than using a hard KL-divergence constraint. PPO implements this via a \textit{clipped surrogate} objective function. Specifically, the full objective comprises three components. First, the \textit{clipped surrogate} objective penalizes large policy deviations. Defining the probability ratio $r_t(\theta) = \frac{\pi_\theta(a|s)}{\pi_{\theta_k}(a|s)}$, this loss is given by: $L^{CLIP}_t(\theta) = $
\begin{equation} \label{eq:ppo-clip-loss}
    \min \Big( r_t(\theta) A^{\pi_{\theta_k}}(s,a), \texttt{clip} \big( r_t(\theta), 1 - \epsilon, 1 + \epsilon \big) A^{\pi_{\theta_k}}(s,a) \Big),
\end{equation}
where $\epsilon$ is a hyperparameter bounding the update. Second, a value function loss $L^{VF}_t(\phi) = (V_{\phi}(s_t) - V^{\text{target}}_t)^2$ ensures accurate advantage estimation. Third, an entropy bonus $S[\pi_\theta](s_t)$ encourages exploration. The final parameters are updated by maximizing the weighted combination of these terms:
\begin{align} \label{eq:ppo-update}
    \theta_{k+1}&, \phi_{k+1} = \\ &\argmax_{\theta, \phi} \underset{s, a \sim \pi_{\theta_k}}{\mathbb{E}} \Big[ L^{CLIP}_t(\theta) - c_1 L^{VF}_t(\phi) + c_2 S[\pi_\theta](s_t) \Big], \nonumber
\end{align}
where $c_1$ and $c_2$ are coefficients weighting the value error and entropy bonus, respectively. See Appendix \ref{sec:appendix-ppo-details} for additional details and pseudocode.

\section{Policy Optimization via\\Smooth Tchebycheff Attention}
\label{sec:scalarized_ppo}

We address the challenge of learning Pareto-optimal policies in non-convex MORL domains, where standard linear scalarization of the reward fails to recover solutions beyond the convex hull. To resolve these complex trade-offs, we integrate \emph{Smooth Tchebycheff (STCH) scalarization}~\cite{lin2024smooth} into an on-policy policy-gradient framework based on PPO~\cite{schulman2017proximal}. Specifically, integrating STCH seeks to unlock the discovery of non-convex solutions while improving the gradient stability essential for deep RL.

Crucially, the STCH analytical derivative naturally functions as a principled \emph{attention mechanism}. This section establishes the core mathematical integration of this insight. However, relying on a static scalarization curvature risks gradient starvation and optimization collapse as the policy evolves. To overcome this, Section \ref{sec:adaptive_controller} introduces a conflict-driven controller to actively modulate the STCH smoothness. This addition completes the ``adaptive'' component of our framework, culminating in our full proposed algorithm: \textbf{PASTA}.

\subsection{Smooth Tchebycheff Scalarization}

We consider $m$ objectives with user-defined preference weights $\mathbf{w} \in \Delta^{m-1}$, such that $w_i  \ge 0, \sum w_i = 1$. Classical Tchebycheff scalarization minimizes the maximum weighted deviation from an ideal (utopia) point $\mathbf{z}^*$. In RL, however, optimization is naturally posed as a maximization problem.

We propose to do this by fixing the utopia point at $z_i^* = \zeta > 1$. % (with $\zeta = 1.05$).
To ensure scale invariance and a stable reference point, we also dynamically normalize raw \textit{returns} $R_i$ using running estimates of the global minimum $R_{i}^{\min}$ and maximum $R_{i}^{\max}$ observed during training: $\bar{r}_i = (R_i - R_{i}^{\min})/(R_{i}^{\max} - R_{i}^{\min} + \epsilon)$, such that $\bar{r}_i \in [0, 1]$. As a result, we can turn our MOP into a SOP (scalarize) using STCH~\cite{lin2024smooth} in maximization form, with respect to the normalized returns:
\begin{equation}
    \mathcal{S}_{\mathrm{STCH}}(\mathbf{\bar{r}})
    =
    -\mu \log \sum_{i=1}^m
    \exp\!\left(
        \frac{w_i (z_i^* - \bar{r}_i)}{\mu}
    \right),
    \label{eq:stch_max_explicit}
\end{equation}
where $\mu > 0$ is the smoothness parameter.

As $\mu \to 0$, Eq.~\eqref{eq:stch_max_explicit} converges to the classical (hard) Tchebycheff operator. As $\mu \to \infty$, it approaches the linear weighted sum $\sum_i w_i \bar{r}_i$. Importantly, $\mu$ controls not only bias--variance trade-offs but the \emph{curvature of the Pareto front approximation} and, consequently, the structure of the induced gradients. See Appendix~\ref{sec:appendix-stch} for illustrations and further analysis of STCH.

\subsection{Smooth Tchebycheff Attention}

A defining property of STCH is that its gradient induces a normalized, positive weighting over objectives:
\begin{equation}
    \delta_i(\mathbf{\bar{r}})
    \;\triangleq\;
    \frac{\partial \mathcal{S}_{\mathrm{STCH}}(\mathbf{\bar{r}})}{\partial \bar{r}_i}
    =
    \frac{
        \exp\!\left(
            w_i (z_i^* - \bar{r}_i)/\mu
        \right)
    }{
        \sum_j
        \exp\!\left(
            w_j (z_j^* - \bar{r}_j)/\mu
        \right)
    }.
    \label{eq:stch_attention}
\end{equation}

\noindent These coefficients admit a precise interpretation:
\begin{itemize}
    \item Objectives with larger weighted deviation from the utopia point receive exponentially higher attention;
    \item $\mu$ controls how sharply attention concentrates on the limiting objective, i.e., the one with largest weighted deviation from the utopia.
\end{itemize}

Crucially, $\boldsymbol{\delta}$ is \emph{not} a heuristic weighting. It is the exact multiplier with which each objective contributes to the scalarized gradient. This observation motivates our design choice, described in the remaining of this text, to propagate STCH attention consistently across policy and value updates. 

However, in the low-$\mu$ regime, the attention weights in Eq.~\eqref{eq:stch_attention} become sharply peaked. While necessary for resolving non-convex Pareto trade-offs, this concentration induces a well-known failure mode in deep learning: \emph{gradient starvation} \cite{pezeshki2021gradient, he2016deep}. Objectives that are temporarily non-limiting receive near-zero gradient signal, causing the shared representation to forget features critical for those objectives.

To counteract this effect, we introduce a \textbf{Maintenance Rate} $\rho \in (0,1)$, which enforces a minimum gradient budget for all objectives. As such, we define the effective attention as
\begin{equation} \label{eq:stch_att_with_grad_preservation}
    \boldsymbol{\eta}
    =
    (1-\rho)\boldsymbol{\delta}
    +
    \rho \mathbf{u},
    \qquad
    \mathbf{u} = (1/m,\dots,1/m).
\end{equation}

This guarantees that each objective receives at least $\rho/m$ of the gradient signal. Conceptually, this mechanism plays a role analogous to persistent excitation in adaptive control: it ensures that the system remains identifiable along all objective dimensions, enabling rapid re-adaptation when trade-offs shift.

\subsection{Attention-Aligned Branched Critic}

To ensure that value learning prioritizes the same objectives indicated by STCH, we employ a \textbf{Branched Critic} architecture with a shared feature extractor and $m$ independent value heads. 
The new critic loss, replacing $L_t^{VF}$ in Eq. \eqref{eq:ppo-update}, is: 
\begin{equation} \label{eq:astch-value-loss}
    L_t^{VF}
    =
    \frac{1}{N}
    \sum_{t=1}^N
    \sum_{i=1}^m
    \eta_{t,i}
    \left(
        V^{(i)}_\phi(s_t) - y_{t}^{(i)}
    \right)^2,
\end{equation}
where $y_{t}^{(i)}$ denotes the target return for objective $i$. Consistent with PPO \cite{schulman2017proximal}, we calculate $y_{t}^{(i)}$ as the $\lambda$-return, using Generalized Advantage Estimation (GAE) \cite{schulman2015high} to obtain a variance-reduced estimate of the discounted return (Sec. \ref{subsec:morl_prelims}).

This attention-aligned critic ensures that representational capacity is allocated where it most reduces variance in the scalarized advantage. This is also consistent with TCH's focus on bottleneck objectives. Without this alignment, the critic would expend capacity modeling objectives that contribute little to the policy improvement.

\subsection{Vectorized Advantage}
\label{subsec:vector_advantages}

Unlike standard scalarized MORL, which collapses returns into a single utility value prior to optimization, we maintain a vector of advantages over which we do gradient projection later on. For each objective $i \in \{1, \dots, m\}$, we compute the Advantage $A_i(s_t, a_t)$ using GAE. We then apply per-objective normalization over the batch: 
\begin{equation} \label{eq:vec-advantage}
    \hat{A}_i(s, a) = \big( A_i(s, a) - \mathbb{E}[A_i] \big) \big/ \big( \sigma(A_i) + \epsilon \big).
\end{equation}
This yields a vector advantage $\hat{\mathbf{A}}(s,a) \in \mathbb{R}^m$. 

Rather than scalarizing $\hat{\mathbf{A}}$ immediately, which effectively destroys information about gradient \textit{conflict}, we use this vector during the actor's optimization. Specifically, deferring scalarization allows the actor to compute distinct policy gradients $\mathbf{g}_i$ for each objective. This separation is a prerequisite for our optimization strategy, as it exposes the geometric conflicts required to mitigate destructive interference, explained next.

\subsection{Conflict-Aware Gradient Projection (PCGrad)}
\label{subsec:pcgrad}

Despite attention-aligned updates, non-convex Pareto trade-offs can still induce destructive gradient interference. To address this issue, we integrate a subtly modified variant of Projecting Conflicting Gradients (PCGrad)~\cite{pcgrad} that operates at the level of individual objectives, rather than tasks, enabling conflict projection directly in objective space.

Let $\mathbf{g}_i = \nabla_\theta L_i^{CLIP}$ (recall Eq. \eqref{eq:ppo-clip-loss}) denote the policy gradient associated with objective $i$ via the vectorized advantage $\hat{A}_i(s, a)$ from Eq. \eqref{eq:vec-advantage}. For randomly sampled pairs of objectives $(i,j)$, PCGrad detects conflicts via the inner product:
\(
    \mathbf{g}_i^\top \mathbf{g}_j < 0.
\)
If a conflict is detected, $\mathbf{g}_i$ is projected onto $\mathbf{g}_j$'s normal plane:
% \vspace{-4mm}
\begin{equation}
    \mathbf{g}_i^{\mathrm{proj}}
    =
    \mathbf{g}_i
    -
    \frac{
        \mathbf{g}_i^\top \mathbf{g}_j
    }{
        \|\mathbf{g}_j\|^2
    }
    \mathbf{g}_j.
\end{equation}

Finally, replacing the standard scalarized PPO update, we sum the projected gradients of $m$ distinct clipped objectives $L_i^{CLIP}$ to update the policy: $\theta_{k+1} = \theta_k + \alpha_\theta \sum_{i=1}^m \mathbf{g}_i^{\mathrm{proj}}$. This ensures that the optimization respects individual trust-region constraints while mitigating geometric conflicts.

Crucially, we note that PCGrad also provides a principled measure of gradient interference. Based on this we define the \textbf{Conflict Ratio} $\kappa_t$ at iteration $t$ as:
\begin{equation} \label{eq:conflict-ratio}
    \kappa_t
    =
    \frac{
        \#\{(i,j) : \mathbf{g}_i^\top \mathbf{g}_j < 0\}
    }{
        \#\{(i,j)\}
    },
\end{equation}
i.e., the fraction of sampled gradient pairs exhibiting conflict. 

A key idea of this work is to \emph{elevate} $\kappa_t$ from a passive diagnostic to an active control signal. We show that $\kappa_t$ provides a reliable, real-time estimate of the agent’s ability to reconcile the current Pareto trade-offs, and we leverage it as the feedback variable in our adaptive smoothness controller, closing the loop between optimization dynamics and scalarization curvature.

\section{Conflict-Driven Adaptive Smoothness}
\label{sec:adaptive_controller}

\begin{figure}
    \centering

    \captionsetup{labelfont=small, textfont=small}
    \includegraphics[width=\linewidth]{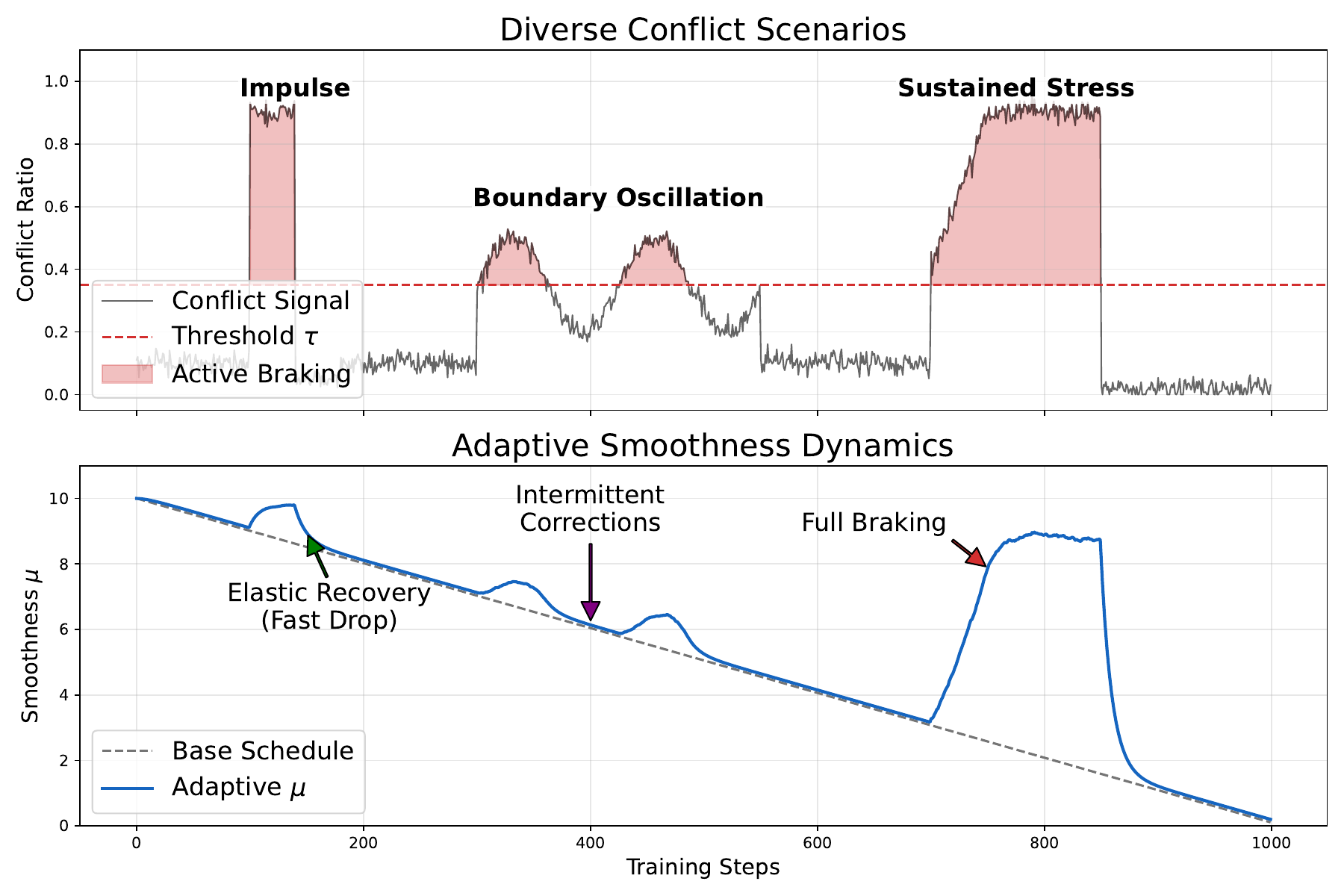} \vspace{-4mm}
    \caption{\small 
        The conflict ratio  $\kappa_t$  (top), computed from inter-objective gradient interference, serves as a feedback signal for adapting the Smooth Tchebycheff parameter $\mu_t$ (bottom). Transient spikes in conflict trigger rapid increases in $\mu_t$ (``elastic recovery") to stabilize optimization, while sustained high conflict induces prolonged braking toward a linear-like scalarization. As conflict subsides, $\mu_t$ decays back toward the base schedule, enabling annealing toward sharper non-convex scalarization.
    }
    \label{fig:adaptive_smoothness_dynamics}
    \vspace{-2mm}
\end{figure}

We introduce a mechanism to adapt the smoothness parameter $\mu$ of STCH during training. The controller leverages objective-gradient interference, quantified by the conflict ratio in Eq.~\eqref{eq:conflict-ratio}, as a real-time feedback signal. This feedback allows the scalarization smoothness to adjust dynamically in response to the evolving optimization landscape.

\subsection{Adaptive Smoothness Mechanism}
The controller is designed as a closed-loop regulator that varies $\mu$ to maintain a ``healthy" level of gradient conflict. It consists of three components: a monotonic decay baseline, a conflict-based boost (brake), and an inertial smoother. Intuitively, this creates an ``elastic'' optimization landscape: the monotonic decay progressively sharpens the scalarization to recover precise solutions within non-convex concavities. 
Simultaneously, the dynamic brake momentarily relaxes this curvature upon high gradient interference, permitting the agent to traverse scenarios that would otherwise stall convergence.

\subsubsection{Base Decay Schedule}
We define a base schedule $\mu_{base}(t)$ that linearly anneals from a stable starting value $\mu_{start}$ to a precise terminal value $\mu_{min}$ over the course of training steps $T$:
\vspace{-2mm}
\begin{equation}
    \mu_{base}(t) = \mu_{start} - (\mu_{start} - \mu_{min}) \times \min\left(1, \frac{t}{T}\right).
\end{equation}
This provides the global pressure to converge toward hard Tchebycheff scalarization.

\subsubsection{Conflict-Driven Braking}
Another core component is the dynamic braking mechanism. We define a \textbf{Conflict Threshold} $\tau$, % (e.g., $\tau=0.4$), 
representing the maximum acceptable ratio of conflicting gradients. At each step $t$, we observe the conflict ratio $\kappa_t$ from the PCGrad update.
If $\kappa_t > \tau$, the optimization landscape is deemed challenging for the current policy stability. The controller calculates a boost factor proportional to the excess conflict:
\begin{equation}
    \beta_{boost} = 
    \begin{cases} 
      \frac{\kappa_t - \tau}{1 - \tau} & \text{if } \kappa_t > \tau, \\
      0 & \text{otherwise.}
   \end{cases}
\end{equation}
The target smoothness $\mu^*_{t}$ is then interpolated between the base schedule and the maximum stability value $\mu_{max}$:
\begin{equation}
    \mu^*_{t} = \mu_{base}(t) + \beta_{boost} \cdot (\mu_{max} - \mu_{base}(t)).
\end{equation}
This mechanism effectively ``brakes" the annealing process, reverting to a more linear-like (convex) formulation when gradient conflicts threaten to destabilize the update.

\subsubsection{Inertial Smoothing}
To prevent rapid oscillations in the objective function, which could violate the stationary distribution assumption of PPO, we apply an Exponential Moving Average (EMA) to the controller's output. The final operational $\mu_t$ is updated as:
\begin{equation}
    \mu_{t} = (1 - \lambda) \mu_{t-1} + \lambda \mu^*_{t}.
\end{equation}
where $\lambda$ is a small smoothing factor. % (e.g., 0.05).

\subsection{Controller Dynamics}
As shown in Fig. \ref{fig:adaptive_smoothness_dynamics}, this control law generates distinct behaviors we found to be critical for adaptive scalarization:
\begin{itemize}
    \item \textit{Elastic Recovery:} In response to transient spikes in conflict (e.g., entering a new area of the state space), $\mu$ rapidly increases to stabilize the gradients, then decays elastically back to the base schedule as the agent adapts.
    \item \textit{Sustained Braking:} If the agent encounters a fundamentally high-conflict trade-off (a ``wall"), the controller maintains a high $\mu$, forcing a linear-like approximation that allows the agent to bypass the local minima before attempting to resolve the non-convexity again.
\end{itemize}

The complete integration seeks that the scalarization curvature ($\mu$) matches the agent's current ability to resolve gradient conflicts, enabling convergence to the non-convex Pareto front.

See Appendix \ref{sec:appendix-astch-pseudocode} for \textit{pseudocode} (Alg. \ref{alg:full_stch_ppo}) describing the integration of our multi-objective policy optimization algorithm with the adaptive STCH smoothness.

\section{Results} \label{sec:results}

\begin{figure*}
    \centering
    \includegraphics[width=\linewidth]{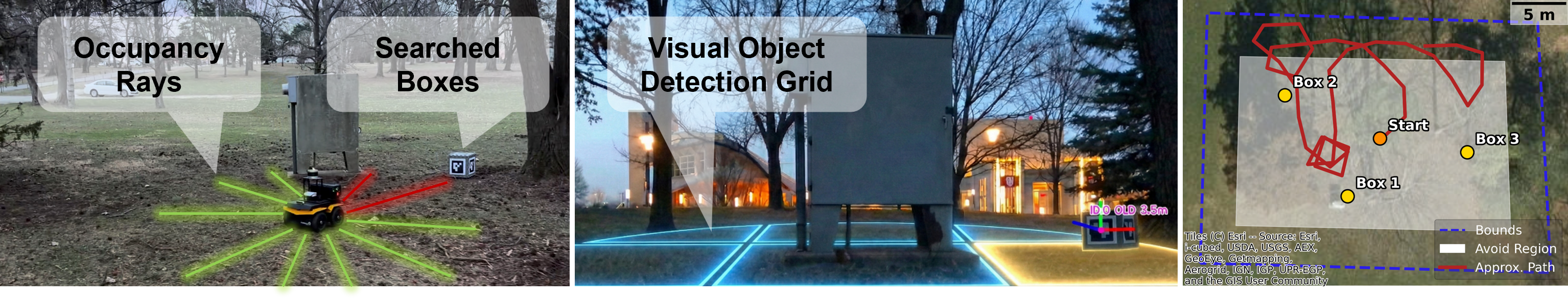} 
    \caption{\small Real-world Stealth Visual Search Experiments. \textit{(Left)} Task setup showing the UGV, searched items and the occupancy ray observation space derived from LiDAR. \textit{(Center)} Onboard camera view showing the grid-representation of the object detection pipeline. \textit{(Right)} Trajectory aerial view. The robot path (red) minimizes intrusion into the ``Avoid Region" (white), entering only as necessary to localize the targets.}
    \label{fig:real-world-stealth}
    \vspace{-2mm}
\end{figure*}

We evaluate the proposed PASTA algorithm across a diverse suite of continuous multi-objective environments, encompassing ground-based stealth visual search, quadrotor navigation among dynamic obstacles, and high-dimensional locomotion tasks. Our analysis focuses on three key dimensions: \textbf{(1)} the ability to recover optimal policies on non-convex Pareto fronts, \textbf{(2)} the sample efficiency improvement, and \textbf{(3)} the robustness of the method to architectural choices. 

Notably, PASTA \textit{uses the same hyperparameters} across all experiments and environments (see Appendix \ref{sec:appendix-hyperparams}), highlighting that our results do not depend on case-specific fine-tuning.

\subsection{Baselines}
\label{subsec:baselines}

We benchmark PASTA against diverse scalarization approaches using identical PPO architectures (see Appendix \ref{sec:appendix-network-arch}) to ensure a fair comparison. These baselines include: \textbf{Linear Scalarization}, which is fundamentally just the standard PPO algorithm, but with the scalarization step (dynamically normalizing and collapsing the reward components into a weighted sum) taking place inside the algorithm rather than within the RL environment—a formulation that inherently fails to explore concave regions of the Pareto front; \textbf{Tchebycheff Scalarization}, which can recover the full front but suffers from gradient starvation and instability; and \textbf{Fixed Smooth Tchebycheff (STCH)} across a spectrum of smoothing parameters $\mu \in \{0.01, 0.1, 0.5, 1.0, 5.0, 10.0\}$. The latter serves to verify if PASTA effectively automates the trade-off search between approximation error and optimization smoothness.

\subsection{Performance Metrics}
\label{subsec:metrics}

We evaluate convergence, diversity, and robustness using a suite of complementary metrics.
Our primary measure is \textbf{Hypervolume} (HV), which captures both convergence to the Pareto front and solution diversity via the dominated objective-space volume.
To assess reliability across preferences, we report the \textbf{Target Dominance Rate} (Win Rate), which measures the fraction of evaluated preferences for which a method attains the highest mean HV, and the \textbf{Objective Dominance Rate}, which quantifies how frequently individual objectives are maximized without excessive trade-offs, i.e., overly sacrificing secondary objectives.
To aggregate performance across seeds and environments while mitigating outlier effects, we use \textbf{Dolan-Moré Performance Profiles} (DMP), reporting the Area Under the Curve (\textbf{AUC}) as a robustness indicator.
For completeness, we additionally report \textbf{Expected Utility} (EU), noting that it may be biased by differing reward scales, rendering it potentially misleading in non-convex MOPs.
Full metric definitions are provided in Appendix \ref{sec:appendix-metrics}.

\begin{figure*}
{
    \centering
    \includegraphics[width=0.98\linewidth]{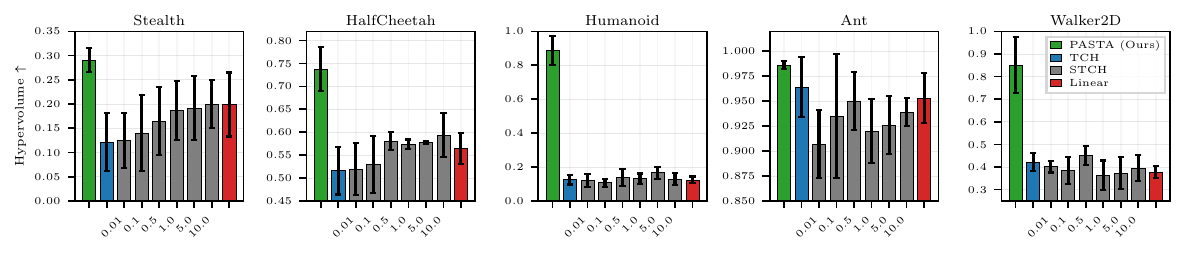}
    \vspace{-2mm}
    \caption{\small Performance on Stealth Visual Search and MuJoCo multi-objective environments. Results are aggregated across five seeds.\vspace{-8pt}
    }
    \label{fig:stealth_baselines_hv_bars}
}
\end{figure*}

\subsection{Evaluation in Stealth Visual Search}
\label{subsec:stealth_env_results}

We evaluate the PASTA algorithm in the real-world with a differential-drive ground-robot traversing unstructured terrain. The robot is tasked with searching for randomly scattered objects while simultaneously minimizing its exposure. This environment serves as a proxy for critical real-world deployments, such as an autonomous rover monitoring endangered species in a fragile habitat or a search-and-rescue robot navigating a radiologically contaminated disaster zone. Specifically, this mission requires the agent to dynamically balance three conflicting objectives: (1) \textit{Visual Object Detection} (maximizing the number of unique targets found), (2) \textit{Stealth} (minimizing exposure time to prevent ecosystem disruption or radiation contamination), and (3) \textit{Exploration} (ensuring thorough workspace coverage). 

The robot operates within a bounded arena, defined by $[-x_{\text{dim}}, x_{\text{dim}}] \times [-y_{\text{dim}}, y_{\text{dim}}]$, and is controlled via a continuous action vector $a_t = [v, \omega]$ that dictates its linear ($v \in [0, 1]$) and angular ($\omega \in [-1, 1]$) velocities. At each time step, the agent receives an observation vector $s_t$ comprising its proprioceptive state---its position $(x, y)$ and heading $(\cos \theta, \sin \theta)$---alongside exteroceptive sensor data. This sensor suite features a flattened $2 \times 3$ spatial grid over a $98^\circ$ field of view to encode detected objects, and a LIDAR point cloud downsampled to 20 rays for obstacle avoidance. The policy was trained in a simulation environment built upon the Gymnasium framework. Further details are provided in Appendix \ref{sec:appendix-stealth-results}.

The real-world experimental scenarios are visualized in Fig. \ref{fig:real-world-stealth} and \textbf{Video 1}\footnote{\textbf{Video 1}: \url{https://youtu.be/xLynYv05Bdc}}.
We validate the agent across \textit{eight} diverse preference vectors $\mathbf{w}$, ranging from balanced policies to extreme specialization.  
Fig. \ref{fig:stealth_baselines_hv_bars} illustrates the Hypervolume performance, while Table~\ref{tab:summary-stealth-results} details the complete statistics across our metrics.
Our method, PASTA, significantly outperforms all baselines, achieving a Hypervolume of $0.291$, which constitutes a $\mathbf{45.5\%}$ improvement over the strongest baseline (STCH $\mu=10.0$).
This superiority is further evidenced by a $\mathbf{100\%}$ Win Rate and an Objective Dominance Rate of $\mathbf{58.3\%}$, vastly exceeding the next best method ($12.5\%$).
Regarding the baselines, we observe interesting behavioral trends in the bar plot: STCH with small smoothness values (e.g., $\mu=0.01$) performs nearly identically to the standard TCH approach ($0.125$ vs $0.122$ HV).
Conversely, as $\mu$ increases to $10.0$, performance aligns closely with the Linear baseline ($0.200$ vs $0.199$ HV), providing empirical evidence that STCH transitions from Tchebycheff-like to linear-like behavior as the smoothness parameter moves away from zero.

Due to space constraints, we provide a detailed breakdown of results per preference vector (Table~\ref{table:stealth-preference-results}) and additional training curves (Fig.~\ref{fig:stealth_training_performance_metrics}) in Appendix \ref{sec:appendix-stealth-results-evidence}.
Notably, it also includes radar chart comparisons (Fig.~\ref{fig:radar_charts}) against the strongest baselines.
These visualizations confirm the robustness of our approach: in most tested cases, PASTA achieves the highest utility in at least two out of the three optimized objectives, demonstrating superior trade-off management.

\begin{table}[h]
{
\centering
\fontsize{6.5pt}{7.5pt}\selectfont
\setlength{\tabcolsep}{3pt}
\caption{\small Comparison against the baselines. % in the Stealth Visual Search task.
Statistics computed over eight preference vectors across five seeds.}
\label{tab:summary-stealth-results}
\begin{tabular}{l c c c c}
\toprule
Method & Hypervolume $\uparrow$ & Win Rate $\uparrow$ [\%] & Obj. Dom. Rate $\uparrow$ [\%] & DMP AUC $\uparrow$ \\
\midrule
PASTA (Ours) & \textbf{0.291 $\pm$ 0.024} & \textbf{100.0} & \textbf{58.3} & \textbf{0.981} \\
Tchebycheff & \textcolor{gray}{0.122 $\pm$ 0.060} & \underline{0.0} & \textcolor{gray}{0.0} & \textcolor{gray}{0.325} \\
STCH (0.01) & 0.125 $\pm$ 0.057 & \underline{0.0} & \textcolor{gray}{0.0} & 0.367 \\
STCH (0.1) & 0.140 $\pm$ 0.079 & \underline{0.0} & 8.3 & 0.434 \\
STCH (0.5) & 0.165 $\pm$ 0.070 & \underline{0.0} & 8.3 & 0.520 \\
STCH (1.0) & 0.187 $\pm$ 0.061 & \underline{0.0} & 4.2 & 0.611 \\
STCH (5.0) & 0.192 $\pm$ 0.066 & \underline{0.0} & 4.2 & 0.642 \\
STCH (10.0) & \underline{0.200 $\pm$ 0.050} & \underline{0.0} & \underline{12.5} & \underline{0.700} \\
Linear & 0.199 $\pm$ 0.066 & \underline{0.0} & 4.2 & 0.678 \\
\bottomrule
\end{tabular}  %\vspace{-10pt}
}
\end{table}

\subsubsection{Ablation Studies} \label{subsec:ablations}

Table \ref{table:ablation_studies_summary} summarizes the ablation analysis, validating the contribution of each component in our framework. Appendix \ref{sec:appendix-ablations} provides more detailed breakdowns. 

First, we observe that mitigating \textit{gradient interference} is critical; removing PCGrad results in a drop in most metrics, confirming the usefulness of relying on gradient projection.
In the \textit{Controller} analysis, while the ``No Conflict'' variant achieves a comparable HV to PASTA, our full method maintains a higher Win Rate ($\mathbf{50.0\%}$), suggesting that the conflict signal contributes to robustness.
Notably, removing the decay schedule (``No Decay'') degrades performance the most, confirming that temporal convergence towards small smoothness values (ideal to closely approximate standard TCH) is key and can be achieved successfully with our controller. 
Regarding the \textit{Critic}, the results reveal a two-fold insight.
First, architecture plays a primary role: the ``Branched'' baselines significantly outperform the ``Shared'' variants ($\textbf{0.277}$ vs $0.249$ average HV), confirming that distinct value heads are necessary to decouple conflicting objective estimates.
However, architecture alone is insufficient; our proposed attention-weighted loss provides the decisive advantage.
Adding this weighting to the branched architecture (PASTA) amplifies performance, increasing the HV by $\mathbf{10.2\%}$ and boosting the DMP AUC to $\mathbf{0.957}$.
This demonstrates that aligning the critic's optimization focus with the actor's scalarization is just as critical as the architectural separation.
Finally, the \textit{Smoothness} comparison highlights the benefit of adaptivity.
While PASTA matches the \textit{average} performance of the best fixed scalarization ($\mu=5.0$), the detailed breakdown in Appendix \ref{sec:appendix-ablation-fixedmu} reveals that the latter's success is skewed by specific outliers.
In fact, as shown in Table \ref{tab:ablations-fixedmu}, for some preference vectors, $\mu=5.0$ yielded the worst HV performance of all configurations.
This underscores that PASTA achieves high performance through robustness rather than chance, effectively automating the otherwise extensive and expensive search for the optimal trade-off surface without the instability and uncertainty inherent to fixed parameters.

\subsubsection{Sensitivity Analysis}
\label{subsec:sensitivity}

Table \ref{table:sensitivity_analysis_summary} summarizes the sensitivity of our algorithm to its three core hyperparameters. We find that \textit{for all choices of hyperparameters our method still beats all the baselines}, speaking to its robustness. In general, we see a distinct performance region for the Conflict Ratio ($\kappa$), indicating that values just below the midpoint of the range are sufficient, while larger values remain viable but suboptimal. For the Moving Average factor ($\lambda$), we confirm that a stable, slowly-updating memory of gradient conflict is preferred over reactive signals. Finally, the Maintenance Rate ($\rho$), shows that while the controller is robust to a wide range of values, a minimum threshold is essential to prevent performance degradation, which we hypothesize arises due to vanishing gradients from STCH. See Appendix \ref{sec:appendix-sensitivity} for additional results, including a study of the effect of the utopia point ($z^*_i$) selection.

\begin{table}[h]
\centering
\fontsize{6pt}{7pt}\selectfont
\setlength{\tabcolsep}{3pt}
\caption{\small Ablation Studies Results. Statistics computed over eight preference vectors across five seeds.}
\label{table:ablation_studies_summary}
\begin{tabular}{l c c c c}
\toprule
Method & Hypervolume $\uparrow$ & Win Rate $\uparrow$ [\%] & Obj. Dom. Rate $\uparrow$ [\%] & DMP AUC $\uparrow$ \\

% Section 3: Sensitivity PCGrad
\midrule\multicolumn{5}{c}{\textbf{PCGrad}} \\\midrule
PASTA (Ours) & \textbf{0.291 $\pm$ 0.024} & \textbf{87.5} & \textbf{58.3} & \textbf{0.962} \\
No PCGrad & \textcolor{gray}{0.187 $\pm$ 0.065} & \textcolor{gray}{0.0} & \textcolor{gray}{12.5} & \textcolor{gray}{0.598} \\
Weighted PCGrad & \underline{0.270 $\pm$ 0.019} & \underline{12.5} & \underline{29.2} & \underline{0.914} \\

% Section 3: Sensitivity Controller
\midrule\multicolumn{5}{c}{\textbf{Controller}} \\\midrule
PASTA (Ours) & \underline{0.291 $\pm$ 0.024} & \textbf{50.0} & \textbf{41.7} & \underline{0.935} \\
No Conflict & \textbf{0.295 $\pm$ 0.023} & \underline{37.5} & \underline{33.3} & \textbf{0.944} \\
No Decay & \textcolor{gray}{0.273 $\pm$ 0.022} & \textcolor{gray}{12.5} & \textcolor{gray}{25.0} & \textcolor{gray}{0.897} \\

% Section 3: Sensitivity Critic
\midrule\multicolumn{5}{c}{\textbf{Critic}} \\\midrule
PASTA (Ours) & \textbf{0.291 $\pm$ 0.024} & \textbf{75.0} & \textbf{45.8} & \textbf{0.957} \\
Branched Unweighted & \underline{0.264 $\pm$ 0.019} & \underline{25.0} & \underline{20.8} & \underline{0.895} \\
Shared Unweighted & \textcolor{gray}{0.246 $\pm$ 0.018} & \textcolor{gray}{0.0} & \underline{20.8} & \textcolor{gray}{0.840} \\
Shared Weighted & 0.253 $\pm$ 0.017 & \textcolor{gray}{0.0} & \textcolor{gray}{12.5} & 0.868 \\

% Section 3: Sensitivity Static Smoothness
\midrule\multicolumn{5}{c}{\textbf{Smoothness}} \\\midrule
PASTA (Ours) & \textbf{0.291 $\pm$ 0.024} & \textbf{37.5} & \underline{25.0} & \underline{0.892} \\
$\mu=0.01$ & 0.276 $\pm$ 0.007 & \textcolor{gray}{0.0} & \textcolor{gray}{4.2} & 0.858 \\
$\mu=0.1$ & 0.278 $\pm$ 0.024 & \underline{12.5} & 8.3 & 0.858 \\
$\mu=0.5$ & 0.280 $\pm$ 0.029 & \textcolor{gray}{0.0} & 8.3 & 0.861 \\
$\mu=1.0$ & 0.281 $\pm$ 0.031 & \underline{12.5} & 16.7 & 0.869 \\
$\mu=5.0$ & \underline{0.290 $\pm$ 0.026} & 37.5 & \textbf{29.2} & \textbf{0.893} \\
$\mu=10.0$ & \textcolor{gray}{0.262 $\pm$ 0.025} & \textcolor{gray}{0.0} & 8.3 & \textcolor{gray}{0.826} \\
\bottomrule
\end{tabular}  \vspace{-10pt}
\end{table}

\begin{table}[H]
\centering
\fontsize{6pt}{7pt}\selectfont
\setlength{\tabcolsep}{3pt}
\caption{\small Sensitivity Analysis Results. Statistics computed over eight preference vectors across five seeds.}
\label{table:sensitivity_analysis_summary}
\begin{tabular}{l c c c c}
\toprule
Method & Hypervolume $\uparrow$ & Win Rate $\uparrow$ [\%] & Obj. Dom. Rate $\uparrow$ [\%] & DMP AUC $\uparrow$ \\

% Section 1: Sensitivity Conflict Ratio
\midrule\multicolumn{5}{c}{\textbf{Conflict Ratio} $\kappa$} \\\midrule
$\kappa = 0.4$ (Ours) & \textbf{0.291 $\pm$ 0.024} & \textbf{50.0} & \textbf{37.5} & \textbf{0.929} \\
$\kappa = 0.2$ & \textcolor{gray}{0.266 $\pm$ 0.020} & \textcolor{gray}{0.0} & 20.8 & \textcolor{gray}{0.874} \\
$\kappa = 0.6$ & 0.276 $\pm$ 0.019 & \underline{37.5} & \textcolor{gray}{16.7} & 0.901 \\
$\kappa = 0.8$ & \underline{0.280 $\pm$ 0.016} & 12.5 & \underline{25.0} & \underline{0.907} \\

% Section 2: Sensitivity Moving Average $\alpha$
\midrule\multicolumn{5}{c}{\textbf{Moving Average $\lambda$}} \\\midrule
$\lambda=0.05$ (Ours) & \textbf{0.291 $\pm$ 0.024} & \textbf{62.5} & \underline{33.3} & \textbf{0.941} \\
$\lambda=0.2$ & \textcolor{gray}{0.265 $\pm$ 0.026} & \underline{25.0} & \textcolor{gray}{29.2} & \textcolor{gray}{0.881} \\
$\lambda=0.5$ & \underline{0.281 $\pm$ 0.014} & \textcolor{gray}{12.5} & \textbf{37.5} & \underline{0.918} \\

% Section 3: Sensitivity Maitenance Rate
\midrule\multicolumn{5}{c}{\textbf{Maintenance Rate} $\rho$} \\\midrule
$\rho=0.15$ (Ours) & \textbf{0.291 $\pm$ 0.024} & \underline{25.0} & \textbf{29.2} & \underline{0.898} \\
$\rho=0.05$ & \textcolor{gray}{0.261 $\pm$ 0.025} & \textcolor{gray}{0.0} & \textcolor{gray}{4.2} & \textcolor{gray}{0.831} \\
$\rho=0.3$ & 0.276 $\pm$ 0.024 & 12.5 & 16.7 & 0.860 \\
$\rho=0.5$ & 0.281 $\pm$ 0.020 & \underline{25.0} & \underline{20.8} & 0.864 \\
$\rho=0.8$ & \underline{0.291 $\pm$ 0.024} & \textbf{37.5} & \underline{29.2} & \textbf{0.900} \\

\bottomrule
\end{tabular}  \vspace{-2pt}
\end{table}

\subsection{Evaluation on MO-Gym Environments} 

\begin{figure}[h] 
\vspace{-2mm}
 {   \centering
    \includegraphics[width=0.24\linewidth]{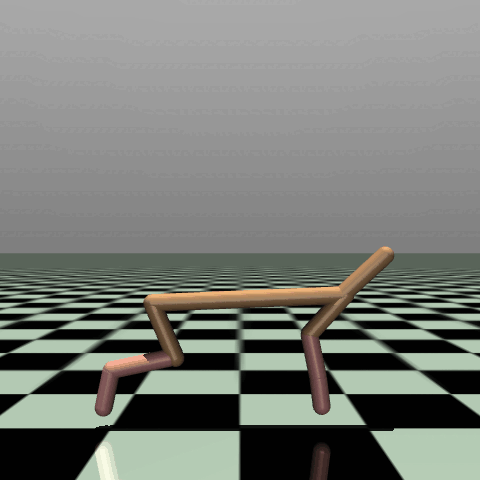}
    \includegraphics[width=0.24\linewidth]{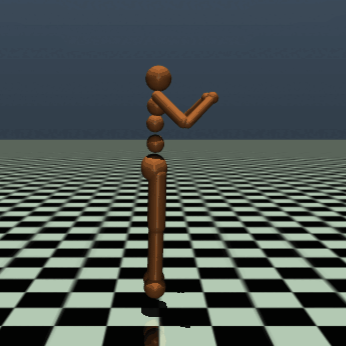}
    \includegraphics[width=0.24\linewidth]{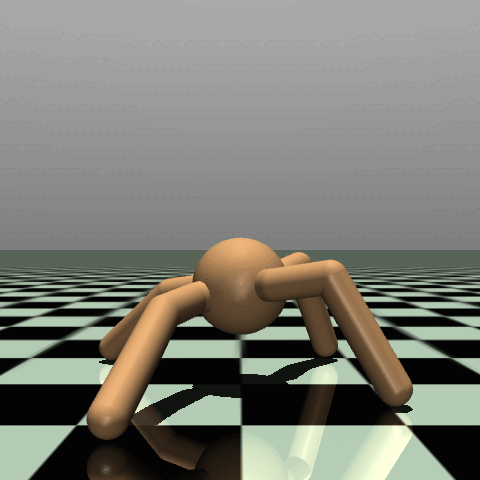}
    \includegraphics[width=0.24\linewidth]{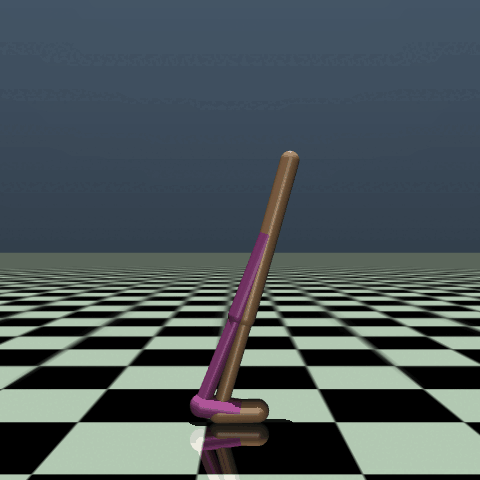}
    \caption{\small Multi-objective MuJoCo evaluation environments.}
    \label{fig:mujoco-envs}
 } 
\vspace{-6mm}
\end{figure}

To evaluate our method in additional settings, we pick continuous high-dimensional environments that have been used to assess on-policy algorithms' performance in widely used libraries like Stable Baselines 3 \cite{stable-baselines3}. Specifically, we use the multi-objective instantiations of MuJoCo's Half Cheetah, Humanoid, Ant and Walker 2D, available through MO-Gym \cite{felten_toolkit_2023} and shown in Fig. \ref{fig:mujoco-envs}. 
To ensure a fair comparison, we pick as the MOP preferences the same as the original environments, normalized between 0 and 1. Fig. \ref{fig:stealth_baselines_hv_bars} shows the hypervolume comparison for our method and the baselines in the four environments. Overall, we can see PASTA achieves a better performance. See Appendix \ref{sec:appendix-mujoco-results} for more details and results.

\subsection{Quadrotor Flights among Random Agents}

\begin{figure}
    \centering
    \includegraphics[width=0.95\linewidth]{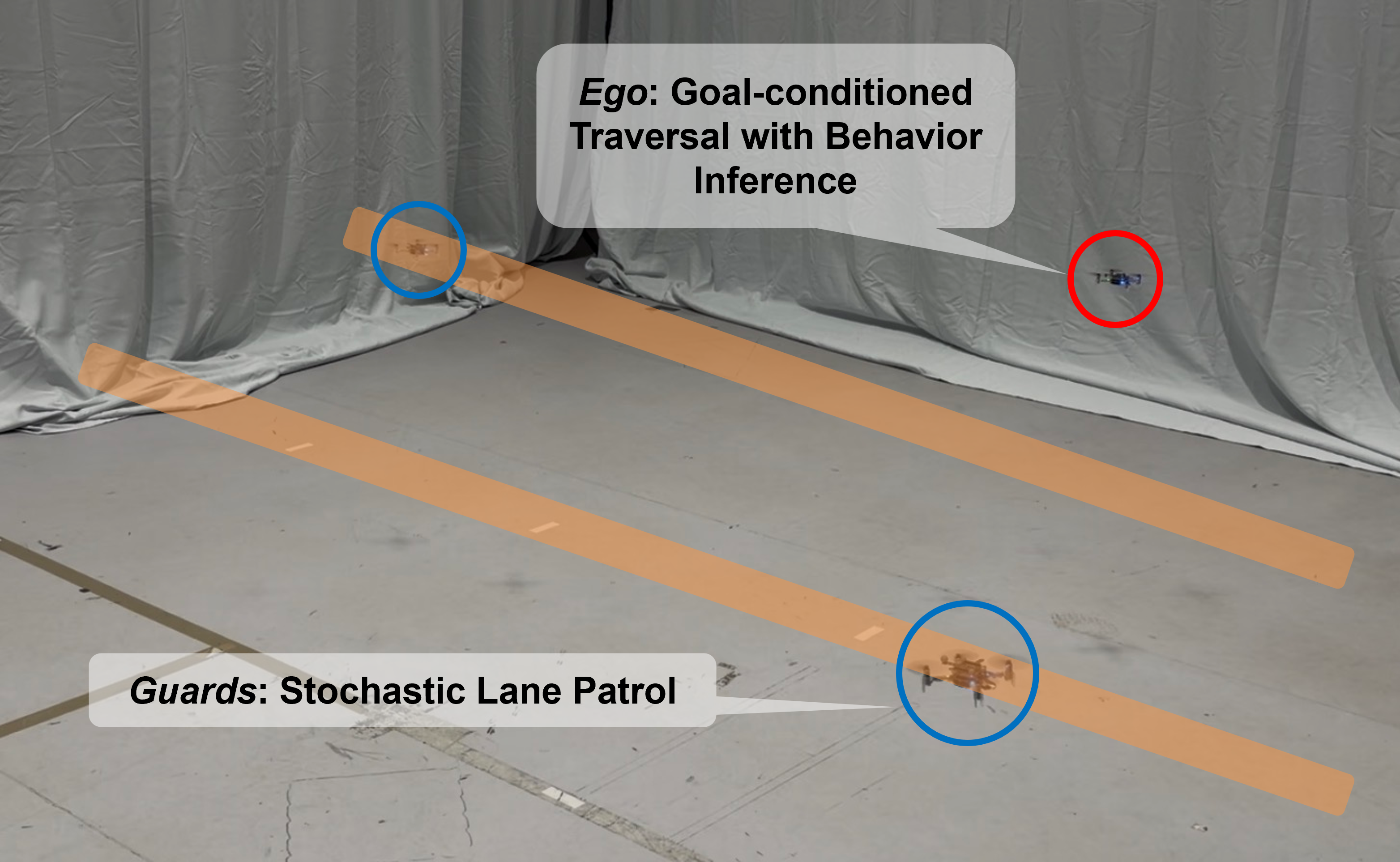}
    \caption{\small Goal-conditioned Quadrotor Traversal among Dynamic Obstacles.}
    \label{fig:drone_setup}
    \vspace{-4mm}
\end{figure}

To demonstrate the real-world applicability of PASTA across different platforms, we deployed it on a quadrotor testbed, shown in Fig. \ref{fig:drone_setup} and \textbf{Video 2}\footnote{\textbf{Video 2}: \url{https://youtu.be/twh3YZ6eczo}}. Specifically, we evaluate our approach in two multi-objective Crazyflie environments: \textit{Frogger} and \textit{Formation}. 

In the \textit{Frogger} scenario, the ego drone must traverse a shared workspace while avoiding two other drones executing independent tasks within assigned lanes. Relying solely on a history of the other agents' positions, the ego agent's policy must predict their behavior and plan accordingly to safely weave through active traffic composed of decentralized, independent actors.  In the \textit{Formation} scenario, a team of $3$ Crazyflies must navigate from a starting zone to a target zone. Throughout the maneuver, the team must maintain an equilateral triangle formation, avoid a dynamic obstacle, and strictly remain within the bounds of the flight arena.

A central challenge in both environments is the presence of dynamic opponents acting as stochastic obstacles. These opponents are governed by a 1D patrolling policy and are constrained to move along a fixed horizontal axis (e.g., $y \in \{-0.3, 0.3\}$ for the \textit{Frogger} task, and $y=0$ for the \textit{Formation} task) with a constant speed magnitude ($|\dot{x}| \in \{0.03, 0.04\}$ m/s). Their motion follows a ``bounce'' logic, where velocity is deterministically inverted upon reaching the arena boundaries ($x_{\text{lim}} \approx \pm 0.95$). Additionally, to simulate unpredictable real-world disturbances, there is a $5\%$ probability at every timestep that an opponent will spontaneously reverse its direction. This combination of boundary-constrained oscillation and stochastic direction switching requires the agents to learn collision avoidance policies capable of adapting to sudden movements.

\begin{figure}
    \centering
    \includegraphics[width=0.96\linewidth]{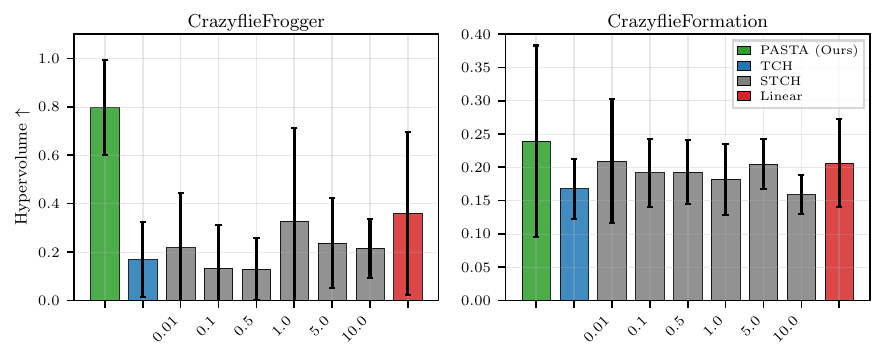}
    \caption{Performance on Crazyflie Frogger and Formation multi-objective environments. Results are aggregated across five seeds.}
    \label{fig:crazy_hypervol_comparison}
    \vspace{-4mm}
\end{figure}

Fig. \ref{fig:crazy_hypervol_comparison} illustrates the HV achieved across both tasks. Overall, our method yielded higher HV compared to the baselines. Further details regarding quadrotor deployment, RL environments, and comprehensive results are provided in Appendix \ref{sec:appendix-quadrotor-demo}.

\section{Discussion and Conclusion} \label{sec:discussion}

In this work, we addressed the tension between optimization stability and geometric precision in non-convex MORL. We proposed an \textit{Adaptive Smooth Tchebycheff} framework that utilizes a novel conflict-driven controller to dynamically modulate the curvature of the scalarization function. By treating gradient interference as a feedback signal, our method anneals towards precise non-convex geometries when feasible, while elastically braking into stable convex approximations upon conflict emergence. Moreover, we introduce the PASTA (\underline{P}olicy-optimization via \underline{A}daptive \underline{S}mooth \underline{T}chebycheff \underline{A}ttention) algorithm which integrates the adaptive scalarization into a multi-objective policy's loss by deriving attention weights directly from the STCH analytical derivative. These weights prioritize the value loss toward those with the largest weighted deviation from the utopia point, but also feature a maintenance mechanism to prevent gradient starvation in non-bottleneck objectives. Overall, we showed that our framework provides a robust way for robots to autonomously resolve complex, conflicting requirements in unstructured real-world scenarios, across ground and aerial platforms, as well as in widely used benchmark environments.

\textit{Limitations:} While our framework enhances stability and geometric precision, it remains dependent on the interplay between the conflict and maintenance rate hyperparameters. The former bottlenecks gradient flow via the smoothness controller, while the latter counteracts this to preserve gradients across all objectives via the attention. The interdependence of these parameters suggests that future work should explore unifying them into a single-parameter or parameter-free feedback mechanism.
Additionally, we focused on preference-specific learning; extending to preference-conditioned policies would better support deployment in robotic use cases where priorities could evolve at decision-time.

{
\bibliographystyle{plainnat}
\bibliography{references}
}

\clearpage

\onecolumn

% ==========================================
% APPENDIX COVER PAGE
% ==========================================
\vspace*{0.15\textheight} % Push content down slightly for aesthetic centering
\begin{center}
    {\huge \textbf{Appendix}} \\[1cm]
    {\Large \textbf{Adaptive Smooth Tchebycheff Attention for Multi-Objective Policy Optimization}} \\[0.5cm]
    {\large Alejandro Murillo-González, Mahmoud Ali and Lantao Liu} \\ % Update with your names
    {\large Indiana University--Bloomington} \\
    {\large \texttt{\{almuri, alimaa, lantao\}@iu.edu}} \\ [2cm]
    {\Large \textbf{Table of Contents}} \\[0.5cm]
\end{center}

% Clickable Index using hyperref
% \ref* gets the section number without creating a nested link
\begin{itemize}[leftmargin=2cm]
    \normalsize % Slightly larger text for readability
    \item \hyperref[sec:appendix-tch-pseudocode]{\textit{Appendix \ref*{sec:appendix-tch-pseudocode}:} Gradient-Based Tchebycheff Optimization}
    \item \hyperref[sec:appendix-ppo-details]{\textit{Appendix \ref*{sec:appendix-ppo-details}:} PPO Additional Details}
    \item \hyperref[sec:appendix-stch] {\textit{Appendix \ref*{sec:appendix-stch}:} Smooth Tchebycheff Scalarization Analysis}
    \item \hyperref[sec:appendix-astch-pseudocode] {A\textit{ppendix \ref*{sec:appendix-astch-pseudocode}:} \underline{P}olicy-optimization via \underline{A}daptive \underline{S}mooth \underline{T}chebycheff \underline{A}ttention (PASTA)}
    \item \hyperref[sec:appendix-experimental-setup] {\textit{Appendix \ref*{sec:appendix-experimental-setup}:} Experimental Setup}
    \item \hyperref[sec:appendix-metrics] {\textit{Appendix \ref*{sec:appendix-metrics}:} Performance Metrics}
    \item \hyperref[sec:appendix-stealth-results] {\textit{Appendix \ref*{sec:appendix-stealth-results}:} Extended Results on Stealth Visual Search Environment}
    \item \hyperref[sec:appendix-ablations] {\textit{Appendix \ref*{sec:appendix-ablations}:} Extended Ablation Studies}
    \item \hyperref[sec:appendix-sensitivity] {\textit{Appendix \ref*{sec:appendix-sensitivity}:} Extended Sensitivity Analysis}
    \item \hyperref[sec:appendix-mujoco-results] {\textit{Appendix \ref*{sec:appendix-mujoco-results}:} Extended Results on MuJoCo Environments}
    \item \hyperref[sec:appendix-quadrotor-demo] {\textit{Appendix \ref*{sec:appendix-quadrotor-demo}:} Extended Results on Quadrotor Flights among Random Agents}
\end{itemize}
\vspace*{0.15\textheight}
\clearpage
% ==========================================

\twocolumn
\appendix

\subsection{Gradient-Based Tchebycheff Optimization} \label{sec:appendix-tch-pseudocode}

Alg. \ref{alg:tchebycheff_step} outlines the SGD updates under Tchebycheff scalarization. This approach suffers from gradient instability due to the non-differentiable switches in the $\argmax$ operator (Line 3) \cite{lin2024smooth}. Additionally, it depends on a theoretical utopia point $\mathbf{z}^*$, which effectively requires heuristic online estimation.

\begin{algorithm}[h]
    \caption{\small Tchebycheff SGD Step}
    \label{alg:tchebycheff_step}
    \begin{algorithmic}[1]
        \Require Parameters $\mathbf{x}$, preference weight $\mathbf{w}$, utopia point $\mathbf{z}^*$, learning rate $\alpha$
        \State Compute all objective values: $\mathbf{v} = \mathbf{F}(\mathbf{x})$
        \State Compute weighted deviations:
            \[
            d_i = w_i (v_i - z_i^*) \quad \text{for} \quad i=1, \dots, m
            \]
        \State Find index of worst objective: $j = \argmax_{i} \{ d_i \}$
        \State Compute gradient for that objective: $\mathbf{g}_j = \nabla_{\mathbf{x}} f_j(\mathbf{x})$
        \State $\mathbf{g} \leftarrow w_j \cdot \mathbf{g}_j$ \Comment{Get scaled \textit{sub}gradient}
        \State $\mathbf{x} \leftarrow \mathbf{x} - \alpha \cdot \mathbf{g}$ \Comment{Perform gradient descent step}
    \end{algorithmic}
\end{algorithm}

\subsection{PPO Additional Details} \label{sec:appendix-ppo-details}

The $\min$ operation from the clipped loss (Eq. \ref{eq:ppo-clip-loss}) creates a pessimistic bound. If the advantage $A^{\pi_{\theta_k}}$ is positive (a good action), the objective is clipped at $(1 + \epsilon) A^{\pi_{\theta_k}}$. If the advantage is negative (a bad action), the objective is clipped at $(1 - \epsilon) A^{\pi_{\theta_k}}$, preventing an over-correction.

Alg. \ref{alg:ppo} presents a common practical implementation that, while equivalent, differs slightly in its formulation. It separates the optimization into two distinct steps: one for the policy $\pi_\theta$ (actor) and one for the value function $V_\phi$ (critic), often using different learning rates ($\alpha_\theta, \alpha_\phi$). %The algorithm's $L^{VF}$ (line 23) is identical in purpose to $L^{VF}$ in Eq. \ref{eq:ppo-full-loss}. 
This separation has been found to be computationally convenient and stable \cite{SpinningUp2018}. For simplicity, the pseudocode also omits the entropy bonus $S$, which in a real implementation would be added to $L^{\text{CLIP}}$ on line 20 before the policy update.

The algorithm utilizes Generalized Advantage Estimation (GAE) \cite{schulman2015high} to compute the targets. GAE computes an estimate of the advantage function using an exponentially-weighted average of temporal difference errors. This approach is controlled by two parameters, the discount factor $\gamma$ and the trace-decay parameter $\lambda \in [0, 1]$, which balances the bias-variance trade-off. 

\begin{algorithm*}[h]
    {\small \caption{\small PPO (Clip) \cite{schulman2017proximal} with GAE \cite{schulman2015high}} \label{alg:ppo}
    \begin{algorithmic}[1]
    \State Initialize policy parameters $\boldsymbol{\theta}$ and value function parameters $\boldsymbol{\phi}$.
    \State Set hyperparameters: $\gamma$ (discount), $\lambda$ (GAE), $\epsilon$ (clip), $E$ (epochs), $M$ (mini-batch size).
    
    \For{iteration $k = 0, 1, 2, \dots$}
        \State Collect batch of $T$ timesteps $\mathcal{D}_k = \{ (s_t, a_t, r_t, s_{t+1}) \}$ by running policy $\pi_{\boldsymbol{\theta}}$.
        \State \Comment{Store $\log \pi_{\boldsymbol{\theta}}(a_t \mid s_t)$ for each step as well.}
        
        \State \Comment{--- Advantage and Value Target Calculation (GAE) ---}
        \State Set $\hat{A}_T = 0$
        \For{$t = T-1$ \textbf{down to} $0$}
            \State \Comment{Use the value network $V_{\boldsymbol{\phi}}$ as the baseline}
            \State $\delta_t = r_t + \gamma V_{\boldsymbol{\phi}}(s_{t+1}) - V_{\boldsymbol{\phi}}(s_t)$
            \State $\hat{A}_t = \delta_t + \gamma \lambda \hat{A}_{t+1}$
            \State $V^{\text{target}}_t = \hat{A}_t + V_{\boldsymbol{\phi}}(s_t)$ \Comment{Target for value function update}
        \EndFor
        
        \State \Comment{--- Policy and Value Updates ---}
        \State Let $\boldsymbol{\theta}_k = \boldsymbol{\theta}$ \Comment{Store old policy parameters}
        \For{epoch $e = 1$ \textbf{to} $E$}
            \For{mini-batch $b = \{ s_t, a_t, \log \pi_{\boldsymbol{\theta}_k}(a_t \mid s_t), \hat{A}_t, V^{\text{target}}_t \} \subset \mathcal{D}_k$}
                \State \Comment{1. Calculate Policy (Actor) Loss and Update}
                \State $r_t(\boldsymbol{\theta}) = \frac{ \pi_{\boldsymbol{\theta}} ( a_t \mid s_t) }{ \pi_{\boldsymbol{\theta}_k} (a_t \mid s_t) }$ \Comment{Ratio using current and old policies}
                \State $L^{\text{CLIP}}(\boldsymbol{\theta}) = \hat{\mathbb{E}}_{b} \Big[ \min \Big( r_t(\boldsymbol{\theta}) \hat{A}_t, \texttt{clip}(r_t(\boldsymbol{\theta}), 1 - \epsilon, 1 + \epsilon) \hat{A}_t \Big) \Big]$ \Comment{Eq. \eqref{eq:ppo-clip-loss}}
                \State $\boldsymbol{\theta} \leftarrow \boldsymbol{\theta} + \alpha_{\theta} \nabla_{\boldsymbol{\theta}} L^{\text{CLIP}}(\boldsymbol{\theta})$ \Comment{Gradient ascent (Entropy bonus $S$ often added here)}
                
                \State \Comment{2. Calculate Value (Critic) Loss and Update}
                \State $L^{VF}(\boldsymbol{\phi}) = \hat{\mathbb{E}}_{b} \Big[ (V_{\boldsymbol{\phi}}(s_t) - V^{\text{target}}_t)^2 \Big]$ \Comment{Mean-squared error (same as $L^{VF}$)}
                \State $\boldsymbol{\phi} \leftarrow \boldsymbol{\phi} - \alpha_{\phi} \nabla_{\boldsymbol{\phi}} L^{VF}(\boldsymbol{\phi})$ \Comment{Gradient descent}
            \EndFor
        \EndFor
        
    \EndFor
    \end{algorithmic}
    }
\end{algorithm*}

\subsection{Smooth Tchebycheff Scalarization Analysis}
\label{sec:appendix-stch}

\begin{figure}
    \centering
    \includegraphics[width=\linewidth]{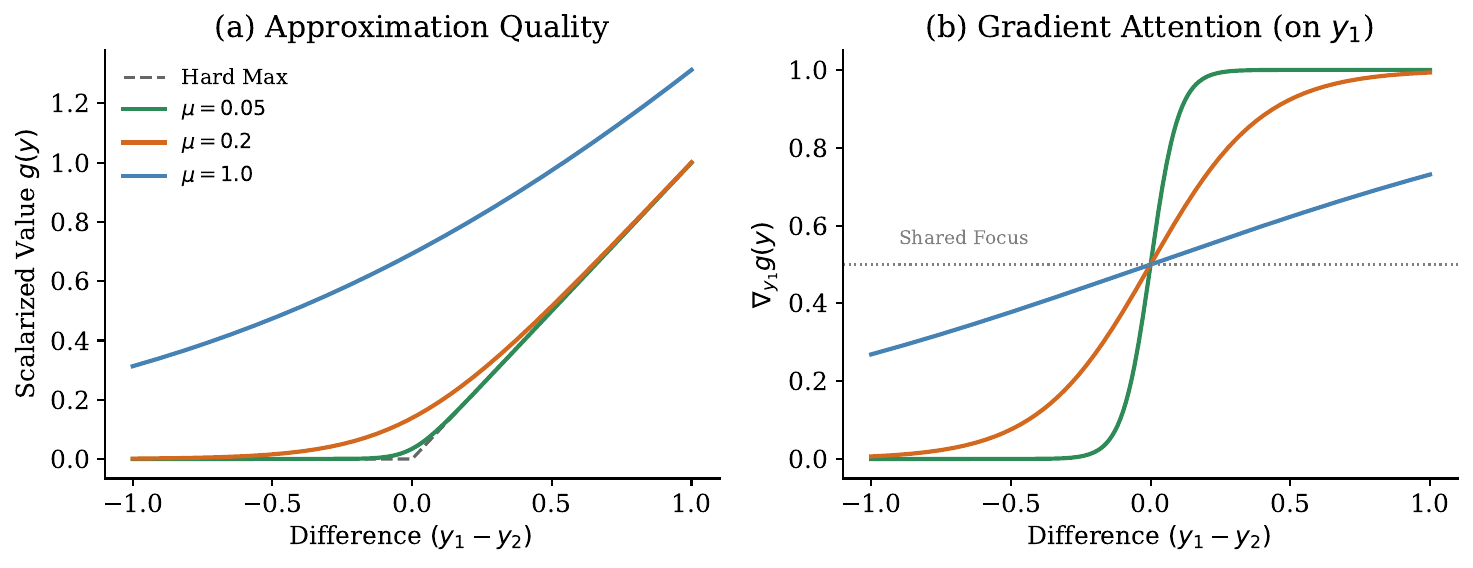}
    \caption{\small Approximation quality and gradient attention of STCH.}
    \label{fig:stch_mu_effect}
    \vspace{-6mm}
\end{figure}

To understand why the Smooth Tchebycheff (STCH) scalarization outperforms where the classical Tchebycheff (TCH) approach struggles with gradient-based optimization, we must analyze the constituent operations of the scalarization function. The STCH scalarization $g_\mu^{(STCH)}$ is defined as a composition of five distinct mathematical operations \cite{lin2024smooth}. For a maximization problem, we have: $g_{\mu}^{(STCH)}(x|\mathbf{w}) =$

\begin{equation} 
\label{eq:stch_expanded} 
\underbrace{-\mu}_{\text{5. Scale}} 
\underbrace{\log}_{\text{4. Damp}} 
\left( 
    \underbrace{\sum_{i=1}^{m}}_{\text{3. Aggregate}} 
    \underbrace{\exp}_{\text{2. Select}} 
    \left( 
        \underbrace{\frac{w_i (z_i^* - f_i(x))}{\mu}}_{\text{1. Inverse Temp.}}
    \right) 
\right) .
\end{equation}

This formulation serves as a smooth approximation of the maximum function \cite[p. 72]{boyd2004convex}. The dynamics of this approximation are controlled by the smoothing parameter $\mu$. We analyze the five operational steps %{\color{red}illustrated in Fig. \ref{fig:stch_mechanism}} 
below.

\subsubsection{The Inverse Temperature ($1/\mu$)} The inner term $\mu$ acts as an inverse temperature coefficient. This operation scales the weighted objective values $w_i (z_i^* - f_i(x))$ before they enter the exponential function. This step determines the distinctness of the objectives. A small $\mu$ amplifies small differences between conflicting objectives, while a large $\mu$ compresses them. That is, as $\mu \rightarrow 0$, even infinitesimal differences between objectives result in effectively infinite differences after scaling. This forces the subsequent steps to focus exclusively on the single worst objective. 

\subsubsection{Exponential Selection ($\exp$)} The exponential function transforms the scaled values into a non-linear space. Because $\exp$ grows super-linearly, the largest scaled value dominates the sum. This effectively ``selects" the worst-performing objective (the maximum) and suppresses the others. As illustrated in Fig. \ref{fig:stch_mu_effect}(b), this step creates an implicit attention mechanism. The gradient of the function with respect to the $i^{th}$ objective is proportional to its exponential weight: \begin{equation} 
        \frac{\partial g}{\partial f_i} \propto \frac{e^{y_i/\mu}}{\sum_j e^{y_j/\mu}}, \quad \text{where} \quad  y_i = w_i (z_i^* - f_i(x))
    \end{equation} 
This is the \textit{Softmax} function. This ensures that the optimization gradient is driven primarily by the objective that is currently performing the worst. 

\subsubsection{Aggregation ($\sum$)} The summation aggregates the exponentiated differences. It becomes dominated by the single, largest exponential term. For example, if $\exp(y_j/\mu) = 1000$ and all other terms are small (e.g., $\le 1$), the sum will be approximately 1000.
\subsubsection{Damping ($\log$)} The logarithm reverts the magnitude of the data to the original scale. While the classical Tchebycheff method takes a ``hard" maximum (ignoring all non-maximal objectives), the Log-Sum operation provides a ``soft" maximum that incorporates information from all objectives, weighted by their proximity to the worst objective.

\subsubsection{Outer Scaling ($\mu$)} The final multiplication by $\mu$ restores the original units of the objective functions. This step ensures that the approximation is bounded. As show in \cite[Propositions 3.3 \& 3.4]{lin2024smooth}, the STCH function approximates the true Tchebycheff value within a tight bound determined by $\mu$:
    \begin{equation}
        \begin{aligned}
            g_\mu^{(\mathrm{STCH})}(\mathbf{x}\mid\mathbf{w}) - \mu\log m 
                &\le g^{(\mathrm{TCH})}(\mathbf{x}\mid\mathbf{w}) \\
                &\le g_\mu^{(\mathrm{STCH})}(\mathbf{x}\mid\mathbf{w})
        \end{aligned}
    \end{equation}
where $m$ corresponds to the dimensionality of the MOP. This bound guarantees that maximizing the smooth proxy $g_\mu^{(STCH)}$ consistently maximizes the true multi-objective problem. 

\subsubsection{The Role of $\mu$ in Gradient Flow} The parameter $\mu$ acts as a control knob for the ``sharpness" of the scalarization, as shown in Fig. \ref{fig:stch_mu_effect}. When $\mu$ is large (Fig. \ref{fig:stch_mu_effect}, blue line), the function behaves like a linear average, distributing the gradient equally across objectives regardless of their value. When $\mu$ is small (Fig. \ref{fig:stch_mu_effect}, green line), the function approximates the ``Hard Max" accurately, but the gradient behaves like a step function (0 or 1), which causes oscillation in gradient-based solvers. The optimal $\mu$, which is usually unknown, creates a differentiable ``corner", allowing the optimizer to smoothly traverse the Pareto front.

\subsection{\underline{P}olicy-optimization via \underline{A}daptive \underline{S}mooth \underline{T}chebycheff \underline{A}ttention (PASTA)} \label{sec:appendix-astch-pseudocode}

Alg. \ref{alg:full_stch_ppo} summarizes the closed-loop interaction between STCH-based attention, conflict-aware updates, and adaptive smoothness.

\begin{algorithm}[h]
    \caption{\small PASTA}
    \label{alg:full_stch_ppo}
    \begin{algorithmic}[1]
        \Require Policy $\pi_\theta$, Branched Critic $V_\phi^{(1:m)}$, Preferences $\mathbf{w}$
        \State Initialize smoothness $\mu$, conflict metric $\kappa \leftarrow 0$
        
        \For{iteration $= 1, \dots, N$}
            \State Collect trajectories $\tau$ via $\pi_\theta$
            \State Compute normalized vector advantages $\hat{\mathbf{A}} \in \mathbb{R}^{m \times T}$
            
            \State \textbf{Adaptive Step:} Update $\mu \leftarrow \text{Controller}(\kappa)$ 
            \State Compute ASTCH attention $\boldsymbol{\eta}$ \Comment{Eq. \eqref{eq:stch_att_with_grad_preservation}}

            \State Initialize conflict history $\mathcal{K} \leftarrow \emptyset$ 
            
            \For{epoch $= 1, \dots, K$}
                \State \textbf{Critic Update:} 
                \State \quad $L^{VF} = \sum \eta_{t,i} (V_\phi^{(i)} - y_{t}^{(i)})^2$  \Comment{Eq. \eqref{eq:astch-value-loss}}
                \State \quad Update value $\phi \leftarrow \phi - \alpha_\phi \nabla_\phi L^{VF}$
                
                \State \textbf{Actor Update:}
                \State \quad Compute objective gradients $\{\mathbf{g}_i\}_{i=1}^m$ using $\hat{A}_i$
                \State \quad $(\{\mathbf{g}_i^{proj}\}, \kappa_{batch}) \leftarrow \text{PCGrad}(\{\mathbf{g}_i\})$ 
                \State \quad Update policy $\theta \leftarrow \theta + \alpha_\theta \sum \mathbf{g}_i^{proj}$
                \State \quad $\mathcal{K} \leftarrow \mathcal{K} \cup \{\kappa_{batch}\}$
            \EndFor
            \State Update global conflict $\kappa \leftarrow \text{mean}(\mathcal{K})$
        \EndFor
    \end{algorithmic}
\end{algorithm}

This design ensures that scalarization sharpness ($\mu$) adapts to the agent's current ability to resolve trade-offs ($\kappa$), while the attention mechanism ($\boldsymbol{\eta}$) explicitly directs value estimation capacity toward the current Pareto bottlenecks.

\subsection{Experimental Setup} \label{sec:appendix-experimental-setup}

\subsubsection{Hyperparameters}  \label{sec:appendix-hyperparams}

Across all experiments, we held the hyperparameters of PASTA constant. We utilized a maintenance rate of $\rho = 0.15$, an exponential moving average factor of $\lambda = 0.05$, and a conflict ratio of $\kappa = 0.4$. We fix the utopia point at $\zeta = 1.05$ which satisfies the recommendation $z_i^* = \zeta > 1$ from \cite[Sec.~2]{steuer1983interactive_compressed}. The decay schedule ranges from $\mu_{start} = 10.0$ to $\mu_{min} = 0.05$.

Regarding the PPO backbone: The network parameters are optimized using Adam with a learning rate of $3 \times 10^{-4}$. Data is collected over a horizon of $T = 2048$ steps per iteration, and updates are performed over $10$ epochs with a minibatch size of $64$. Regarding the PPO-specific hyperparameters (Eq. \ref{eq:ppo-update}), we set the clipping range $\epsilon = 0.2$, the value function coefficient $c_1 = 0.5$, and the entropy coefficient $c_2 = 0.01$ to encourage exploration. For advantage estimation, we employ GAE \cite{schulman2015high} with a discount factor $\gamma = 0.99$ and a smoothing parameter $\lambda = 0.95$.

\subsubsection{Network Architecture} \label{sec:appendix-network-arch}

We implement our multi-objective policy and value function approximations using neural networks conditioned on both the state $s$ and the scalarization weights $\mathbf{w}$. All networks process a concatenated input vector $[s, \mathbf{w}]$.

\textbf{Actor.}
The actor network utilizes a multi-layer perceptron (MLP) backbone with two hidden layers of $64$ units each, using Tanh activations. The network outputs the mean $\mu$ of a Gaussian distribution via a final linear layer followed by a Sigmoid activation function. The standard deviation $\sigma$ is parameterized as a state-independent, learnable log-variable optimized alongside the network weights.

\textbf{Branched Critic.}
To estimate the vector of value functions for the $m$ objectives, we employ a Branched Critic architecture. This network consists of a shared feature extraction trunk (two hidden layers of 64 units, Tanh) followed by $m$ independent heads. Each head contains an additional hidden layer (64 units, Tanh) and a final linear projection to estimate the scalar value for its specific objective.

Table \ref{tab:network_arch} summarizes the structural details of these modules.

\begin{table}[h]
    \centering
    \fontsize{6pt}{7pt}\selectfont
    \caption{\small Actor and Critic Neural Network Architectures.}
    \label{tab:network_arch}
    \begin{tabular}{l l l l}
        \toprule
        \textbf{Module} & \textbf{Input} & \textbf{Hidden Layers} & \textbf{Output Layer} \\
        \midrule
        \textbf{Continuous Actor} & & & \\
        \quad \textit{Backbone} & $[s, \mathbf{w}]$ & 2 $\times$ [Linear(64), Tanh] & N/A \\
        \quad \textit{Heads} & Backbone Feat. & N/A & $\mu$: Sigmoid, $\sigma$: Parameter \\
        \midrule
        \textbf{Branched Critic} & & & \\
        \quad \textit{Shared Trunk} & $[s, \mathbf{w}]$ & 2 $\times$ [Linear(64), Tanh] & N/A \\
        \quad \textit{Objective Heads} & Trunk Feat. & 1 $\times$ [Linear(64), Tanh] & Linear(1) (per objective) \\
        \midrule
        \textbf{Shared Critic} & & & \\
        \quad \textit{Shared Trunk} & $[s, \mathbf{w}]$ & 2 $\times$ [Linear(64), Tanh] & N/A \\
        \quad \textit{Shared Head} & Trunk Feat. & 1 $\times$ [Linear(64), Tanh] & Linear(m) (all objectives) \\
        \bottomrule
    \end{tabular}
\end{table}

\subsection{Performance Metrics}
\label{sec:appendix-metrics}

We evaluate convergence, diversity, and robustness using the following metrics, where $\mathcal{B}$ is the set of methods and $\Xi$ is the set of preference targets.

\textbf{Hypervolume (HV).} As our primary figure of merit, HV measures the size of the objective space region dominated by a solution set relative to a reference point $\mathbf{z}_{ref}$. We normalize objectives to $[0, 1]$ and use $\mathbf{z}_{ref} = \mathbf{0}$, ensuring that maximizing HV drives both convergence to the Pareto front and solution diversity as well as let us order point sets \cite{guerreiro2021hypervolume}.

\textbf{Target Dominance Rate (Win Rate).} To assess reliability, we measure the fraction of evaluated preferences for which a method attains the highest mean HV. Let $\bar{H}_{b, \mathbf{w}}$ be the mean peak hypervolume for method $b \in \mathcal{B}$. The rate is defined as $\frac{1}{|\Xi|} \sum_{\mathbf{w} \in \Xi} \mathbb{I} [\bar{H}_{b, \mathbf{w}} = \max_{b' \in \mathcal{B}} \bar{H}_{b', \mathbf{w}}]$. A high win rate indicates consistent dominance across the diverse preference spectrum $\Xi$.

\textbf{Objective Dominance Rate.} This metric provides a scale-independent ranking of boundary-seeking behavior. We verify if an algorithm maximizes individual objective $i \in \{1, \dots, m\}$ relative to baselines. Calculated as $\frac{1}{|\Xi| \times m} \sum_{\mathbf{w} \in \Xi} \sum_{i=1}^{m} \mathbb{I} [ f_i(b, \mathbf{w}) \approx \max_{b'} f_i(b', \mathbf{w}) ]$, this aggregates how frequently a method achieves the best observed value for any individual objective. 
High scores indicate that the method manages to maximize the primary objective without overly sacrificing secondary ones, i.e., trade-off management.

\textbf{Dolan-Moré Performance Profiles (DMP)} \cite{dolan2002benchmarking}\textbf{.} To aggregate performance across all seeds and environments without outlier skew, we compute a performance ratio $r_{p,b} = (\max_{\beta} HV_{p,\beta}) / HV_{p,b}$ for each problem instance $p$ and method $b \in \mathcal{B}$. We report the \textit{Area Under the Curve} (\textbf{AUC}) of the cumulative distribution of these ratios. A high AUC (closer to 1.0) denotes a method that is robust and frequently optimal.

\textbf{Expected Utility (EU).} We report the scalarized expected return $\mathbb{E}[\mathbf{w}^T \mathbf{r}]$ for completeness. However, we caution that EU could be biased by differing reward scales, rendering it a potentially misleading metric for non-convex MOPs.

\subsection{Extended Results on Stealth Visual Search Environment} \label{sec:appendix-stealth-results}

\subsubsection{Environment Details} \label{sec:appendix-stealth-results-details}

We evaluate our proposed method in a continuous 2D simulation built on the Gymnasium framework. The environment models a mobile differential-drive robot tasked with locating hidden objects in a cluttered arena while managing competing objectives of stealth, safety and exploration.

\textbf{State and Observation Space.}
The agent operates within a bounded 2D arena $[-x_{dim}, x_{dim}] \times [-y_{dim}, y_{dim}]$. The observation vector $s_t$ consists of:
\begin{itemize}
    \item \textit{Proprioception:} The agent's position $(x, y)$ and heading vector $(\cos \theta, \sin \theta)$.
    \item \textit{Visual Sensor:} A 6-dimensional vector representing a $2 \times 3$ grid over the agent's field of view (FOV $\approx 98^\circ$, range $0.6$). This provides density readings of unscanned targets.
    \item \textit{LIDAR:} A 20-ray range scanner (max range $0.35$) detecting distances to arena boundaries, unscanned targets, and static obstacles.
\end{itemize}

\textbf{Action Space and Dynamics.}
The agent is controlled via a continuous action vector $a_t = [v, \omega]$, where $v \in [0, 1]$ controls linear velocity and $\omega \in [-1, 1]$ controls angular velocity. The simulation uses Dubins dynamics for the robot with discrete time steps ($dt=0.05$). The environment populates the arena with random static obstacles (circles and rectangles) using rejection sampling. It strictly enforces collision physics; if a move results in a collision with an obstacle or boundary, the agent's position is reset to the previous step.

\textbf{Multi-Objective Reward Structure.}
The environment computes a reward vector $\mathbf{r}_t = [r_{\text{score}}, r_{\text{stealth}}, r_{\text{expl}}]^\top$ at each time step $t$, defined as follows:

\begin{itemize}
    \item \textit{Search Score ($r_{\text{score}}$):} This objective incentivizes locating targets. It combines a sparse reward for distinct discoveries with a dense signal for tracking visible targets:
    \begin{equation}
        r_{\text{score}} = \text{clip}\left( 10 \cdot N_{\text{new}} + 0.05 \sum_{k=1}^{6} v_k, \ 0, \ 10 N_{\text{obj}} \right),
    \end{equation}
    where $N_{\text{new}}$ is the count of newly scanned objects in the current step, and $v_k$ represents the intensity readings from the agent's 6-cell vision grid.

    \item \textit{Stealth ($r_{\text{stealth}}$):} This objective penalizes exposure by measuring the agent's distance from the ``safe zones" located at the arena borders. Let $p_t = (x_t, y_t)$ be the agent's position and $L_{\text{safe}}$ be the boundary of the central danger zone. The exposure distance $d_{\text{risk}}$ is defined as:
    \begin{equation}
        d_{\text{risk}} = \max\left(0, \ \min(L_{\text{safe}}^x - |x_t|, \ L_{\text{safe}}^y - |y_t|) \right).
    \end{equation}
    The stealth reward is then calculated as the normalized proximity to safety, with a penalty $\mathbb{I}_{\text{coll}}$ for collisions:
    \begin{equation}
        r_{\text{stealth}} = \text{clip}\left( \left(1 - \frac{d_{\text{risk}}}{d_{\text{max}}}\right) - \mathbb{I}_{\text{coll}}, \ 0, \ 1 \right).
    \end{equation}

    \item \textit{Exploration ($r_{\text{expl}}$):} This objective encourages movement and area coverage: 
    \begin{equation}
        r_{\text{expl}} = \text{clip}\left( 2.0 \cdot \|p_t - p_{t-1}\|_2, \ 0, \ 1 \right).
    \end{equation}
\end{itemize}

\begin{table}[h]
    \centering
    \fontsize{6.5pt}{7.5pt}\selectfont
    \caption{\small Parameters for the Stealth Visual Search Environment.}
    \label{tab:env_params}
    \begin{tabular}{l c}
        \toprule
        \textbf{Parameter} & \textbf{Value} \\
        \midrule
        Arena Dimensions & $x_{dim}=1.0, y_{dim}=1.0$ \\
        Agent Radius & $0.05$ \\
        Sensor FOV / Range & $1.715 \text{ rad} \approx 98^\circ$ / $0.6$ \\
        LIDAR Rays / Range & $20$ / $0.35$ \\
        Effective Action Space & $v \in [0, 1], \omega \in [-1, 1]$ \\
        Obstacles & Random Circles \& Rectangles \\
        \bottomrule
    \end{tabular}
\end{table}

\subsubsection{System Specifications}  \label{sec:appendix-stealth-system}

We now describe the real-world experimental setups. The stealth visual search task is validated using a mobile robotics platform across distinct outdoor environments. We employ the Clearpath Jackal, a differential-drive UGV with a control input vector $\mathbf{u} = [v, \omega]^\top \in [-1, 1]^2$ representing the linear and angular velocities, respectively. The system operates on the ROS 2 middleware and is equipped with a Velodyne 16-beam LiDAR which serves a dual purpose: it facilitates real-time localization via the \texttt{KISS-ICP} package \cite{vizzo2023ral}, and provides point cloud data as input to the policy for obstacle avoidance. Additionally, an Intel RealSense D435i RGB-D camera is utilized for object detection. To maintain focus on the search algorithm rather than computer vision contributions, ArUco markers are used to simulate target objects, providing a robust method for visual target detection and localization. The testing environments feature targets hidden among trees (boxes with ArUco markers) on uneven terrain consisting of mud and grass. Here, virtual borders restrict the robot to stealth corridors peripheral to a protected region, ensuring it maintains proximity to cover while strictly limiting interaction with the off-limits area.

\subsubsection{Results} \label{sec:appendix-stealth-results-evidence}
We validated the agent across eight diverse preference vectors $\mathbf{w}$, ranging from balanced policies ($[1/3, 1/3, 1/3]$) to extreme specializations (e.g., Stealth-focused $[0.1, 0.7, 0.2]$). Full preference list: $[0.1, 0.7, 0.2]$, $[0.2, 0.2, 0.6]$, $[0.2, 0.6, 0.2]$, $[1/3, 1/3, 1/3]$, $[0.4, 0.4, 0.2]$, $[0.5, 0.3, 0.2]$, $[0.6, 0.3, 0.1]$, $[0.8, 0.1, 0.1]$. 

\begin{figure*}[t]
    \centering
    
    % --- First Row ---
    \begin{subfigure}[b]{0.485\textwidth}
        \centering
        \includegraphics[width=\linewidth]{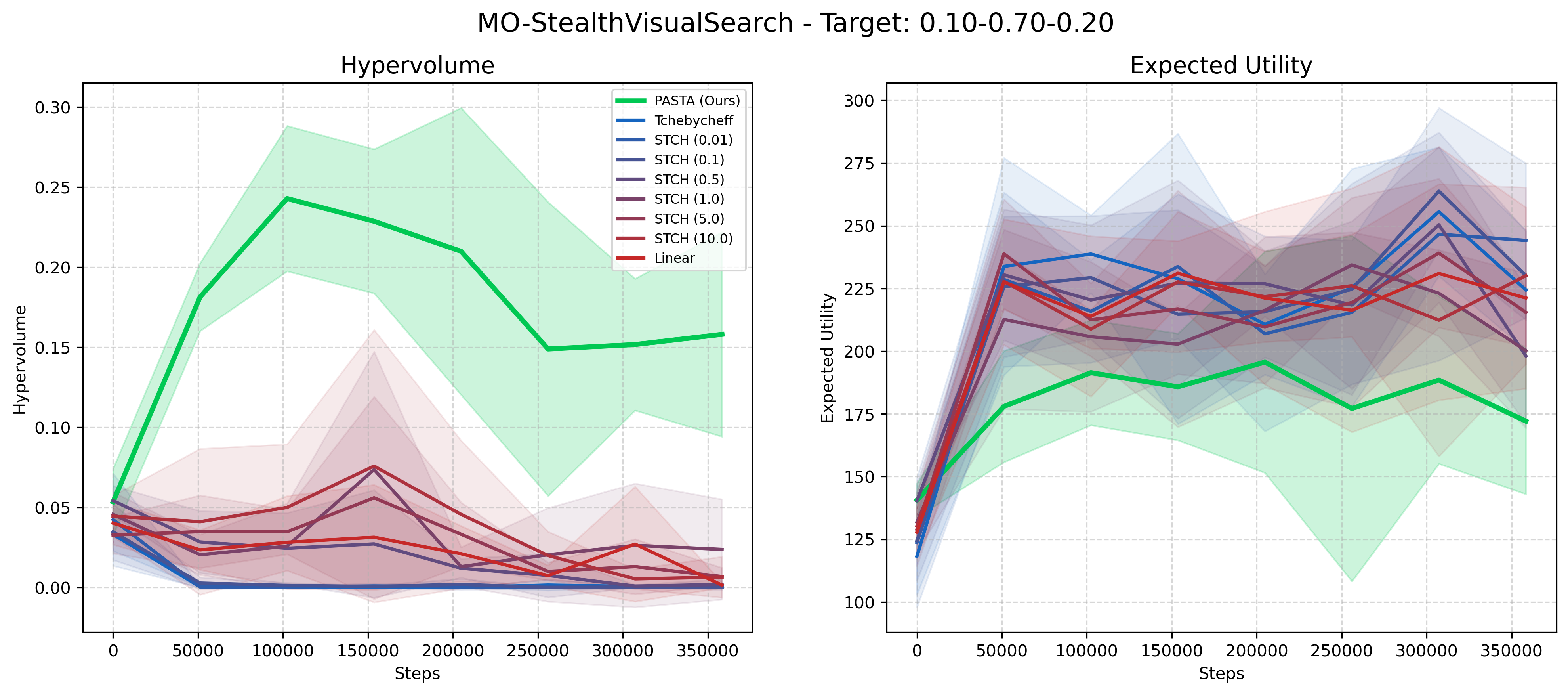}
        \caption{\small 0.10-0.70-0.20}
        \label{fig:top-left}
    \end{subfigure}
    \hfill
    \begin{subfigure}[b]{0.485\textwidth}
        \centering 
        \includegraphics[width=\linewidth]{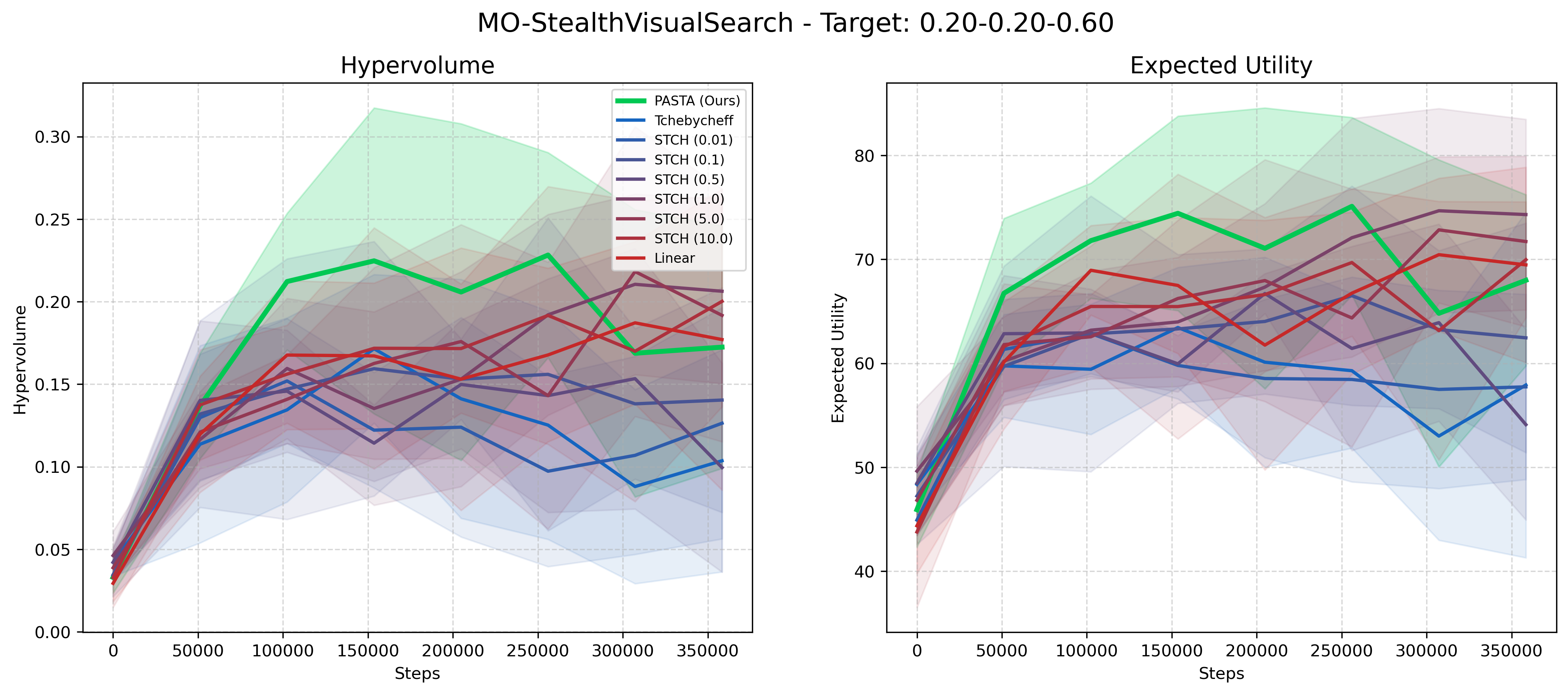}  
        \caption{\small 0.20-0.20-0.60}
        \label{fig:top-right}
    \end{subfigure}
    
    \vspace{0.2cm}
    
    % --- Second Row ---
    \begin{subfigure}[b]{0.485\textwidth}
        \centering
        \includegraphics[width=\linewidth]{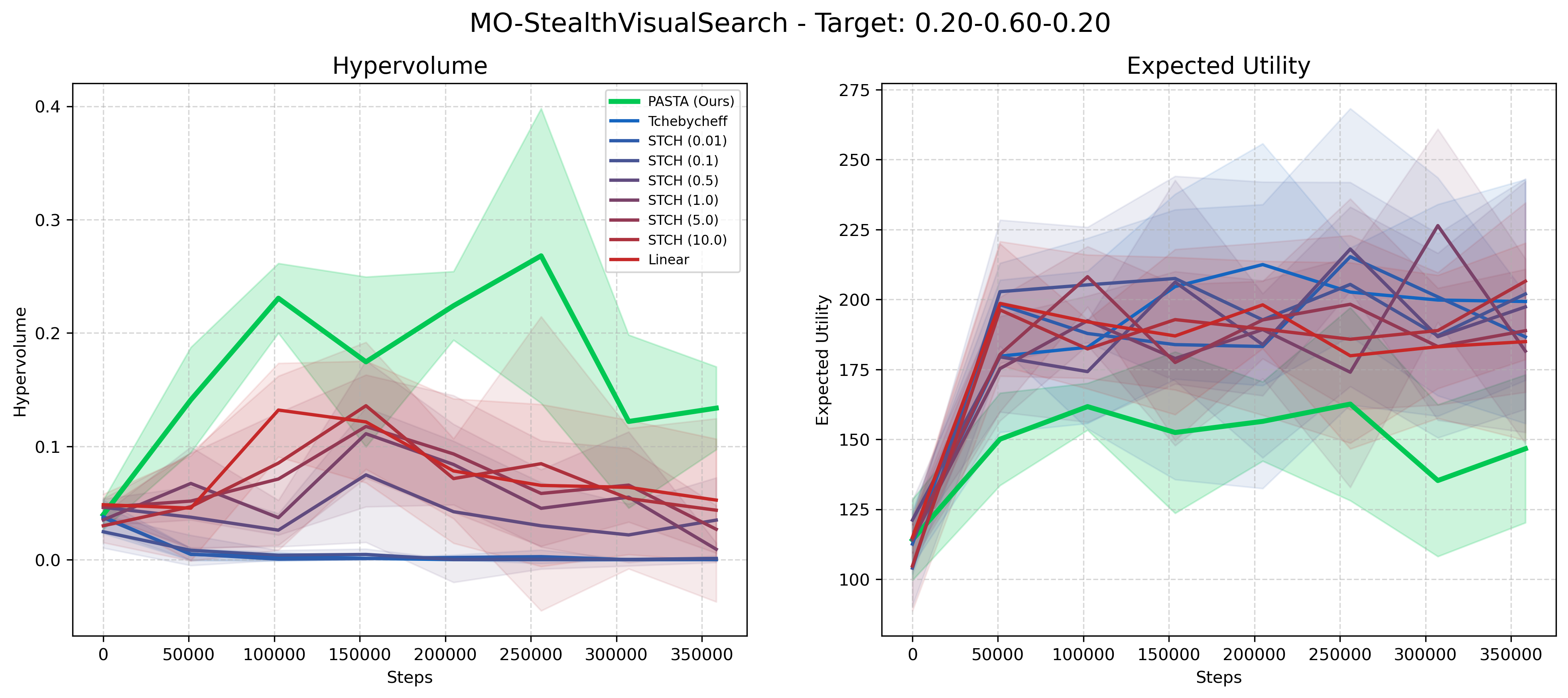}
        \caption{\small 0.20-0.60-0.20}
        \label{fig:bottom-left}
    \end{subfigure}
    \hfill
    \begin{subfigure}[b]{0.485\textwidth}
        \centering 
        \includegraphics[width=\linewidth]{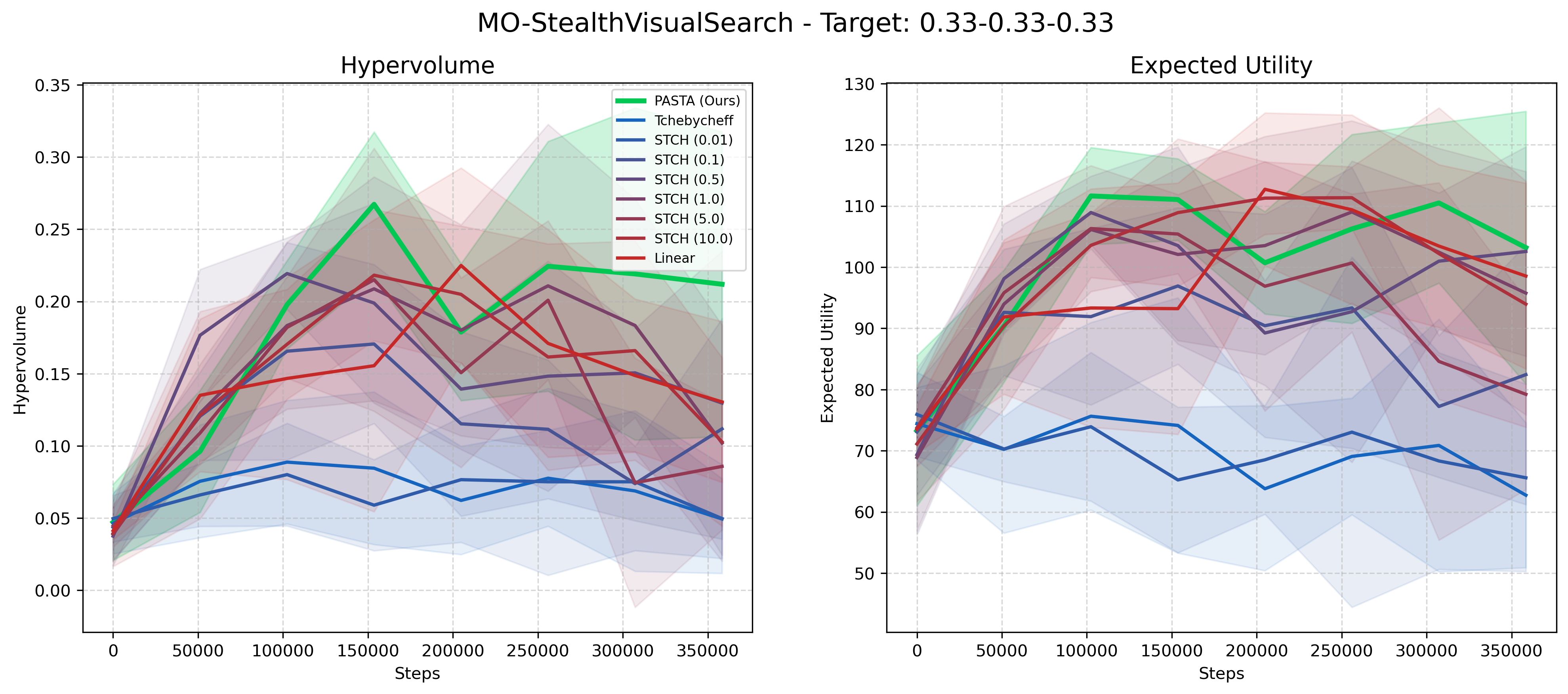} 
        \caption{\small 0.33-0.33-0.33}
        \label{fig:bottom-right}
    \end{subfigure}
    
    \vspace{0.2cm} % Adjust vertical space between rows

    % --- Third Row ---
    \begin{subfigure}[b]{0.485\textwidth}
        \centering
        \includegraphics[width=\linewidth]{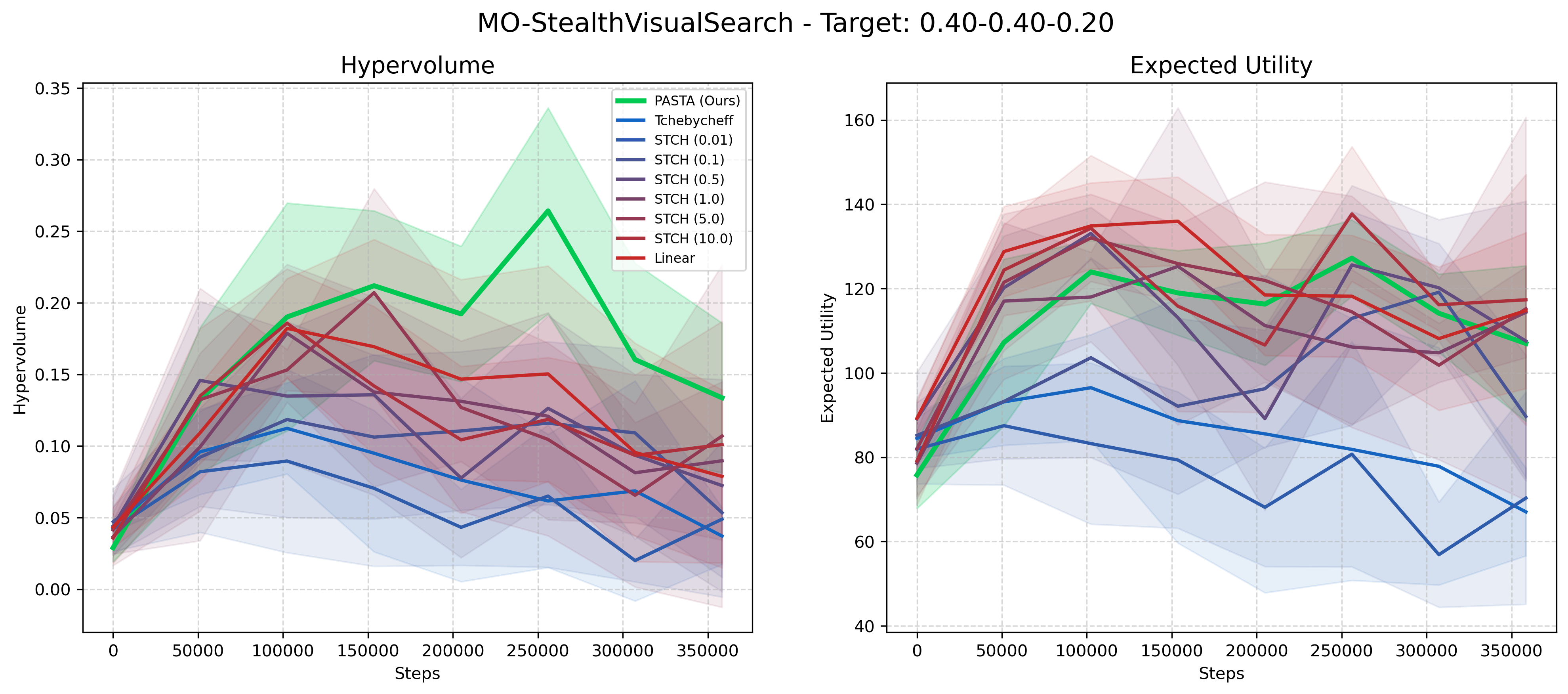} 
        \caption{\small 0.40-0.40-0.20}
        \label{fig:bottom-left}
    \end{subfigure}
    \hfill
    \begin{subfigure}[b]{0.485\textwidth}
        \centering
        \includegraphics[width=\linewidth]{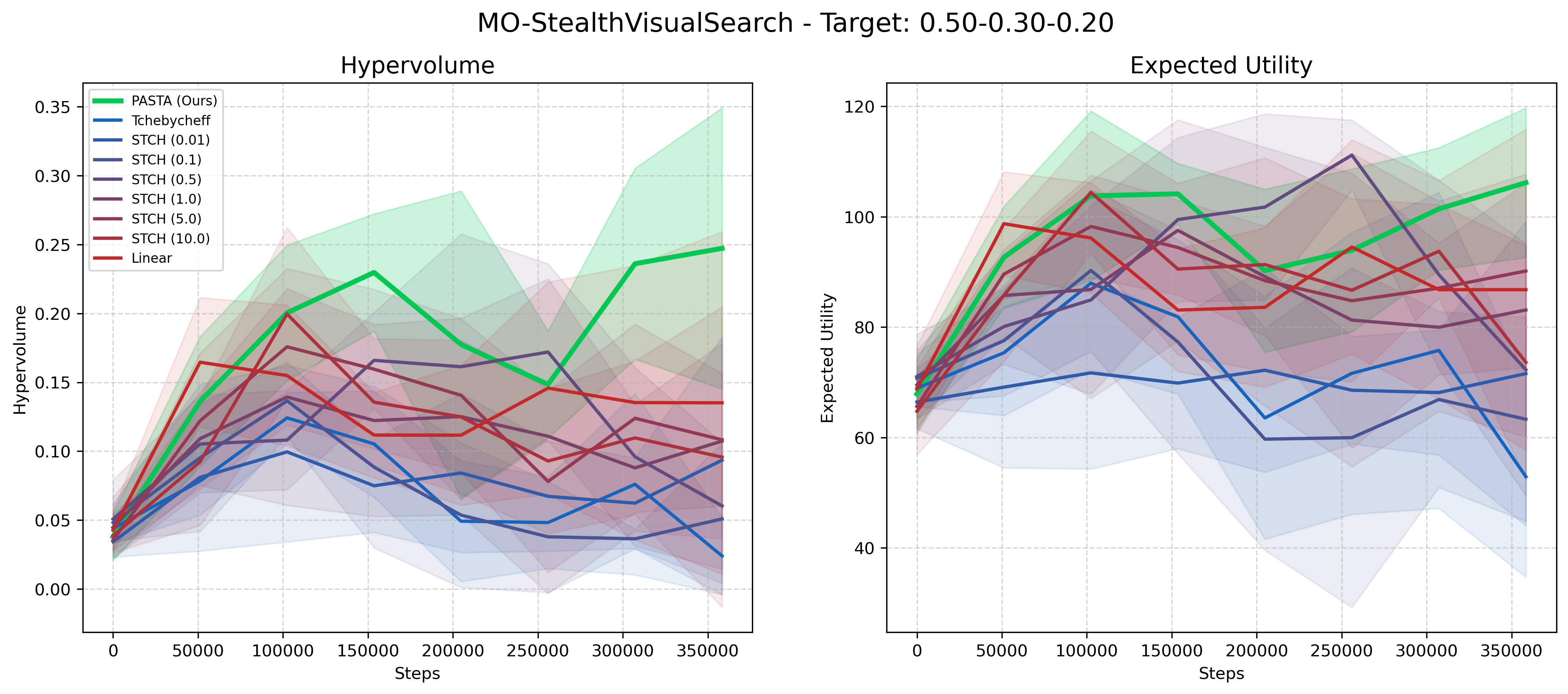} 
        \caption{\small 0.50-0.30-0.20}
        \label{fig:bottom-right}
    \end{subfigure}
    
    \vspace{0.2cm} % Adjust vertical space between rows

    % --- Fourth Row ---
    \begin{subfigure}[b]{0.485\textwidth}
        \centering
        \includegraphics[width=\linewidth]{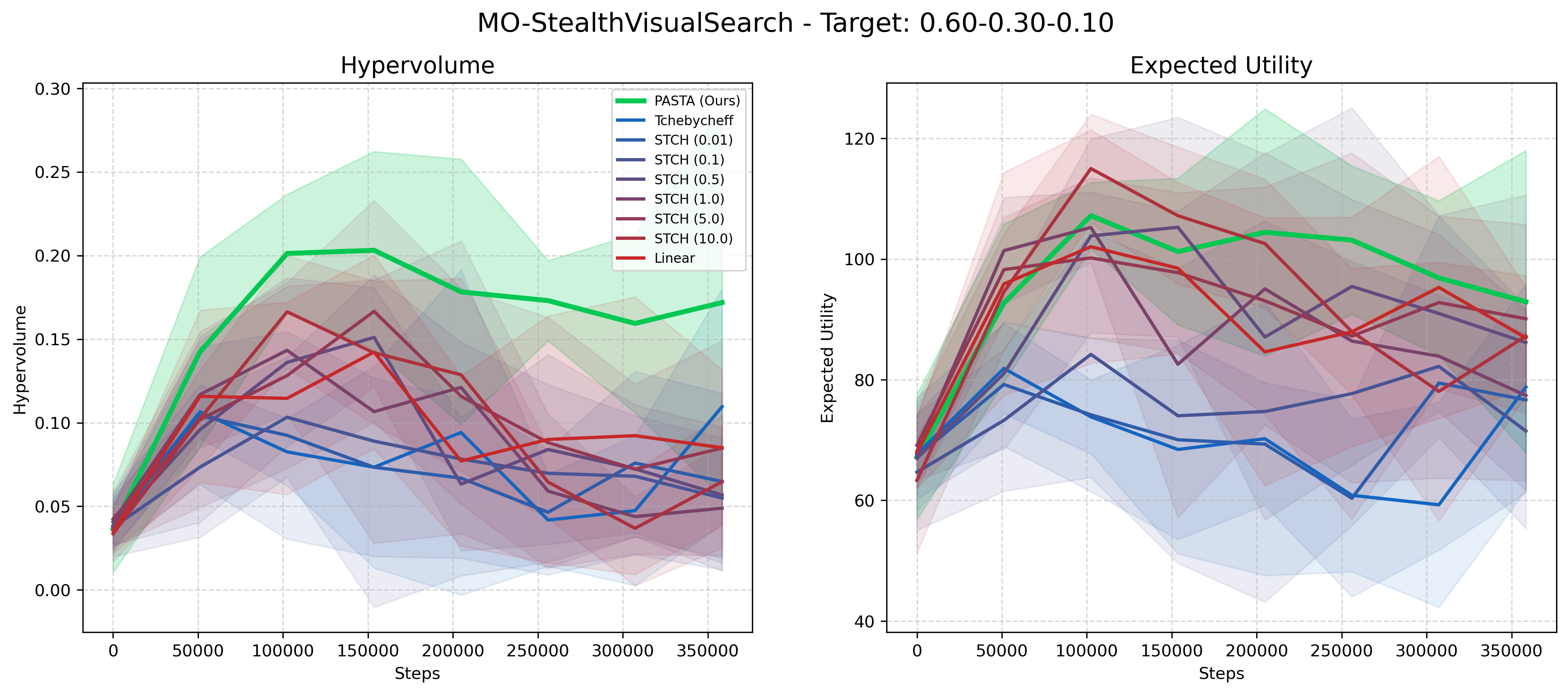} 
        \caption{\small 0.60-0.30-0.10}
        \label{fig:bottom-left}
    \end{subfigure}
    \hfill
    \begin{subfigure}[b]{0.485\textwidth}
        \centering
        \includegraphics[width=\linewidth]{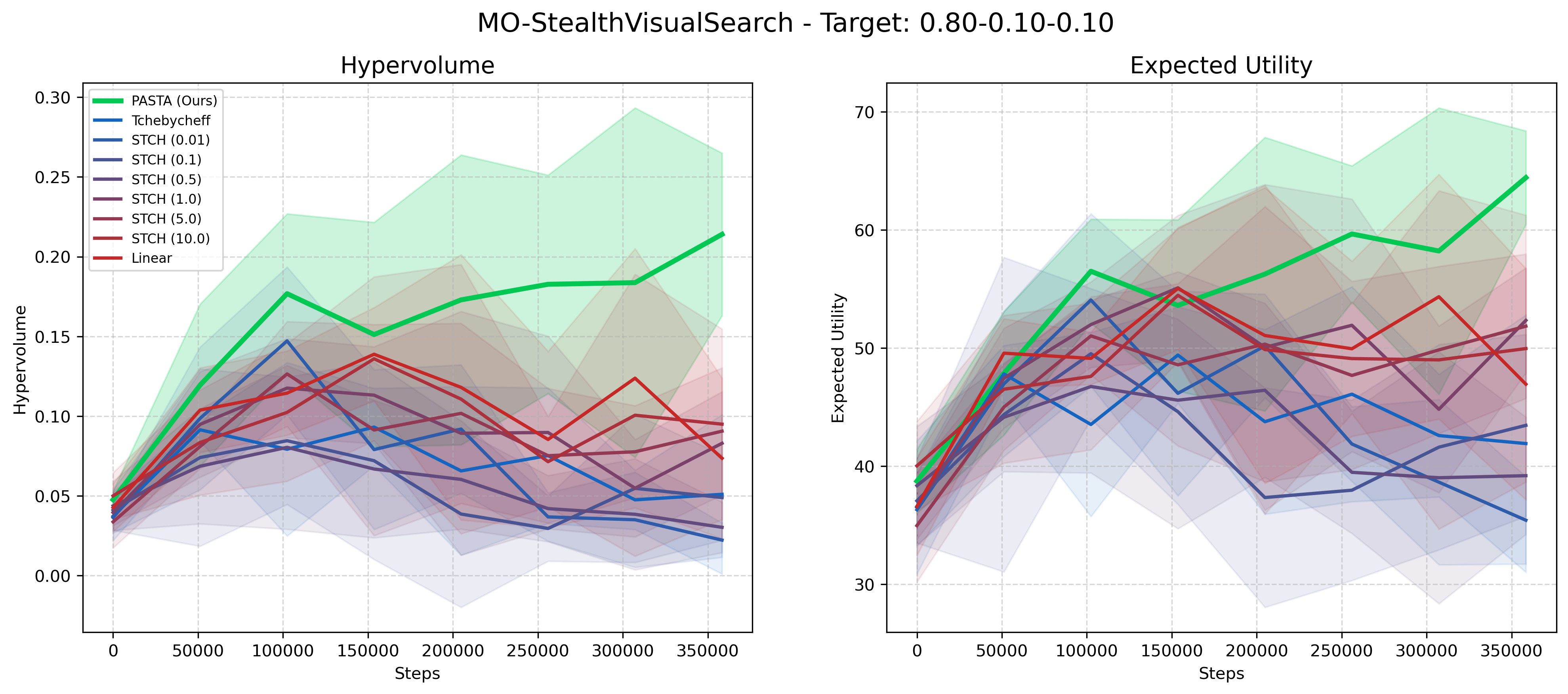} 
        \caption{\small 0.80-0.10-0.10}
        \label{fig:bottom-right}
    \end{subfigure}
    
    \caption{\small Training statistics evaluated on validation episodes in the Stealth Visual Search Environment. Each subplot shows the results for a given preference, as follows: \textit{(Left)} Evolution of the Hypervolume indicator, serving as a proxy for the quality and diversity of the approximated Pareto front. \textit{(Right)} The Expected Utility metric, measuring the alignment of the learned policies with specific preference weights. The solid lines represent the mean performance averaged over five seeds, while the shaded regions denote the standard deviation.}
    \label{fig:stealth_training_performance_metrics}
\end{figure*}

Table \ref{table:stealth-preference-results} presents the breakdown of the results for each preference across five seeds. Additionally, in Fig. \ref{fig:stealth_training_performance_metrics} we present the training progress measured over validation episodes, with the Hypervolume indicator shown on the left and Expected Utility on the right of each subplot. Shaded regions represent the standard deviation across five seeds.

\begin{figure*}[h]
    \centering
    \includegraphics[width=\linewidth]{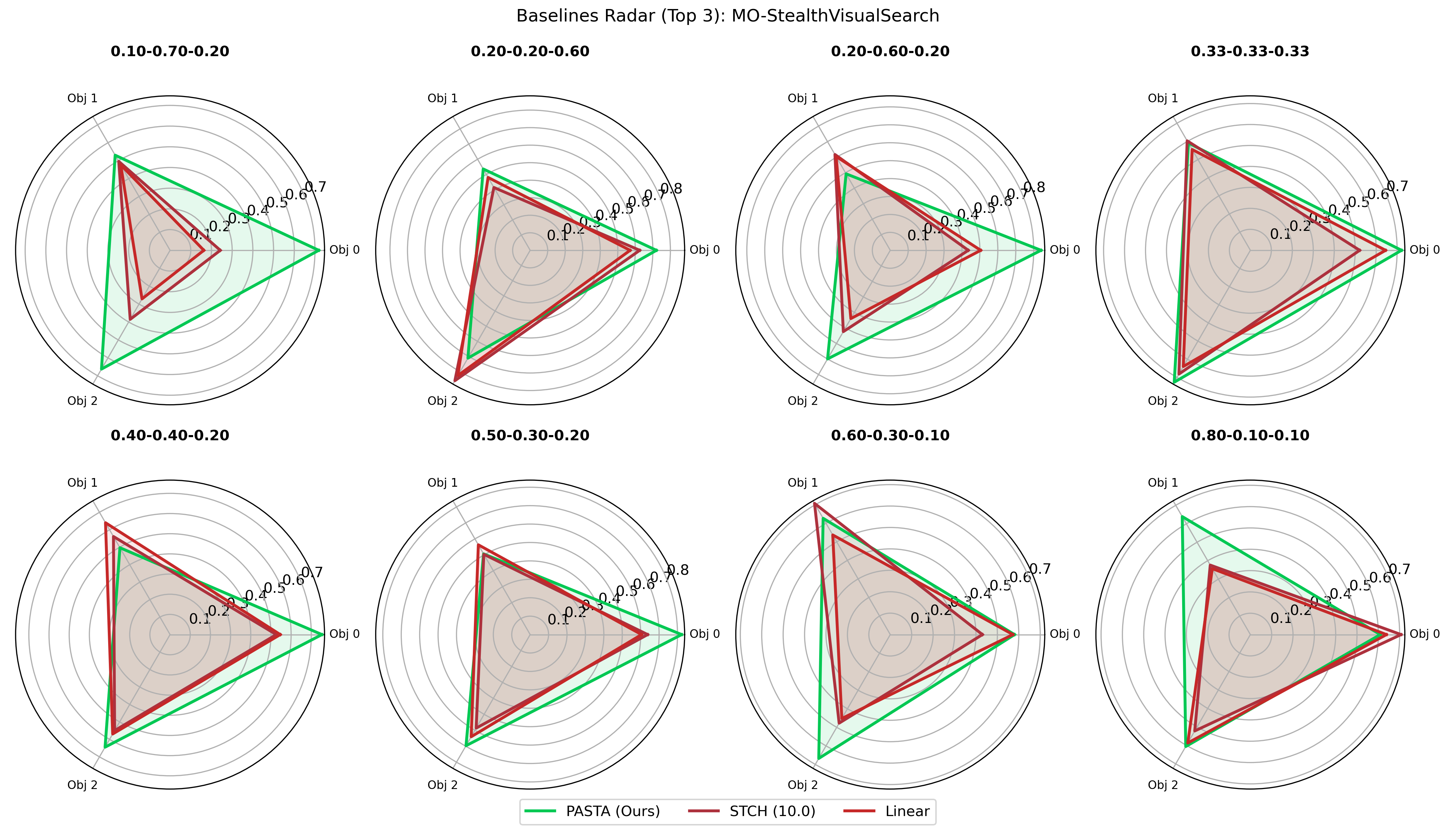}
    \caption{\small Normalized performance (utility) across 8 diverse preference vectors. To reduce visual clutter, we show PASTA and the two baselines with better performance (in terms of hypervolume).} 
    \label{fig:radar_charts}
\end{figure*}

Fig. \ref{fig:radar_charts} presents a radar chart comparison of our method against the top two performing methods (in terms of hypervolume): \textbf{STCH ($\mu=10.0$)}, and \textbf{Linear Scalarization}. The axes represent the normalized scores for the key training objectives: \textit{Search}, \textit{Stealth}, and \textit{Exploration}. As illustrated by the coverage area, \textbf{PASTA} demonstrates a superior balance across most objectives, confirming its capability to optimize conflicting objectives simultaneously and achieve a better solution to the MOP.

% PACKAGES REQUIRED: \usepackage{booktabs}, \usepackage{xcolor}
\begin{table*}[h] 
\centering
\fontsize{6pt}{7pt}\selectfont
\caption{Results (Baselines): MO-StealthVisualSearch. Metrics computed over 5 seeds.}
\label{table:stealth-preference-results}
\begin{tabular}{lcccccc}
\toprule
Method & Hypervolume $\uparrow$ & E. Utility $\uparrow$ & Obj 0 $\uparrow$ & Obj 1 $\uparrow$ & Obj 2 $\uparrow$ \\
\midrule
\multicolumn{6}{c}{\textbf{Preferences: 0.10-0.70-0.20}} \\
\midrule
PASTA (Ours) & \textbf{0.280 $\pm$ 0.015} & 195.274 $\pm$ 5.948 & \textbf{49.537 $\pm$ 5.084} & 265.126 $\pm$ 8.716 & \textbf{23.660 $\pm$ 1.866} \\
Tchebycheff & \textcolor{gray}{0.033 $\pm$ 0.010} & \textcolor{gray}{118.249 $\pm$ 15.331} & 22.706 $\pm$ 2.470 & \textcolor{gray}{162.370 $\pm$ 21.555} & \textcolor{gray}{11.599 $\pm$ 0.745} \\
STCH (0.01) & 0.042 $\pm$ 0.029 & 123.720 $\pm$ 25.823 & 23.254 $\pm$ 7.034 & 170.048 $\pm$ 35.164 & 11.805 $\pm$ 2.661 \\
STCH (0.1) & 0.035 $\pm$ 0.018 & 124.448 $\pm$ 15.954 & 21.593 $\pm$ 4.572 & 171.374 $\pm$ 22.157 & 11.635 $\pm$ 1.573 \\
STCH (0.5) & 0.058 $\pm$ 0.016 & 158.465 $\pm$ 42.440 & 23.679 $\pm$ 2.959 & 219.121 $\pm$ 60.849 & 13.561 $\pm$ 1.014 \\
STCH (1.0) & 0.094 $\pm$ 0.061 & \underline{195.848 $\pm$ 27.135} & 23.361 $\pm$ 4.648 & \underline{271.894 $\pm$ 38.520} & 15.932 $\pm$ 6.930 \\
STCH (5.0) & 0.074 $\pm$ 0.051 & \textbf{216.418 $\pm$ 41.818} & 20.200 $\pm$ 6.953 & \textbf{302.628 $\pm$ 60.734} & 12.795 $\pm$ 4.612 \\
STCH (10.0) & \underline{0.099 $\pm$ 0.074} & 185.073 $\pm$ 52.085 & \underline{24.224 $\pm$ 7.096} & 256.161 $\pm$ 73.669 & \underline{16.691 $\pm$ 5.256} \\
Linear & 0.059 $\pm$ 0.023 & 180.621 $\pm$ 49.517 & \textcolor{gray}{19.992 $\pm$ 5.116} & 251.223 $\pm$ 70.549 & 13.828 $\pm$ 2.562 \\
\midrule
\multicolumn{6}{c}{\textbf{Preferences: 0.20-0.20-0.60}} \\
\midrule
PASTA (Ours) & \textbf{0.297 $\pm$ 0.024} & \underline{78.147 $\pm$ 3.245} & \textbf{49.639 $\pm$ 4.819} & \underline{266.512 $\pm$ 17.545} & \textcolor{gray}{24.861 $\pm$ 1.108} \\
Tchebycheff & 0.210 $\pm$ 0.022 & 68.012 $\pm$ 1.176 & 41.243 $\pm$ 5.834 & 215.385 $\pm$ 10.148 & 27.812 $\pm$ 1.438 \\
STCH (0.01) & \textcolor{gray}{0.196 $\pm$ 0.026} & \textcolor{gray}{66.251 $\pm$ 3.174} & 40.428 $\pm$ 2.790 & \textcolor{gray}{208.216 $\pm$ 11.763} & 27.538 $\pm$ 2.378 \\
STCH (0.1) & 0.240 $\pm$ 0.053 & 74.576 $\pm$ 3.282 & 37.369 $\pm$ 7.296 & 247.850 $\pm$ 16.566 & \textbf{29.220 $\pm$ 1.604} \\
STCH (0.5) & 0.211 $\pm$ 0.021 & 71.098 $\pm$ 1.645 & 37.236 $\pm$ 3.370 & 235.870 $\pm$ 9.646 & 27.461 $\pm$ 1.104 \\
STCH (1.0) & 0.253 $\pm$ 0.022 & \textbf{78.680 $\pm$ 2.480} & \textcolor{gray}{35.804 $\pm$ 4.253} & \textbf{270.305 $\pm$ 11.842} & \underline{29.098 $\pm$ 1.953} \\
STCH (5.0) & 0.261 $\pm$ 0.047 & 76.124 $\pm$ 5.082 & 40.252 $\pm$ 3.968 & 255.619 $\pm$ 18.042 & 28.250 $\pm$ 2.727 \\
STCH (10.0) & \underline{0.263 $\pm$ 0.038} & 73.075 $\pm$ 5.209 & \underline{44.578 $\pm$ 2.756} & 235.016 $\pm$ 25.163 & 28.594 $\pm$ 0.750 \\
Linear & 0.257 $\pm$ 0.047 & 75.400 $\pm$ 7.048 & 41.663 $\pm$ 5.541 & 252.095 $\pm$ 36.332 & 27.747 $\pm$ 0.820 \\
\midrule
\multicolumn{6}{c}{\textbf{Preferences: 0.20-0.60-0.20}} \\
\midrule
PASTA (Ours) & \textbf{0.315 $\pm$ 0.064} & 169.322 $\pm$ 21.980 & \textbf{56.074 $\pm$ 7.651} & 255.315 $\pm$ 36.800 & \textbf{24.589 $\pm$ 1.513} \\
Tchebycheff & 0.038 $\pm$ 0.015 & 114.665 $\pm$ 9.689 & 20.589 $\pm$ 6.240 & 179.946 $\pm$ 16.088 & 12.896 $\pm$ 0.879 \\
STCH (0.01) & 0.037 $\pm$ 0.011 & 112.568 $\pm$ 10.671 & 21.677 $\pm$ 3.781 & 176.316 $\pm$ 17.407 & 12.216 $\pm$ 1.085 \\
STCH (0.1) & \textcolor{gray}{0.025 $\pm$ 0.014} & \textcolor{gray}{103.971 $\pm$ 13.603} & \textcolor{gray}{16.741 $\pm$ 6.253} & \textcolor{gray}{164.014 $\pm$ 20.938} & \textcolor{gray}{11.071 $\pm$ 1.948} \\
STCH (0.5) & 0.087 $\pm$ 0.053 & \underline{196.160 $\pm$ 42.204} & 19.984 $\pm$ 3.619 & \textbf{315.346 $\pm$ 70.386} & 14.778 $\pm$ 5.029 \\
STCH (1.0) & 0.129 $\pm$ 0.040 & 189.700 $\pm$ 12.436 & 33.307 $\pm$ 7.079 & 300.122 $\pm$ 20.501 & 14.830 $\pm$ 6.474 \\
STCH (5.0) & 0.162 $\pm$ 0.019 & \textbf{198.629 $\pm$ 6.425} & 34.648 $\pm$ 5.041 & \underline{314.024 $\pm$ 10.787} & 16.425 $\pm$ 3.950 \\
STCH (10.0) & \underline{0.182 $\pm$ 0.089} & 183.409 $\pm$ 26.055 & 34.473 $\pm$ 12.731 & 287.466 $\pm$ 47.446 & \underline{20.174 $\pm$ 4.345} \\
Linear & 0.172 $\pm$ 0.030 & 181.728 $\pm$ 24.046 & \underline{38.159 $\pm$ 10.158} & 284.139 $\pm$ 43.196 & 18.062 $\pm$ 2.066 \\
\midrule
\multicolumn{6}{c}{\textbf{Preferences: 0.33-0.33-0.33}} \\
\midrule
PASTA (Ours) & \textbf{0.320 $\pm$ 0.023} & \textbf{118.530 $\pm$ 4.456} & \textbf{49.680 $\pm$ 5.938} & 280.646 $\pm$ 17.195 & \textbf{25.263 $\pm$ 2.799} \\
Tchebycheff & \textcolor{gray}{0.127 $\pm$ 0.028} & \textcolor{gray}{88.946 $\pm$ 11.371} & \textcolor{gray}{38.972 $\pm$ 6.686} & \textcolor{gray}{209.113 $\pm$ 32.419} & 18.752 $\pm$ 2.254 \\
STCH (0.01) & 0.130 $\pm$ 0.031 & 90.460 $\pm$ 12.079 & 42.429 $\pm$ 4.522 & 211.846 $\pm$ 36.549 & \textcolor{gray}{17.104 $\pm$ 2.473} \\
STCH (0.1) & 0.222 $\pm$ 0.037 & 105.425 $\pm$ 10.206 & 43.218 $\pm$ 3.866 & 249.842 $\pm$ 30.389 & 23.215 $\pm$ 2.642 \\
STCH (0.5) & 0.251 $\pm$ 0.032 & 113.993 $\pm$ 6.432 & 40.413 $\pm$ 4.171 & 276.947 $\pm$ 19.184 & \underline{24.619 $\pm$ 0.673} \\
STCH (1.0) & \underline{0.268 $\pm$ 0.079} & \underline{116.322 $\pm$ 9.803} & 45.493 $\pm$ 8.009 & \underline{280.958 $\pm$ 24.860} & 22.515 $\pm$ 2.253 \\
STCH (5.0) & 0.263 $\pm$ 0.055 & 109.944 $\pm$ 6.732 & \underline{47.111 $\pm$ 8.451} & 258.470 $\pm$ 22.327 & 24.250 $\pm$ 2.728 \\
STCH (10.0) & 0.244 $\pm$ 0.023 & 115.666 $\pm$ 3.226 & 39.056 $\pm$ 3.005 & \textbf{283.808 $\pm$ 11.678} & 24.135 $\pm$ 2.657 \\
Linear & 0.257 $\pm$ 0.046 & 113.414 $\pm$ 10.948 & 45.629 $\pm$ 6.364 & 271.521 $\pm$ 35.987 & 23.091 $\pm$ 2.708 \\
\midrule
\multicolumn{6}{c}{\textbf{Preferences: 0.40-0.40-0.20}} \\
\midrule
PASTA (Ours) & \textbf{0.275 $\pm$ 0.065} & 127.797 $\pm$ 10.253 & \textbf{51.372 $\pm$ 9.268} & 256.518 $\pm$ 25.545 & \textbf{23.205 $\pm$ 1.841} \\
Tchebycheff & 0.155 $\pm$ 0.055 & 106.622 $\pm$ 19.273 & 41.334 $\pm$ 7.951 & 215.099 $\pm$ 48.232 & 20.243 $\pm$ 2.588 \\
STCH (0.01) & \textcolor{gray}{0.131 $\pm$ 0.044} & \textcolor{gray}{100.883 $\pm$ 11.600} & \textcolor{gray}{38.022 $\pm$ 5.477} & \textcolor{gray}{204.364 $\pm$ 26.958} & 19.641 $\pm$ 3.835 \\
STCH (0.1) & 0.178 $\pm$ 0.035 & 125.230 $\pm$ 14.972 & 40.386 $\pm$ 4.171 & 263.261 $\pm$ 39.706 & \textcolor{gray}{18.857 $\pm$ 3.160} \\
STCH (0.5) & 0.205 $\pm$ 0.042 & 129.158 $\pm$ 14.846 & 40.237 $\pm$ 6.119 & 272.099 $\pm$ 41.565 & 21.119 $\pm$ 4.503 \\
STCH (1.0) & 0.202 $\pm$ 0.028 & 133.747 $\pm$ 13.926 & 42.963 $\pm$ 3.569 & 281.955 $\pm$ 35.580 & 18.901 $\pm$ 4.848 \\
STCH (5.0) & \underline{0.253 $\pm$ 0.039} & \underline{135.150 $\pm$ 10.721} & \underline{43.789 $\pm$ 8.146} & \underline{282.748 $\pm$ 33.706} & \underline{22.679 $\pm$ 1.638} \\
STCH (10.0) & 0.204 $\pm$ 0.038 & 129.118 $\pm$ 7.988 & 39.340 $\pm$ 3.820 & 273.045 $\pm$ 20.036 & 20.823 $\pm$ 2.209 \\
Linear & 0.231 $\pm$ 0.018 & \textbf{137.782 $\pm$ 5.908} & 40.377 $\pm$ 4.244 & \textbf{293.424 $\pm$ 18.849} & 21.310 $\pm$ 1.663 \\
\midrule
\multicolumn{6}{c}{\textbf{Preferences: 0.50-0.30-0.20}} \\
\midrule
PASTA (Ours) & \textbf{0.314 $\pm$ 0.046} & \underline{110.091 $\pm$ 8.617} & \textbf{55.086 $\pm$ 5.878} & 258.818 $\pm$ 31.954 & \textbf{24.517 $\pm$ 1.208} \\
Tchebycheff & \textcolor{gray}{0.145 $\pm$ 0.022} & 95.358 $\pm$ 13.489 & \textcolor{gray}{38.321 $\pm$ 9.785} & 241.311 $\pm$ 56.614 & 19.022 $\pm$ 2.232 \\
STCH (0.01) & 0.165 $\pm$ 0.044 & \textcolor{gray}{91.740 $\pm$ 10.541} & \underline{47.594 $\pm$ 3.477} & \textcolor{gray}{213.869 $\pm$ 34.748} & 18.912 $\pm$ 2.548 \\
STCH (0.1) & 0.151 $\pm$ 0.025 & 95.374 $\pm$ 10.700 & 40.041 $\pm$ 7.375 & 238.792 $\pm$ 35.613 & \textcolor{gray}{18.579 $\pm$ 3.631} \\
STCH (0.5) & 0.225 $\pm$ 0.053 & \textbf{111.854 $\pm$ 6.530} & 43.984 $\pm$ 8.413 & \textbf{286.564 $\pm$ 24.677} & 19.467 $\pm$ 2.238 \\
STCH (1.0) & 0.220 $\pm$ 0.058 & 108.654 $\pm$ 6.839 & 44.559 $\pm$ 6.272 & \underline{274.778 $\pm$ 19.875} & 19.706 $\pm$ 3.325 \\
STCH (5.0) & 0.189 $\pm$ 0.052 & 99.044 $\pm$ 10.436 & 40.554 $\pm$ 8.373 & 248.240 $\pm$ 40.657 & 21.473 $\pm$ 2.966 \\
STCH (10.0) & 0.227 $\pm$ 0.047 & 104.396 $\pm$ 7.843 & 45.247 $\pm$ 6.944 & 258.078 $\pm$ 30.894 & 21.746 $\pm$ 2.620 \\
Linear & \underline{0.246 $\pm$ 0.044} & 108.403 $\pm$ 7.811 & 43.499 $\pm$ 8.426 & 273.439 $\pm$ 34.265 & \underline{23.105 $\pm$ 2.168} \\
\midrule
\multicolumn{6}{c}{\textbf{Preferences: 0.60-0.30-0.10}} \\
\midrule
PASTA (Ours) & \textbf{0.266 $\pm$ 0.043} & \underline{114.676 $\pm$ 5.266} & 42.179 $\pm$ 4.961 & \underline{289.966 $\pm$ 18.784} & \textbf{23.787 $\pm$ 3.410} \\
Tchebycheff & 0.157 $\pm$ 0.075 & \textcolor{gray}{90.136 $\pm$ 13.506} & \underline{43.115 $\pm$ 11.111} & \textcolor{gray}{207.584 $\pm$ 33.617} & \underline{19.913 $\pm$ 2.362} \\
STCH (0.01) & \textcolor{gray}{0.141 $\pm$ 0.042} & 91.828 $\pm$ 13.907 & 38.829 $\pm$ 5.717 & 222.041 $\pm$ 39.750 & 19.184 $\pm$ 4.994 \\
STCH (0.1) & 0.156 $\pm$ 0.064 & 96.750 $\pm$ 15.014 & \textbf{46.642 $\pm$ 8.945} & 223.439 $\pm$ 51.029 & 17.332 $\pm$ 2.630 \\
STCH (0.5) & 0.171 $\pm$ 0.026 & 110.291 $\pm$ 12.320 & 38.749 $\pm$ 3.061 & 284.430 $\pm$ 41.731 & 17.131 $\pm$ 2.014 \\
STCH (1.0) & 0.177 $\pm$ 0.067 & 112.732 $\pm$ 8.444 & 41.166 $\pm$ 11.459 & 288.000 $\pm$ 40.072 & \textcolor{gray}{16.330 $\pm$ 3.887} \\
STCH (5.0) & \underline{0.190 $\pm$ 0.041} & 108.705 $\pm$ 8.258 & 40.994 $\pm$ 11.632 & 273.971 $\pm$ 39.044 & 19.172 $\pm$ 1.764 \\
STCH (10.0) & 0.186 $\pm$ 0.033 & \textbf{115.650 $\pm$ 7.418} & \textcolor{gray}{34.210 $\pm$ 4.885} & \textbf{310.745 $\pm$ 29.634} & 19.009 $\pm$ 1.977 \\
Linear & 0.181 $\pm$ 0.031 & 106.954 $\pm$ 8.981 & 41.790 $\pm$ 7.808 & 266.806 $\pm$ 34.406 & 18.384 $\pm$ 1.928 \\
\midrule
\multicolumn{6}{c}{\textbf{Preferences: 0.80-0.10-0.10}} \\
\midrule
PASTA (Ours) & \textbf{0.260 $\pm$ 0.047} & \textbf{66.517 $\pm$ 5.864} & 43.731 $\pm$ 7.953 & \textbf{293.086 $\pm$ 38.983} & \textbf{22.240 $\pm$ 0.426} \\
Tchebycheff & \textcolor{gray}{0.112 $\pm$ 0.015} & \textcolor{gray}{49.715 $\pm$ 3.563} & \textcolor{gray}{34.751 $\pm$ 9.264} & 198.069 $\pm$ 49.479 & 21.067 $\pm$ 1.441 \\
STCH (0.01) & 0.159 $\pm$ 0.029 & 55.951 $\pm$ 4.639 & 41.425 $\pm$ 5.415 & 206.102 $\pm$ 14.101 & \underline{22.005 $\pm$ 1.500} \\
STCH (0.1) & 0.118 $\pm$ 0.039 & 51.352 $\pm$ 5.498 & 37.200 $\pm$ 4.427 & \textcolor{gray}{196.838 $\pm$ 27.809} & 19.080 $\pm$ 3.107 \\
STCH (0.5) & 0.114 $\pm$ 0.029 & 53.784 $\pm$ 4.581 & 37.969 $\pm$ 6.628 & 217.561 $\pm$ 39.083 & \textcolor{gray}{16.531 $\pm$ 3.817} \\
STCH (1.0) & 0.149 $\pm$ 0.044 & 58.601 $\pm$ 8.234 & 42.720 $\pm$ 8.552 & \underline{225.517 $\pm$ 64.010} & 18.729 $\pm$ 3.641 \\
STCH (5.0) & 0.145 $\pm$ 0.043 & 55.085 $\pm$ 4.813 & 38.872 $\pm$ 2.712 & 220.629 $\pm$ 27.344 & 19.244 $\pm$ 2.068 \\
STCH (10.0) & \underline{0.194 $\pm$ 0.041} & \underline{63.673 $\pm$ 3.687} & \textbf{48.917 $\pm$ 3.703} & 225.269 $\pm$ 12.486 & 20.125 $\pm$ 2.658 \\
Linear & 0.188 $\pm$ 0.046 & 60.418 $\pm$ 3.653 & \underline{45.264 $\pm$ 3.604} & 220.304 $\pm$ 22.441 & 21.760 $\pm$ 2.981 \\
\bottomrule
\end{tabular}
\end{table*}

\subsection{Extended Ablation Studies} \label{sec:appendix-ablations}

\subsubsection{PC Grad} \label{sec:appendix-ablation-pcgrad}
% PACKAGES REQUIRED: \usepackage{booktabs}, \usepackage{xcolor}
\begin{table*}[h]
\centering
\fontsize{6pt}{7pt}\selectfont
\caption{Results (Ablations PCGrad): MO-StealthVisualSearch. Metrics computed over 5 seeds.} \label{tab:ablations-pcgrad}
\begin{tabular}{lcccccc}
\toprule
Method & Hypervolume $\uparrow$ & E. Utility $\uparrow$ & Obj 0 $\uparrow$ & Obj 1 $\uparrow$ & Obj 2 $\uparrow$ \\
\midrule
\multicolumn{6}{c}{\textbf{Preferences: 0.10-0.70-0.20}} \\
\midrule
PASTA (Ours) & \textbf{0.280 $\pm$ 0.015} & \underline{195.274 $\pm$ 5.948} & \textbf{49.537 $\pm$ 5.084} & \underline{265.126 $\pm$ 8.716} & \textbf{23.660 $\pm$ 1.866} \\
No PCGrad & \textcolor{gray}{0.070 $\pm$ 0.040} & \textcolor{gray}{171.787 $\pm$ 50.599} & \textcolor{gray}{23.252 $\pm$ 7.206} & \textcolor{gray}{237.891 $\pm$ 72.844} & \textcolor{gray}{14.692 $\pm$ 4.109} \\
Weighted PCGrad & \underline{0.254 $\pm$ 0.044} & \textbf{201.893 $\pm$ 18.281} & \underline{43.750 $\pm$ 5.220} & \textbf{275.533 $\pm$ 26.280} & \underline{23.226 $\pm$ 1.803} \\
\midrule
\multicolumn{6}{c}{\textbf{Preferences: 0.20-0.20-0.60}} \\
\midrule
PASTA (Ours) & \textbf{0.297 $\pm$ 0.024} & \textbf{78.147 $\pm$ 3.245} & \textbf{49.639 $\pm$ 4.819} & \underline{266.512 $\pm$ 17.545} & \underline{24.861 $\pm$ 1.108} \\
No PCGrad & \underline{0.267 $\pm$ 0.036} & \underline{77.647 $\pm$ 1.669} & \textcolor{gray}{40.829 $\pm$ 8.198} & \textcolor{gray}{263.517 $\pm$ 16.567} & \textbf{27.963 $\pm$ 1.358} \\
Weighted PCGrad & \textcolor{gray}{0.265 $\pm$ 0.030} & \textcolor{gray}{76.777 $\pm$ 6.758} & \underline{49.362 $\pm$ 6.559} & \textbf{266.780 $\pm$ 41.161} & \textcolor{gray}{22.581 $\pm$ 1.728} \\
\midrule
\multicolumn{6}{c}{\textbf{Preferences: 0.20-0.60-0.20}} \\
\midrule
PASTA (Ours) & \textbf{0.315 $\pm$ 0.064} & \textcolor{gray}{169.322 $\pm$ 21.980} & \textbf{56.074 $\pm$ 7.651} & \textcolor{gray}{255.315 $\pm$ 36.800} & \textbf{24.589 $\pm$ 1.513} \\
No PCGrad & \textcolor{gray}{0.165 $\pm$ 0.042} & \textbf{184.927 $\pm$ 12.939} & \textcolor{gray}{31.503 $\pm$ 7.197} & \textbf{291.101 $\pm$ 24.050} & \textcolor{gray}{19.828 $\pm$ 3.055} \\
Weighted PCGrad & \underline{0.246 $\pm$ 0.025} & \underline{184.407 $\pm$ 10.234} & \underline{40.479 $\pm$ 3.694} & \underline{286.094 $\pm$ 17.643} & \underline{23.273 $\pm$ 2.202} \\
\midrule
\multicolumn{6}{c}{\textbf{Preferences: 0.33-0.33-0.33}} \\
\midrule
PASTA (Ours) & \textbf{0.320 $\pm$ 0.023} & \textbf{118.530 $\pm$ 4.456} & \textbf{49.680 $\pm$ 5.938} & \textbf{280.646 $\pm$ 17.195} & \textbf{25.263 $\pm$ 2.799} \\
No PCGrad & \textcolor{gray}{0.258 $\pm$ 0.062} & \textcolor{gray}{108.499 $\pm$ 11.431} & \textcolor{gray}{47.170 $\pm$ 5.371} & \textcolor{gray}{254.190 $\pm$ 35.072} & \textcolor{gray}{24.138 $\pm$ 3.685} \\
Weighted PCGrad & \underline{0.304 $\pm$ 0.039} & \underline{115.715 $\pm$ 8.485} & \underline{49.340 $\pm$ 8.369} & \underline{272.704 $\pm$ 27.233} & \underline{25.100 $\pm$ 2.546} \\
\midrule
\multicolumn{6}{c}{\textbf{Preferences: 0.40-0.40-0.20}} \\
\midrule
PASTA (Ours) & \textbf{0.275 $\pm$ 0.065} & \textcolor{gray}{127.797 $\pm$ 10.253} & \textbf{51.372 $\pm$ 9.268} & \textcolor{gray}{256.518 $\pm$ 25.545} & \textbf{23.205 $\pm$ 1.841} \\
No PCGrad & \textcolor{gray}{0.225 $\pm$ 0.036} & \textbf{139.020 $\pm$ 9.296} & \textcolor{gray}{41.426 $\pm$ 5.720} & \textbf{296.154 $\pm$ 27.047} & \textcolor{gray}{19.939 $\pm$ 1.610} \\
Weighted PCGrad & \underline{0.264 $\pm$ 0.062} & \underline{137.022 $\pm$ 4.090} & \underline{45.392 $\pm$ 6.320} & \underline{286.150 $\pm$ 10.230} & \underline{22.027 $\pm$ 3.599} \\
\midrule
\multicolumn{6}{c}{\textbf{Preferences: 0.50-0.30-0.20}} \\
\midrule
PASTA (Ours) & \textbf{0.314 $\pm$ 0.046} & \underline{110.091 $\pm$ 8.617} & \textbf{55.086 $\pm$ 5.878} & \textcolor{gray}{258.818 $\pm$ 31.954} & \textbf{24.517 $\pm$ 1.208} \\
No PCGrad & \textcolor{gray}{0.201 $\pm$ 0.057} & \textcolor{gray}{106.612 $\pm$ 8.925} & \textcolor{gray}{43.556 $\pm$ 8.864} & \underline{269.947 $\pm$ 41.332} & \textcolor{gray}{19.248 $\pm$ 3.681} \\
Weighted PCGrad & \underline{0.283 $\pm$ 0.062} & \textbf{111.340 $\pm$ 8.351} & \underline{47.600 $\pm$ 3.581} & \textbf{276.203 $\pm$ 26.768} & \underline{23.396 $\pm$ 2.478} \\
\midrule
\multicolumn{6}{c}{\textbf{Preferences: 0.60-0.30-0.10}} \\
\midrule
PASTA (Ours) & \underline{0.266 $\pm$ 0.043} & \textbf{114.676 $\pm$ 5.266} & \underline{42.179 $\pm$ 4.961} & \textbf{289.966 $\pm$ 18.784} & \underline{23.787 $\pm$ 3.410} \\
No PCGrad & \textcolor{gray}{0.159 $\pm$ 0.018} & \textcolor{gray}{103.790 $\pm$ 4.699} & \textcolor{gray}{37.599 $\pm$ 3.492} & \textcolor{gray}{264.816 $\pm$ 21.333} & \textcolor{gray}{17.857 $\pm$ 1.245} \\
Weighted PCGrad & \textbf{0.285 $\pm$ 0.042} & \underline{113.418 $\pm$ 4.991} & \textbf{46.630 $\pm$ 7.210} & \underline{276.693 $\pm$ 10.701} & \textbf{24.322 $\pm$ 2.410} \\
\midrule
\multicolumn{6}{c}{\textbf{Preferences: 0.80-0.10-0.10}} \\
\midrule
PASTA (Ours) & \textbf{0.260 $\pm$ 0.047} & \textbf{66.517 $\pm$ 5.864} & \underline{43.731 $\pm$ 7.953} & \textbf{293.086 $\pm$ 38.983} & \underline{22.240 $\pm$ 0.426} \\
No PCGrad & \textcolor{gray}{0.150 $\pm$ 0.053} & \textcolor{gray}{56.598 $\pm$ 5.478} & \textcolor{gray}{41.266 $\pm$ 6.595} & \textcolor{gray}{216.507 $\pm$ 28.883} & \textcolor{gray}{19.342 $\pm$ 2.372} \\
Weighted PCGrad & \underline{0.259 $\pm$ 0.066} & \underline{65.862 $\pm$ 5.932} & \textbf{46.621 $\pm$ 5.674} & \underline{262.110 $\pm$ 34.798} & \textbf{23.541 $\pm$ 3.570} \\
\bottomrule
\end{tabular}
\end{table*}

In Table \ref{tab:ablations-pcgrad}, we evaluate the impact of gradient projection on the learning process.
We compare the standard PASTA implementation against a baseline with no gradient projection (``No PCGrad'') and a variation where the projected gradients are weighted by the ASTCH attention mechanism (``Weighted PCGrad'').
The results confirm that standard PCGrad is crucial for preventing performance collapse in conflicting objectives.

\subsubsection{Smoothness Controller} \label{sec:appendix-ablation-controller}
% PACKAGES REQUIRED: \usepackage{booktabs}, \usepackage{xcolor}
\begin{table*}[h]
\centering
\fontsize{6pt}{7pt}\selectfont
\caption{Results (Ablations Controller): MO-StealthVisualSearch. Metrics computed over 5 seeds.} \label{tab:ablations-controller}
\begin{tabular}{lcccccc}
\toprule
Method & Hypervolume $\uparrow$ & E. Utility $\uparrow$ & Obj 0 $\uparrow$ & Obj 1 $\uparrow$ & Obj 2 $\uparrow$ \\
\midrule
\multicolumn{6}{c}{\textbf{Preferences: 0.10-0.70-0.20}} \\
\midrule
PASTA (Ours) & \textbf{0.280 $\pm$ 0.015} & \underline{195.274 $\pm$ 5.948} & \underline{49.537 $\pm$ 5.084} & \underline{265.126 $\pm$ 8.716} & \textbf{23.660 $\pm$ 1.866} \\
Controller No Conflict & \underline{0.270 $\pm$ 0.065} & \textcolor{gray}{189.204 $\pm$ 11.247} & \textbf{49.756 $\pm$ 6.527} & \textcolor{gray}{256.511 $\pm$ 16.071} & \underline{23.351 $\pm$ 3.056} \\
Controller No Decay & \textcolor{gray}{0.238 $\pm$ 0.026} & \textbf{197.590 $\pm$ 9.272} & \textcolor{gray}{42.790 $\pm$ 3.989} & \textbf{269.602 $\pm$ 13.719} & \textcolor{gray}{22.944 $\pm$ 2.674} \\
\midrule
\multicolumn{6}{c}{\textbf{Preferences: 0.20-0.20-0.60}} \\
\midrule
PASTA (Ours) & \textcolor{gray}{0.297 $\pm$ 0.024} & \textcolor{gray}{78.147 $\pm$ 3.245} & \underline{49.639 $\pm$ 4.819} & \textcolor{gray}{266.512 $\pm$ 17.545} & \underline{24.861 $\pm$ 1.108} \\
Controller No Conflict & \textbf{0.319 $\pm$ 0.027} & \textbf{82.259 $\pm$ 4.200} & \textbf{50.992 $\pm$ 6.776} & \underline{288.808 $\pm$ 31.088} & \textcolor{gray}{23.831 $\pm$ 2.173} \\
Controller No Decay & \underline{0.298 $\pm$ 0.047} & \underline{81.970 $\pm$ 2.766} & \textcolor{gray}{44.592 $\pm$ 7.844} & \textbf{289.202 $\pm$ 18.633} & \textbf{25.353 $\pm$ 2.676} \\
\midrule
\multicolumn{6}{c}{\textbf{Preferences: 0.20-0.60-0.20}} \\
\midrule
PASTA (Ours) & \textbf{0.315 $\pm$ 0.064} & \textcolor{gray}{169.322 $\pm$ 21.980} & \textbf{56.074 $\pm$ 7.651} & \textcolor{gray}{255.315 $\pm$ 36.800} & \textbf{24.589 $\pm$ 1.513} \\
Controller No Conflict & \textcolor{gray}{0.274 $\pm$ 0.075} & \underline{174.383 $\pm$ 7.383} & \textcolor{gray}{47.561 $\pm$ 12.854} & \underline{266.844 $\pm$ 9.281} & \textcolor{gray}{23.821 $\pm$ 1.051} \\
Controller No Decay & \underline{0.283 $\pm$ 0.019} & \textbf{177.142 $\pm$ 11.887} & \underline{48.333 $\pm$ 7.545} & \textbf{271.079 $\pm$ 22.082} & \underline{24.139 $\pm$ 2.008} \\
\midrule
\multicolumn{6}{c}{\textbf{Preferences: 0.33-0.33-0.33}} \\
\midrule
PASTA (Ours) & \textbf{0.320 $\pm$ 0.023} & \textbf{118.530 $\pm$ 4.456} & \textbf{49.680 $\pm$ 5.938} & \textbf{280.646 $\pm$ 17.195} & \textbf{25.263 $\pm$ 2.799} \\
Controller No Conflict & \underline{0.281 $\pm$ 0.039} & \textcolor{gray}{111.710 $\pm$ 3.272} & \underline{49.518 $\pm$ 4.295} & \textcolor{gray}{261.494 $\pm$ 7.698} & \underline{24.119 $\pm$ 2.745} \\
Controller No Decay & \textcolor{gray}{0.257 $\pm$ 0.042} & \underline{112.129 $\pm$ 10.792} & \textcolor{gray}{47.572 $\pm$ 4.106} & \underline{266.336 $\pm$ 33.496} & \textcolor{gray}{22.479 $\pm$ 1.377} \\
\midrule
\multicolumn{6}{c}{\textbf{Preferences: 0.40-0.40-0.20}} \\
\midrule
PASTA (Ours) & \textcolor{gray}{0.275 $\pm$ 0.065} & \textcolor{gray}{127.797 $\pm$ 10.253} & \textbf{51.372 $\pm$ 9.268} & \textcolor{gray}{256.518 $\pm$ 25.545} & \textcolor{gray}{23.205 $\pm$ 1.841} \\
Controller No Conflict & \underline{0.288 $\pm$ 0.035} & \textbf{133.554 $\pm$ 6.614} & \underline{48.390 $\pm$ 5.482} & \textbf{273.490 $\pm$ 20.515} & \underline{24.009 $\pm$ 1.653} \\
Controller No Decay & \textbf{0.289 $\pm$ 0.042} & \underline{131.299 $\pm$ 6.243} & \textcolor{gray}{47.526 $\pm$ 10.101} & \underline{267.934 $\pm$ 23.386} & \textbf{25.575 $\pm$ 1.754} \\
\midrule
\multicolumn{6}{c}{\textbf{Preferences: 0.50-0.30-0.20}} \\
\midrule
PASTA (Ours) & \textbf{0.314 $\pm$ 0.046} & \textcolor{gray}{110.091 $\pm$ 8.617} & \textbf{55.086 $\pm$ 5.878} & \textcolor{gray}{258.818 $\pm$ 31.954} & \underline{24.517 $\pm$ 1.208} \\
Controller No Conflict & \underline{0.307 $\pm$ 0.032} & \textbf{112.785 $\pm$ 5.251} & \underline{50.570 $\pm$ 5.994} & \textbf{275.478 $\pm$ 14.374} & \textcolor{gray}{24.283 $\pm$ 2.481} \\
Controller No Decay & \textcolor{gray}{0.292 $\pm$ 0.044} & \underline{110.423 $\pm$ 6.249} & \textcolor{gray}{46.336 $\pm$ 1.332} & \underline{274.093 $\pm$ 18.836} & \textbf{25.137 $\pm$ 1.665} \\
\midrule
\multicolumn{6}{c}{\textbf{Preferences: 0.60-0.30-0.10}} \\
\midrule
PASTA (Ours) & \underline{0.266 $\pm$ 0.043} & \textbf{114.676 $\pm$ 5.266} & \textcolor{gray}{42.179 $\pm$ 4.961} & \textbf{289.966 $\pm$ 18.784} & \underline{23.787 $\pm$ 3.410} \\
Controller No Conflict & \textbf{0.288 $\pm$ 0.039} & \underline{110.839 $\pm$ 5.811} & \textbf{49.019 $\pm$ 8.538} & \underline{263.003 $\pm$ 35.151} & \textbf{25.267 $\pm$ 3.160} \\
Controller No Decay & \textcolor{gray}{0.251 $\pm$ 0.032} & \textcolor{gray}{106.071 $\pm$ 5.330} & \underline{48.919 $\pm$ 3.746} & \textcolor{gray}{247.986 $\pm$ 20.658} & \textcolor{gray}{23.236 $\pm$ 1.058} \\
\midrule
\multicolumn{6}{c}{\textbf{Preferences: 0.80-0.10-0.10}} \\
\midrule
PASTA (Ours) & \textcolor{gray}{0.260 $\pm$ 0.047} & \textcolor{gray}{66.517 $\pm$ 5.864} & \textcolor{gray}{43.731 $\pm$ 7.953} & \textbf{293.086 $\pm$ 38.983} & \textcolor{gray}{22.240 $\pm$ 0.426} \\
Controller No Conflict & \textbf{0.336 $\pm$ 0.047} & \textbf{73.437 $\pm$ 5.580} & \textbf{52.840 $\pm$ 7.360} & \underline{287.586 $\pm$ 14.437} & \textbf{24.064 $\pm$ 1.671} \\
Controller No Decay & \underline{0.272 $\pm$ 0.047} & \underline{67.744 $\pm$ 4.435} & \underline{48.404 $\pm$ 4.787} & \textcolor{gray}{267.032 $\pm$ 22.184} & \underline{23.175 $\pm$ 0.833} \\
\bottomrule
\end{tabular}
\end{table*}

Table \ref{tab:ablations-controller} dissects the Adaptive Smoothness Controller by isolating its two primary inputs: the conflict ratio and the time-step decay signal.
The ``No Conflict'' ablation removes the gradient conflict term, relying solely on the time decay, while ``No Decay'' removes the temporal component.
This comparison allows us to attribute the controller's effectiveness primarily to the annealing schedule, with the conflict signal providing auxiliary robustness.

\subsubsection{Critic Architecture} \label{sec:appendix-ablation-critic}
% PACKAGES REQUIRED: \usepackage{booktabs}, \usepackage{xcolor}
\begin{table*}[h]
\centering
\fontsize{6pt}{7pt}\selectfont
\caption{Results (Ablations Critic): MO-StealthVisualSearch. Metrics computed over 5 seeds.} \label{tab:ablations-critic}
\begin{tabular}{lcccccc}
\toprule
Method & Hypervolume $\uparrow$ & E. Utility $\uparrow$ & Obj 0 $\uparrow$ & Obj 1 $\uparrow$ & Obj 2 $\uparrow$ \\
\midrule
\multicolumn{6}{c}{\textbf{Preferences: 0.10-0.70-0.20}} \\
\midrule
PASTA (Ours) & \underline{0.280 $\pm$ 0.015} & 195.274 $\pm$ 5.948 & \textbf{49.537 $\pm$ 5.084} & 265.126 $\pm$ 8.716 & 23.660 $\pm$ 1.866 \\
Branched Unweighted & \textbf{0.282 $\pm$ 0.064} & \underline{199.818 $\pm$ 18.531} & \underline{46.752 $\pm$ 3.190} & \underline{271.858 $\pm$ 26.275} & \underline{24.209 $\pm$ 2.498} \\
Shared Unweighted & 0.264 $\pm$ 0.036 & \textcolor{gray}{192.786 $\pm$ 10.343} & 44.613 $\pm$ 4.144 & \textcolor{gray}{261.893 $\pm$ 14.502} & \textbf{25.000 $\pm$ 0.448} \\
Shared Weighted & \textcolor{gray}{0.247 $\pm$ 0.044} & \textbf{202.315 $\pm$ 9.721} & \textcolor{gray}{42.382 $\pm$ 4.943} & \textbf{276.342 $\pm$ 14.200} & \textcolor{gray}{23.189 $\pm$ 2.385} \\
\midrule
\multicolumn{6}{c}{\textbf{Preferences: 0.20-0.20-0.60}} \\
\midrule
PASTA (Ours) & \textbf{0.297 $\pm$ 0.024} & \textbf{78.147 $\pm$ 3.245} & \textbf{49.639 $\pm$ 4.819} & 266.512 $\pm$ 17.545 & \textbf{24.861 $\pm$ 1.108} \\
Branched Unweighted & 0.247 $\pm$ 0.033 & 76.981 $\pm$ 3.879 & 43.574 $\pm$ 9.958 & \underline{270.539 $\pm$ 24.042} & \underline{23.597 $\pm$ 1.385} \\
Shared Unweighted & \textcolor{gray}{0.241 $\pm$ 0.076} & \textcolor{gray}{75.648 $\pm$ 7.396} & \textcolor{gray}{42.880 $\pm$ 8.265} & \textcolor{gray}{266.194 $\pm$ 26.990} & \textcolor{gray}{23.055 $\pm$ 2.749} \\
Shared Weighted & \underline{0.267 $\pm$ 0.025} & \underline{77.559 $\pm$ 2.166} & \underline{46.585 $\pm$ 5.752} & \textbf{270.845 $\pm$ 11.737} & 23.456 $\pm$ 1.853 \\
\midrule
\multicolumn{6}{c}{\textbf{Preferences: 0.20-0.60-0.20}} \\
\midrule
PASTA (Ours) & \textbf{0.315 $\pm$ 0.064} & \textcolor{gray}{169.322 $\pm$ 21.980} & \textbf{56.074 $\pm$ 7.651} & \textcolor{gray}{255.315 $\pm$ 36.800} & \textbf{24.589 $\pm$ 1.513} \\
Branched Unweighted & 0.243 $\pm$ 0.053 & \underline{178.081 $\pm$ 20.883} & \textcolor{gray}{43.705 $\pm$ 3.731} & \textbf{274.860 $\pm$ 34.535} & 22.120 $\pm$ 1.997 \\
Shared Unweighted & \underline{0.272 $\pm$ 0.045} & \textbf{178.787 $\pm$ 8.415} & \underline{46.477 $\pm$ 4.418} & \underline{274.628 $\pm$ 15.158} & \underline{23.573 $\pm$ 4.040} \\
Shared Weighted & \textcolor{gray}{0.235 $\pm$ 0.046} & 170.776 $\pm$ 18.689 & 45.961 $\pm$ 5.693 & 262.017 $\pm$ 33.543 & \textcolor{gray}{21.869 $\pm$ 3.562} \\
\midrule
\multicolumn{6}{c}{\textbf{Preferences: 0.33-0.33-0.33}} \\
\midrule
PASTA (Ours) & \textbf{0.320 $\pm$ 0.023} & \textbf{118.530 $\pm$ 4.456} & \textbf{49.680 $\pm$ 5.938} & \underline{280.646 $\pm$ 17.195} & \textbf{25.263 $\pm$ 2.799} \\
Branched Unweighted & 0.246 $\pm$ 0.047 & 115.131 $\pm$ 6.476 & 43.015 $\pm$ 5.318 & 280.151 $\pm$ 17.244 & \textcolor{gray}{22.226 $\pm$ 1.703} \\
Shared Unweighted & \textcolor{gray}{0.229 $\pm$ 0.065} & \textcolor{gray}{110.082 $\pm$ 12.126} & \textcolor{gray}{40.361 $\pm$ 4.544} & \textcolor{gray}{266.774 $\pm$ 32.338} & 23.112 $\pm$ 2.291 \\
Shared Weighted & \underline{0.263 $\pm$ 0.042} & \underline{116.234 $\pm$ 11.650} & \underline{44.455 $\pm$ 3.727} & \textbf{281.052 $\pm$ 37.600} & \underline{23.196 $\pm$ 2.920} \\
\midrule
\multicolumn{6}{c}{\textbf{Preferences: 0.40-0.40-0.20}} \\
\midrule
PASTA (Ours) & \textbf{0.275 $\pm$ 0.065} & 127.797 $\pm$ 10.253 & \textbf{51.372 $\pm$ 9.268} & 256.518 $\pm$ 25.545 & 23.205 $\pm$ 1.841 \\
Branched Unweighted & \underline{0.257 $\pm$ 0.031} & \textbf{139.622 $\pm$ 4.419} & \textcolor{gray}{41.159 $\pm$ 2.843} & \textbf{296.487 $\pm$ 12.808} & \textcolor{gray}{22.819 $\pm$ 2.723} \\
Shared Unweighted & \textcolor{gray}{0.218 $\pm$ 0.039} & \textcolor{gray}{120.384 $\pm$ 10.132} & 41.316 $\pm$ 5.860 & \textcolor{gray}{247.631 $\pm$ 24.992} & \textbf{24.024 $\pm$ 1.029} \\
Shared Weighted & 0.249 $\pm$ 0.042 & \underline{129.810 $\pm$ 8.741} & \underline{42.594 $\pm$ 4.316} & \underline{269.969 $\pm$ 24.176} & \underline{23.925 $\pm$ 2.457} \\
\midrule
\multicolumn{6}{c}{\textbf{Preferences: 0.50-0.30-0.20}} \\
\midrule
PASTA (Ours) & \textbf{0.314 $\pm$ 0.046} & \underline{110.091 $\pm$ 8.617} & \textbf{55.086 $\pm$ 5.878} & \textcolor{gray}{258.818 $\pm$ 31.954} & 24.517 $\pm$ 1.208 \\
Branched Unweighted & \underline{0.297 $\pm$ 0.063} & \textbf{110.511 $\pm$ 7.794} & \underline{48.929 $\pm$ 9.150} & \underline{270.235 $\pm$ 28.667} & \textbf{24.881 $\pm$ 1.368} \\
Shared Unweighted & \textcolor{gray}{0.255 $\pm$ 0.077} & \textcolor{gray}{107.706 $\pm$ 8.215} & \textcolor{gray}{42.938 $\pm$ 7.102} & \textbf{271.689 $\pm$ 23.074} & \textcolor{gray}{23.652 $\pm$ 2.917} \\
Shared Weighted & 0.270 $\pm$ 0.065 & 107.949 $\pm$ 6.577 & 45.834 $\pm$ 10.136 & 267.069 $\pm$ 26.868 & \underline{24.553 $\pm$ 2.658} \\
\midrule
\multicolumn{6}{c}{\textbf{Preferences: 0.60-0.30-0.10}} \\
\midrule
PASTA (Ours) & \textbf{0.266 $\pm$ 0.043} & \textbf{114.676 $\pm$ 5.266} & \textcolor{gray}{42.179 $\pm$ 4.961} & \textbf{289.966 $\pm$ 18.784} & \underline{23.787 $\pm$ 3.410} \\
Branched Unweighted & \underline{0.266 $\pm$ 0.046} & \underline{113.237 $\pm$ 6.184} & \textbf{44.456 $\pm$ 3.326} & \underline{280.843 $\pm$ 16.260} & 23.102 $\pm$ 1.433 \\
Shared Unweighted & 0.240 $\pm$ 0.046 & \textcolor{gray}{105.553 $\pm$ 5.820} & \underline{42.453 $\pm$ 5.527} & \textcolor{gray}{258.856 $\pm$ 15.574} & \textbf{24.243 $\pm$ 1.831} \\
Shared Weighted & \textcolor{gray}{0.225 $\pm$ 0.030} & 107.395 $\pm$ 6.674 & 42.357 $\pm$ 7.345 & 265.731 $\pm$ 33.690 & \textcolor{gray}{22.619 $\pm$ 2.904} \\
\midrule
\multicolumn{6}{c}{\textbf{Preferences: 0.80-0.10-0.10}} \\
\midrule
PASTA (Ours) & 0.260 $\pm$ 0.047 & \textbf{66.517 $\pm$ 5.864} & 43.731 $\pm$ 7.953 & \textbf{293.086 $\pm$ 38.983} & \textcolor{gray}{22.240 $\pm$ 0.426} \\
Branched Unweighted & \textbf{0.273 $\pm$ 0.073} & \textcolor{gray}{63.981 $\pm$ 5.851} & \textcolor{gray}{42.271 $\pm$ 7.151} & 276.092 $\pm$ 17.316 & \textbf{25.546 $\pm$ 2.795} \\
Shared Unweighted & \textcolor{gray}{0.251 $\pm$ 0.036} & 65.310 $\pm$ 5.732 & \textbf{45.065 $\pm$ 6.565} & \textcolor{gray}{269.603 $\pm$ 24.742} & 22.974 $\pm$ 1.718 \\
Shared Weighted & \underline{0.267 $\pm$ 0.036} & \underline{65.461 $\pm$ 5.154} & \underline{44.156 $\pm$ 8.067} & \underline{277.166 $\pm$ 15.378} & \underline{24.200 $\pm$ 1.228} \\
\bottomrule
\end{tabular}
\end{table*}

Table \ref{tab:ablations-critic} presents a comparison of different critic architectures.
We evaluate ``Shared'' versus ``Branched'' heads for value estimation and assess the impact of weighting the value loss by the proposed ASTCH attention weights (Eq. \ref{eq:stch_att_with_grad_preservation}).
The data indicates that separating the value heads (Branched) is significantly more effective than sharing parameters, likely due to the distinct reward scales and dynamics of the competing objectives. Furthermore, weighting the branched heads' loss using the ASTCH attention, as described in Eq. \eqref{eq:astch-value-loss}, proved crucial to further improve the overall performance.

\subsubsection{Fixed Smoothness} \label{sec:appendix-ablation-fixedmu}
% PACKAGES REQUIRED: \usepackage{booktabs}, \usepackage{xcolor}
\begin{table*}[h]
\centering
\fontsize{6pt}{7pt}\selectfont
\caption{Results (Ablations Fixed Smoothness): MO-StealthVisualSearch. Metrics computed over 5 seeds.} \label{tab:ablations-fixedmu}
\begin{tabular}{lcccccc}
\toprule
Method & Hypervolume $\uparrow$ & E. Utility $\uparrow$ & Obj 0 $\uparrow$ & Obj 1 $\uparrow$ & Obj 2 $\uparrow$ \\
\midrule
\multicolumn{6}{c}{\textbf{Preferences: 0.10-0.70-0.20}} \\
\midrule
PASTA (Ours) & 0.280 $\pm$ 0.015 & 195.274 $\pm$ 5.948 & \underline{49.537 $\pm$ 5.084} & 265.126 $\pm$ 8.716 & 23.660 $\pm$ 1.866 \\
Mu=0.01 & 0.275 $\pm$ 0.060 & 199.244 $\pm$ 12.228 & 45.681 $\pm$ 5.207 & 271.192 $\pm$ 16.861 & 24.210 $\pm$ 2.093 \\
Mu=0.1 & 0.283 $\pm$ 0.079 & \underline{210.295 $\pm$ 14.407} & \textcolor{gray}{43.649 $\pm$ 5.677} & \underline{287.203 $\pm$ 19.625} & \underline{24.443 $\pm$ 4.324} \\
Mu=0.5 & \underline{0.284 $\pm$ 0.057} & \textcolor{gray}{188.617 $\pm$ 24.551} & \textbf{51.168 $\pm$ 5.428} & \textcolor{gray}{255.200 $\pm$ 35.290} & 24.301 $\pm$ 2.309 \\
Mu=1.0 & 0.267 $\pm$ 0.028 & 199.883 $\pm$ 13.932 & 46.168 $\pm$ 5.113 & 272.254 $\pm$ 20.252 & 23.441 $\pm$ 1.068 \\
Mu=5.0 & \textbf{0.331 $\pm$ 0.057} & \textbf{211.849 $\pm$ 7.707} & 45.659 $\pm$ 5.364 & \textbf{288.293 $\pm$ 11.689} & \textbf{27.389 $\pm$ 3.623} \\
Mu=10.0 & \textcolor{gray}{0.259 $\pm$ 0.036} & 194.273 $\pm$ 20.110 & 46.693 $\pm$ 3.959 & 264.171 $\pm$ 28.973 & \textcolor{gray}{23.419 $\pm$ 2.188} \\
\midrule
\multicolumn{6}{c}{\textbf{Preferences: 0.20-0.20-0.60}} \\
\midrule
PASTA (Ours) & 0.297 $\pm$ 0.024 & 78.147 $\pm$ 3.245 & \underline{49.639 $\pm$ 4.819} & 266.512 $\pm$ 17.545 & \underline{24.861 $\pm$ 1.108} \\
Mu=0.01 & 0.279 $\pm$ 0.024 & 77.245 $\pm$ 3.744 & 48.167 $\pm$ 4.393 & 265.294 $\pm$ 20.599 & 24.254 $\pm$ 1.615 \\
Mu=0.1 & \textbf{0.310 $\pm$ 0.039} & 80.377 $\pm$ 4.857 & \textbf{50.767 $\pm$ 11.255} & 276.876 $\pm$ 29.144 & 24.748 $\pm$ 1.490 \\
Mu=0.5 & \underline{0.308 $\pm$ 0.051} & \underline{81.647 $\pm$ 3.152} & 47.327 $\pm$ 5.778 & \underline{286.680 $\pm$ 14.509} & 24.742 $\pm$ 3.077 \\
Mu=1.0 & \textcolor{gray}{0.260 $\pm$ 0.053} & \textcolor{gray}{74.992 $\pm$ 3.551} & \textcolor{gray}{45.525 $\pm$ 7.989} & \textcolor{gray}{254.178 $\pm$ 11.650} & \textbf{25.087 $\pm$ 1.424} \\
Mu=5.0 & 0.275 $\pm$ 0.036 & 79.261 $\pm$ 3.873 & 49.430 $\pm$ 3.546 & 282.239 $\pm$ 19.152 & \textcolor{gray}{21.546 $\pm$ 1.953} \\
Mu=10.0 & 0.291 $\pm$ 0.051 & \textbf{83.014 $\pm$ 4.186} & 46.193 $\pm$ 5.551 & \textbf{301.489 $\pm$ 20.571} & 22.463 $\pm$ 1.181 \\
\midrule
\multicolumn{6}{c}{\textbf{Preferences: 0.20-0.60-0.20}} \\
\midrule
PASTA (Ours) & \textbf{0.315 $\pm$ 0.064} & \textcolor{gray}{169.322 $\pm$ 21.980} & \textbf{56.074 $\pm$ 7.651} & \textcolor{gray}{255.315 $\pm$ 36.800} & \underline{24.589 $\pm$ 1.513} \\
Mu=0.01 & 0.284 $\pm$ 0.033 & 172.024 $\pm$ 15.869 & 50.637 $\pm$ 1.401 & 261.916 $\pm$ 26.628 & 23.735 $\pm$ 1.204 \\
Mu=0.1 & \underline{0.303 $\pm$ 0.051} & \underline{178.337 $\pm$ 14.910} & \underline{51.487 $\pm$ 3.217} & 272.191 $\pm$ 24.256 & 23.625 $\pm$ 1.183 \\
Mu=0.5 & 0.272 $\pm$ 0.068 & 174.349 $\pm$ 23.204 & 48.934 $\pm$ 8.144 & 266.602 $\pm$ 39.149 & \textcolor{gray}{23.006 $\pm$ 1.454} \\
Mu=1.0 & 0.290 $\pm$ 0.037 & \textbf{191.634 $\pm$ 14.000} & 43.713 $\pm$ 4.599 & \textbf{296.696 $\pm$ 24.191} & 24.366 $\pm$ 2.937 \\
Mu=5.0 & \textcolor{gray}{0.268 $\pm$ 0.062} & 178.091 $\pm$ 19.491 & \textcolor{gray}{42.387 $\pm$ 5.249} & \underline{274.257 $\pm$ 32.330} & \textbf{25.296 $\pm$ 2.989} \\
Mu=10.0 & 0.273 $\pm$ 0.035 & 175.800 $\pm$ 11.821 & 46.272 $\pm$ 6.290 & 269.486 $\pm$ 21.361 & 24.269 $\pm$ 1.987 \\
\midrule
\multicolumn{6}{c}{\textbf{Preferences: 0.33-0.33-0.33}} \\
\midrule
PASTA (Ours) & \textbf{0.320 $\pm$ 0.023} & \underline{118.530 $\pm$ 4.456} & \underline{49.680 $\pm$ 5.938} & \underline{280.646 $\pm$ 17.195} & \textbf{25.263 $\pm$ 2.799} \\
Mu=0.01 & 0.265 $\pm$ 0.025 & 112.263 $\pm$ 5.905 & \textbf{52.001 $\pm$ 10.960} & \textcolor{gray}{262.778 $\pm$ 27.054} & \textcolor{gray}{22.011 $\pm$ 1.859} \\
Mu=0.1 & 0.261 $\pm$ 0.042 & 114.200 $\pm$ 14.068 & 46.971 $\pm$ 6.759 & 272.718 $\pm$ 48.927 & 22.910 $\pm$ 1.971 \\
Mu=0.5 & 0.282 $\pm$ 0.044 & 116.417 $\pm$ 11.589 & 46.295 $\pm$ 7.475 & 278.654 $\pm$ 37.627 & 24.303 $\pm$ 2.468 \\
Mu=1.0 & 0.294 $\pm$ 0.042 & 114.472 $\pm$ 4.972 & 48.586 $\pm$ 6.201 & 270.175 $\pm$ 15.698 & 24.655 $\pm$ 1.257 \\
Mu=5.0 & \underline{0.305 $\pm$ 0.018} & \textbf{119.201 $\pm$ 8.039} & 47.341 $\pm$ 8.622 & \textbf{285.174 $\pm$ 31.509} & \underline{25.088 $\pm$ 1.779} \\
Mu=10.0 & \textcolor{gray}{0.228 $\pm$ 0.036} & \textcolor{gray}{110.379 $\pm$ 11.693} & \textcolor{gray}{40.391 $\pm$ 1.877} & 267.343 $\pm$ 36.054 & 23.404 $\pm$ 1.675 \\
\midrule
\multicolumn{6}{c}{\textbf{Preferences: 0.40-0.40-0.20}} \\
\midrule
PASTA (Ours) & 0.275 $\pm$ 0.065 & \textcolor{gray}{127.797 $\pm$ 10.253} & \underline{51.372 $\pm$ 9.268} & \textcolor{gray}{256.518 $\pm$ 25.545} & 23.205 $\pm$ 1.841 \\
Mu=0.01 & 0.276 $\pm$ 0.037 & 132.230 $\pm$ 14.466 & 49.346 $\pm$ 4.962 & 269.652 $\pm$ 39.543 & 23.154 $\pm$ 2.892 \\
Mu=0.1 & \textcolor{gray}{0.269 $\pm$ 0.075} & 128.765 $\pm$ 9.404 & 48.940 $\pm$ 8.024 & 261.459 $\pm$ 20.637 & \textcolor{gray}{23.025 $\pm$ 2.097} \\
Mu=0.5 & \underline{0.317 $\pm$ 0.029} & \underline{139.748 $\pm$ 8.039} & 48.827 $\pm$ 4.049 & \textbf{288.265 $\pm$ 21.791} & 24.555 $\pm$ 1.744 \\
Mu=1.0 & \textbf{0.352 $\pm$ 0.042} & \textbf{139.989 $\pm$ 6.336} & \textbf{53.020 $\pm$ 5.617} & \underline{284.256 $\pm$ 13.828} & \underline{25.391 $\pm$ 1.334} \\
Mu=5.0 & 0.286 $\pm$ 0.023 & 131.319 $\pm$ 6.696 & 47.942 $\pm$ 7.392 & 267.924 $\pm$ 23.074 & 24.862 $\pm$ 0.805 \\
Mu=10.0 & 0.289 $\pm$ 0.025 & 132.389 $\pm$ 6.578 & \textcolor{gray}{46.070 $\pm$ 7.185} & 272.071 $\pm$ 21.676 & \textbf{25.662 $\pm$ 0.995} \\
\midrule
\multicolumn{6}{c}{\textbf{Preferences: 0.50-0.30-0.20}} \\
\midrule
PASTA (Ours) & \textbf{0.314 $\pm$ 0.046} & \textbf{110.091 $\pm$ 8.617} & \textbf{55.086 $\pm$ 5.878} & 258.818 $\pm$ 31.954 & \textbf{24.517 $\pm$ 1.208} \\
Mu=0.01 & \underline{0.267 $\pm$ 0.063} & \underline{108.576 $\pm$ 8.826} & 47.432 $\pm$ 8.852 & 267.476 $\pm$ 24.051 & 23.085 $\pm$ 0.809 \\
Mu=0.1 & \textcolor{gray}{0.238 $\pm$ 0.037} & \textcolor{gray}{104.491 $\pm$ 5.949} & 46.104 $\pm$ 7.288 & 256.372 $\pm$ 25.561 & \textcolor{gray}{22.635 $\pm$ 2.201} \\
Mu=0.5 & 0.266 $\pm$ 0.071 & 108.077 $\pm$ 9.274 & 46.006 $\pm$ 6.431 & \underline{267.981 $\pm$ 23.880} & 23.396 $\pm$ 1.871 \\
Mu=1.0 & 0.260 $\pm$ 0.040 & 108.555 $\pm$ 4.915 & 45.662 $\pm$ 6.158 & \textbf{270.193 $\pm$ 20.076} & 23.329 $\pm$ 2.542 \\
Mu=5.0 & 0.249 $\pm$ 0.031 & 104.700 $\pm$ 6.342 & \underline{47.969 $\pm$ 4.772} & \textcolor{gray}{253.826 $\pm$ 19.607} & 22.839 $\pm$ 0.813 \\
Mu=10.0 & 0.243 $\pm$ 0.022 & 104.935 $\pm$ 4.394 & \textcolor{gray}{45.016 $\pm$ 7.280} & 259.138 $\pm$ 18.374 & \underline{23.427 $\pm$ 1.125} \\
\midrule
\multicolumn{6}{c}{\textbf{Preferences: 0.60-0.30-0.10}} \\
\midrule
PASTA (Ours) & 0.266 $\pm$ 0.043 & \textbf{114.676 $\pm$ 5.266} & \textcolor{gray}{42.179 $\pm$ 4.961} & \textbf{289.966 $\pm$ 18.784} & 23.787 $\pm$ 3.410 \\
Mu=0.01 & 0.286 $\pm$ 0.042 & 112.161 $\pm$ 5.793 & 49.767 $\pm$ 7.271 & 266.375 $\pm$ 6.287 & \underline{23.881 $\pm$ 1.433} \\
Mu=0.1 & 0.263 $\pm$ 0.056 & 109.496 $\pm$ 8.381 & 47.160 $\pm$ 5.801 & 262.853 $\pm$ 25.563 & 23.439 $\pm$ 2.291 \\
Mu=0.5 & \underline{0.287 $\pm$ 0.030} & \underline{114.430 $\pm$ 7.292} & 48.810 $\pm$ 3.986 & \underline{275.977 $\pm$ 30.278} & 23.504 $\pm$ 1.945 \\
Mu=1.0 & 0.264 $\pm$ 0.027 & 108.389 $\pm$ 5.726 & \underline{50.707 $\pm$ 5.040} & \textcolor{gray}{252.146 $\pm$ 26.997} & \textcolor{gray}{23.207 $\pm$ 1.416} \\
Mu=5.0 & \textbf{0.298 $\pm$ 0.045} & 111.239 $\pm$ 5.646 & \textbf{53.140 $\pm$ 7.911} & 256.361 $\pm$ 19.786 & \textbf{24.465 $\pm$ 0.906} \\
Mu=10.0 & \textcolor{gray}{0.232 $\pm$ 0.052} & \textcolor{gray}{105.706 $\pm$ 7.509} & 42.611 $\pm$ 8.000 & 259.252 $\pm$ 29.888 & 23.641 $\pm$ 2.895 \\
\midrule
\multicolumn{6}{c}{\textbf{Preferences: 0.80-0.10-0.10}} \\
\midrule
PASTA (Ours) & 0.260 $\pm$ 0.047 & 66.517 $\pm$ 5.864 & \textcolor{gray}{43.731 $\pm$ 7.953} & \textbf{293.086 $\pm$ 38.983} & 22.240 $\pm$ 0.426 \\
Mu=0.01 & 0.278 $\pm$ 0.046 & 66.860 $\pm$ 3.043 & 46.161 $\pm$ 5.906 & 275.085 $\pm$ 35.581 & 24.226 $\pm$ 2.193 \\
Mu=0.1 & \underline{0.294 $\pm$ 0.033} & \underline{67.397 $\pm$ 1.456} & \underline{46.945 $\pm$ 3.871} & 273.025 $\pm$ 24.638 & \textbf{25.381 $\pm$ 2.625} \\
Mu=0.5 & \textcolor{gray}{0.222 $\pm$ 0.018} & \textcolor{gray}{64.236 $\pm$ 3.067} & 44.112 $\pm$ 3.483 & 268.632 $\pm$ 19.010 & \textcolor{gray}{20.828 $\pm$ 1.936} \\
Mu=1.0 & 0.264 $\pm$ 0.052 & 65.520 $\pm$ 5.928 & 45.641 $\pm$ 9.307 & \textcolor{gray}{265.862 $\pm$ 17.510} & 24.217 $\pm$ 1.442 \\
Mu=5.0 & \textbf{0.310 $\pm$ 0.047} & \textbf{70.942 $\pm$ 5.992} & \textbf{51.477 $\pm$ 9.792} & 273.153 $\pm$ 23.621 & \underline{24.454 $\pm$ 1.580} \\
Mu=10.0 & 0.280 $\pm$ 0.045 & 66.341 $\pm$ 4.161 & 44.870 $\pm$ 4.888 & \underline{280.054 $\pm$ 6.660} & 24.392 $\pm$ 2.480 \\
\bottomrule
\end{tabular}
\end{table*}

In Table \ref{tab:ablations-fixedmu}, we compare the adaptive nature of the PASTA algorithm against a sweep of fixed smoothness values ranging from $\mu=0.01$ to $\mu=10.0$.
This sweep characterizes the trade-off surface: low values mimic Tchebycheff behavior while high values approximate linear scalarization.
The results demonstrate that PASTA successfully converges to the high-performing region of this spectrum ($\mu \approx 5.0$) without the need for the extensive and expensive grid search performed here.

\subsection{Extended Sensitivity Analysis} \label{sec:appendix-sensitivity}

This section provides granular experimental results for the sensitivity analysis of the key hyperparameters.
For the \textit{Conflict Ratio} ($\kappa$), Table \ref{tab:appendix_sensitivity_kappa} reveals a clear performance peak, with deviations in either direction reducing HV, identifying $\kappa = 0.4$ as the optimal threshold for balancing conflict sensitivity.
The sweep over the \textit{Moving Average} factor ($\lambda$), reported in Table \ref{tab:appendix_sensitivity_lambda}, favors a low value ($\lambda = 0.05$), confirming that a stable, slowly-updating memory of gradient conflict is preferred over highly reactive signals.
Finally, Table \ref{tab:appendix_sensitivity_rho} shows that the \textit{Maintenance Rate} ($\rho$) is highly robust across a broad range of values (from our chosen $0.15$ up to $0.8$), while setting it too low ($\rho = 0.05$) causes a significant performance drop, supporting the need for a minimum maintenance threshold to prevent degradation, which we hypothesize arises from vanishing gradients in STCH.

Additionally, we add a sensitivity analysis for the utopia point $z^*$ in Table \ref{tab:ablation_utopia_all}. Note that \citet[Sec. 2]{steuer1983interactive_compressed} define it as $z^*_i = \max \{f_i(x) \mid x\in \mathcal{X}\} + \epsilon_i$, where $\epsilon_i \geq 0$ but $\epsilon_i = 0$ \textit{only} under very specific cases. While setting $\epsilon_i = 0$ (i.e., $z^*=1.0$) yields a marginally higher Hypervolume, the gap is negligible. Since strictly infeasible reference points ($\epsilon_i > 0$) carry fewer theoretical restrictions in TCH formulations \cite[Sec.~2]{steuer1983interactive_compressed}, we maintain the $z^*=1.05$ recommendation. % as our recommended default. 
Furthermore, since the Win Rate (WR) is a zero-sum metric distributed across the evaluated pool, omitting the $z^*=1.0$ configuration naturally redistributes its win share, neutralizing its apparent advantage.  Crucially, %consistent with our other analyses, 
\underline{every} evaluated $z^*$ %configuration %still strictly 
outperforms all baseline methods. This shows that our method is robust to its exact selection, %greatly 
lowering the fine-tuning requirements.

% PACKAGES REQUIRED: \usepackage{booktabs}, \usepackage{xcolor}
\begin{table*}[h]
\centering
\fontsize{6pt}{7pt}\selectfont
\caption{Results (Sensitivity Conflict Ratio $\kappa$): MO-StealthVisualSearch. Metrics computed over 5 seeds.} \label{tab:appendix_sensitivity_kappa}
\begin{tabular}{lcccccc}
\toprule
Method & Hypervolume $\uparrow$ & E. Utility $\uparrow$ & Obj 0 $\uparrow$ & Obj 1 $\uparrow$ & Obj 2 $\uparrow$ \\
\midrule
\multicolumn{6}{c}{\textbf{Preferences: 0.10-0.70-0.20}} \\
\midrule
PASTA (Ours) & \underline{0.280 $\pm$ 0.015} & 195.274 $\pm$ 5.948 & \textbf{49.537 $\pm$ 5.084} & 265.126 $\pm$ 8.716 & \textcolor{gray}{23.660 $\pm$ 1.866} \\
CR=0.2 & \textcolor{gray}{0.272 $\pm$ 0.045} & \textcolor{gray}{192.625 $\pm$ 19.792} & \underline{47.353 $\pm$ 4.224} & \textcolor{gray}{261.429 $\pm$ 28.247} & \underline{24.448 $\pm$ 2.561} \\
CR=0.6 & \textbf{0.283 $\pm$ 0.020} & \underline{210.605 $\pm$ 16.327} & 43.103 $\pm$ 1.642 & \underline{287.588 $\pm$ 23.740} & \textbf{24.917 $\pm$ 1.873} \\
CR=0.8 & 0.274 $\pm$ 0.033 & \textbf{212.469 $\pm$ 11.555} & \textcolor{gray}{42.953 $\pm$ 4.847} & \textbf{290.533 $\pm$ 17.287} & 24.004 $\pm$ 2.809 \\
\midrule
\multicolumn{6}{c}{\textbf{Preferences: 0.20-0.20-0.60}} \\
\midrule
PASTA (Ours) & \textbf{0.297 $\pm$ 0.024} & \underline{78.147 $\pm$ 3.245} & \underline{49.639 $\pm$ 4.819} & 266.512 $\pm$ 17.545 & \textbf{24.861 $\pm$ 1.108} \\
CR=0.2 & \underline{0.297 $\pm$ 0.053} & \textcolor{gray}{76.512 $\pm$ 5.074} & \textbf{51.925 $\pm$ 4.965} & \textcolor{gray}{256.796 $\pm$ 20.716} & \underline{24.613 $\pm$ 1.267} \\
CR=0.6 & \textcolor{gray}{0.279 $\pm$ 0.074} & 77.793 $\pm$ 5.075 & 48.421 $\pm$ 5.104 & \underline{271.480 $\pm$ 19.438} & \textcolor{gray}{23.022 $\pm$ 2.924} \\
CR=0.8 & 0.294 $\pm$ 0.097 & \textbf{79.520 $\pm$ 4.025} & \textcolor{gray}{48.278 $\pm$ 8.289} & \textbf{279.006 $\pm$ 12.407} & 23.439 $\pm$ 3.276 \\
\midrule
\multicolumn{6}{c}{\textbf{Preferences: 0.20-0.60-0.20}} \\
\midrule
PASTA (Ours) & \textbf{0.315 $\pm$ 0.064} & 169.322 $\pm$ 21.980 & \textbf{56.074 $\pm$ 7.651} & 255.315 $\pm$ 36.800 & \underline{24.589 $\pm$ 1.513} \\
CR=0.2 & \textcolor{gray}{0.253 $\pm$ 0.039} & \textbf{189.050 $\pm$ 15.813} & \textcolor{gray}{42.284 $\pm$ 5.725} & \textbf{293.595 $\pm$ 26.726} & \textcolor{gray}{22.177 $\pm$ 0.997} \\
CR=0.6 & \underline{0.273 $\pm$ 0.046} & \underline{175.032 $\pm$ 10.091} & 45.749 $\pm$ 6.698 & \underline{268.194 $\pm$ 16.042} & \textbf{24.827 $\pm$ 4.614} \\
CR=0.8 & 0.264 $\pm$ 0.029 & \textcolor{gray}{162.360 $\pm$ 10.274} & \underline{49.392 $\pm$ 4.263} & \textcolor{gray}{245.953 $\pm$ 16.953} & 24.549 $\pm$ 2.513 \\
\midrule
\multicolumn{6}{c}{\textbf{Preferences: 0.33-0.33-0.33}} \\
\midrule
PASTA (Ours) & \textbf{0.320 $\pm$ 0.023} & \textbf{118.530 $\pm$ 4.456} & \textbf{49.680 $\pm$ 5.938} & \underline{280.646 $\pm$ 17.195} & \textbf{25.263 $\pm$ 2.799} \\
CR=0.2 & \textcolor{gray}{0.240 $\pm$ 0.014} & \underline{115.660 $\pm$ 9.147} & \underline{44.889 $\pm$ 4.487} & \textbf{281.067 $\pm$ 32.347} & \textcolor{gray}{21.025 $\pm$ 1.241} \\
CR=0.6 & 0.251 $\pm$ 0.016 & \textcolor{gray}{110.453 $\pm$ 6.709} & \textcolor{gray}{43.379 $\pm$ 2.591} & \textcolor{gray}{263.481 $\pm$ 23.382} & 24.499 $\pm$ 1.724 \\
CR=0.8 & \underline{0.267 $\pm$ 0.054} & 112.730 $\pm$ 3.313 & 43.419 $\pm$ 7.820 & 269.624 $\pm$ 13.152 & \underline{25.147 $\pm$ 2.070} \\
\midrule
\multicolumn{6}{c}{\textbf{Preferences: 0.40-0.40-0.20}} \\
\midrule
PASTA (Ours) & 0.275 $\pm$ 0.065 & 127.797 $\pm$ 10.253 & \textbf{51.372 $\pm$ 9.268} & 256.518 $\pm$ 25.545 & \textcolor{gray}{23.205 $\pm$ 1.841} \\
CR=0.2 & \textcolor{gray}{0.270 $\pm$ 0.036} & \textcolor{gray}{126.090 $\pm$ 6.904} & \underline{49.683 $\pm$ 3.872} & \textcolor{gray}{253.620 $\pm$ 17.756} & \underline{23.843 $\pm$ 1.153} \\
CR=0.6 & \textbf{0.309 $\pm$ 0.040} & \textbf{134.643 $\pm$ 4.924} & 48.710 $\pm$ 6.475 & \underline{275.230 $\pm$ 10.330} & \textbf{25.337 $\pm$ 1.074} \\
CR=0.8 & \underline{0.290 $\pm$ 0.060} & \underline{134.464 $\pm$ 8.432} & \textcolor{gray}{47.895 $\pm$ 6.043} & \textbf{276.348 $\pm$ 18.129} & 23.835 $\pm$ 2.485 \\
\midrule
\multicolumn{6}{c}{\textbf{Preferences: 0.50-0.30-0.20}} \\
\midrule
PASTA (Ours) & \textbf{0.314 $\pm$ 0.046} & \textbf{110.091 $\pm$ 8.617} & \textbf{55.086 $\pm$ 5.878} & \textcolor{gray}{258.818 $\pm$ 31.954} & \textbf{24.517 $\pm$ 1.208} \\
CR=0.2 & \textcolor{gray}{0.246 $\pm$ 0.034} & 108.411 $\pm$ 4.816 & \textcolor{gray}{45.417 $\pm$ 4.302} & \underline{270.855 $\pm$ 21.165} & \textcolor{gray}{22.229 $\pm$ 3.183} \\
CR=0.6 & \underline{0.269 $\pm$ 0.024} & \underline{109.934 $\pm$ 4.597} & \underline{46.737 $\pm$ 4.704} & \textbf{272.967 $\pm$ 21.171} & 23.376 $\pm$ 2.131 \\
CR=0.8 & 0.258 $\pm$ 0.041 & \textcolor{gray}{105.627 $\pm$ 4.163} & 45.863 $\pm$ 5.738 & 259.535 $\pm$ 20.070 & \underline{24.174 $\pm$ 2.314} \\
\midrule
\multicolumn{6}{c}{\textbf{Preferences: 0.60-0.30-0.10}} \\
\midrule
PASTA (Ours) & \underline{0.266 $\pm$ 0.043} & 114.676 $\pm$ 5.266 & \textcolor{gray}{42.179 $\pm$ 4.961} & \underline{289.966 $\pm$ 18.784} & \underline{23.787 $\pm$ 3.410} \\
CR=0.2 & 0.263 $\pm$ 0.058 & \underline{115.364 $\pm$ 12.586} & \underline{44.711 $\pm$ 7.170} & 287.588 $\pm$ 48.098 & \textcolor{gray}{22.608 $\pm$ 3.229} \\
CR=0.6 & \textcolor{gray}{0.254 $\pm$ 0.037} & \textcolor{gray}{110.944 $\pm$ 7.189} & 44.013 $\pm$ 5.399 & \textcolor{gray}{274.023 $\pm$ 30.039} & 23.290 $\pm$ 1.912 \\
CR=0.8 & \textbf{0.305 $\pm$ 0.043} & \textbf{118.609 $\pm$ 4.786} & \textbf{45.155 $\pm$ 8.557} & \textbf{296.745 $\pm$ 26.572} & \textbf{24.925 $\pm$ 2.042} \\
\midrule
\multicolumn{6}{c}{\textbf{Preferences: 0.80-0.10-0.10}} \\
\midrule
PASTA (Ours) & \textcolor{gray}{0.260 $\pm$ 0.047} & \textcolor{gray}{66.517 $\pm$ 5.864} & \textcolor{gray}{43.731 $\pm$ 7.953} & \textbf{293.086 $\pm$ 38.983} & \textcolor{gray}{22.240 $\pm$ 0.426} \\
CR=0.2 & \underline{0.288 $\pm$ 0.046} & 67.747 $\pm$ 0.849 & \textbf{47.374 $\pm$ 2.599} & \textcolor{gray}{274.141 $\pm$ 21.772} & \textbf{24.339 $\pm$ 2.770} \\
CR=0.6 & \textbf{0.293 $\pm$ 0.039} & \underline{67.994 $\pm$ 3.381} & 46.021 $\pm$ 4.717 & \underline{287.442 $\pm$ 39.461} & \underline{24.336 $\pm$ 1.864} \\
CR=0.8 & 0.286 $\pm$ 0.032 & \textbf{68.390 $\pm$ 5.594} & \underline{46.945 $\pm$ 5.985} & 284.922 $\pm$ 17.608 & 23.427 $\pm$ 1.793 \\
\bottomrule
\end{tabular}
\end{table*}

% PACKAGES REQUIRED: \usepackage{booktabs}, \usepackage{xcolor}
\begin{table*}[h]
\centering
\fontsize{6pt}{7pt}\selectfont
\caption{Results (Sensitivity EMA $\lambda$): MO-StealthVisualSearch. Metrics computed over 5 seeds.} \label{tab:appendix_sensitivity_lambda}
\begin{tabular}{lcccccc}
\toprule
Method & Hypervolume $\uparrow$ & E. Utility $\uparrow$ & Obj 0 $\uparrow$ & Obj 1 $\uparrow$ & Obj 2 $\uparrow$ \\
\midrule
\multicolumn{6}{c}{\textbf{Preferences: 0.10-0.70-0.20}} \\
\midrule
PASTA (Ours) & \textbf{0.280 $\pm$ 0.015} & \underline{195.274 $\pm$ 5.948} & \textbf{49.537 $\pm$ 5.084} & \underline{265.126 $\pm$ 8.716} & \underline{23.660 $\pm$ 1.866} \\
EMA=0.2 & \underline{0.277 $\pm$ 0.055} & \textbf{210.304 $\pm$ 16.259} & \textcolor{gray}{46.255 $\pm$ 6.227} & \textbf{287.341 $\pm$ 23.965} & \textcolor{gray}{22.699 $\pm$ 2.747} \\
EMA=0.5 & \textcolor{gray}{0.277 $\pm$ 0.066} & \textcolor{gray}{191.803 $\pm$ 6.683} & \underline{47.801 $\pm$ 4.918} & \textcolor{gray}{260.192 $\pm$ 8.593} & \textbf{24.441 $\pm$ 3.125} \\
\midrule
\multicolumn{6}{c}{\textbf{Preferences: 0.20-0.20-0.60}} \\
\midrule
PASTA (Ours) & \underline{0.297 $\pm$ 0.024} & \underline{78.147 $\pm$ 3.245} & \underline{49.639 $\pm$ 4.819} & \textcolor{gray}{266.512 $\pm$ 17.545} & \underline{24.861 $\pm$ 1.108} \\
EMA=0.2 & \textbf{0.315 $\pm$ 0.058} & \textbf{79.642 $\pm$ 4.379} & \textbf{50.719 $\pm$ 4.277} & \textbf{272.571 $\pm$ 21.730} & \textbf{24.973 $\pm$ 2.391} \\
EMA=0.5 & \textcolor{gray}{0.269 $\pm$ 0.051} & \textcolor{gray}{77.308 $\pm$ 4.836} & \textcolor{gray}{46.450 $\pm$ 6.256} & \underline{267.685 $\pm$ 17.258} & \textcolor{gray}{24.135 $\pm$ 4.805} \\
\midrule
\multicolumn{6}{c}{\textbf{Preferences: 0.20-0.60-0.20}} \\
\midrule
PASTA (Ours) & \textbf{0.315 $\pm$ 0.064} & \textcolor{gray}{169.322 $\pm$ 21.980} & \textbf{56.074 $\pm$ 7.651} & \textcolor{gray}{255.315 $\pm$ 36.800} & \underline{24.589 $\pm$ 1.513} \\
EMA=0.2 & \textcolor{gray}{0.254 $\pm$ 0.048} & \underline{173.402 $\pm$ 23.442} & \underline{47.392 $\pm$ 5.878} & \underline{265.729 $\pm$ 39.816} & \textcolor{gray}{22.429 $\pm$ 1.184} \\
EMA=0.5 & \underline{0.296 $\pm$ 0.054} & \textbf{180.753 $\pm$ 17.105} & \textcolor{gray}{46.013 $\pm$ 2.037} & \textbf{277.470 $\pm$ 28.166} & \textbf{25.342 $\pm$ 2.691} \\
\midrule
\multicolumn{6}{c}{\textbf{Preferences: 0.33-0.33-0.33}} \\
\midrule
PASTA (Ours) & \textbf{0.320 $\pm$ 0.023} & \textbf{118.530 $\pm$ 4.456} & \underline{49.680 $\pm$ 5.938} & \underline{280.646 $\pm$ 17.195} & \textbf{25.263 $\pm$ 2.799} \\
EMA=0.2 & \textcolor{gray}{0.245 $\pm$ 0.026} & \underline{117.341 $\pm$ 11.195} & \textcolor{gray}{42.264 $\pm$ 1.933} & \textbf{287.720 $\pm$ 34.543} & \textcolor{gray}{22.040 $\pm$ 1.545} \\
EMA=0.5 & \underline{0.301 $\pm$ 0.044} & \textcolor{gray}{116.365 $\pm$ 9.958} & \textbf{49.848 $\pm$ 5.279} & \textcolor{gray}{275.132 $\pm$ 31.741} & \underline{24.115 $\pm$ 1.011} \\
\midrule
\multicolumn{6}{c}{\textbf{Preferences: 0.40-0.40-0.20}} \\
\midrule
PASTA (Ours) & \textbf{0.275 $\pm$ 0.065} & \underline{127.797 $\pm$ 10.253} & \textbf{51.372 $\pm$ 9.268} & \textcolor{gray}{256.518 $\pm$ 25.545} & \underline{23.205 $\pm$ 1.841} \\
EMA=0.2 & \textcolor{gray}{0.250 $\pm$ 0.023} & \textcolor{gray}{127.152 $\pm$ 6.288} & \textcolor{gray}{46.827 $\pm$ 5.161} & \underline{259.574 $\pm$ 15.956} & \textcolor{gray}{22.956 $\pm$ 1.325} \\
EMA=0.5 & \underline{0.271 $\pm$ 0.047} & \textbf{132.694 $\pm$ 11.373} & \underline{47.240 $\pm$ 9.134} & \textbf{272.811 $\pm$ 28.184} & \textbf{23.366 $\pm$ 1.418} \\
\midrule
\multicolumn{6}{c}{\textbf{Preferences: 0.50-0.30-0.20}} \\
\midrule
PASTA (Ours) & \textbf{0.314 $\pm$ 0.046} & \textbf{110.091 $\pm$ 8.617} & \textbf{55.086 $\pm$ 5.878} & \underline{258.818 $\pm$ 31.954} & \textbf{24.517 $\pm$ 1.208} \\
EMA=0.2 & \textcolor{gray}{0.236 $\pm$ 0.041} & \underline{109.641 $\pm$ 7.843} & \textcolor{gray}{41.242 $\pm$ 7.977} & \textbf{281.654 $\pm$ 36.218} & \textcolor{gray}{22.620 $\pm$ 2.877} \\
EMA=0.5 & \underline{0.276 $\pm$ 0.054} & \textcolor{gray}{106.647 $\pm$ 6.113} & \underline{49.607 $\pm$ 8.023} & \textcolor{gray}{256.701 $\pm$ 19.587} & \underline{24.166 $\pm$ 1.846} \\
\midrule
\multicolumn{6}{c}{\textbf{Preferences: 0.60-0.30-0.10}} \\
\midrule
PASTA (Ours) & \underline{0.266 $\pm$ 0.043} & \textbf{114.676 $\pm$ 5.266} & \textcolor{gray}{42.179 $\pm$ 4.961} & \textbf{289.966 $\pm$ 18.784} & \textcolor{gray}{23.787 $\pm$ 3.410} \\
EMA=0.2 & \textbf{0.282 $\pm$ 0.038} & \underline{113.275 $\pm$ 5.955} & \textbf{46.828 $\pm$ 3.202} & \textcolor{gray}{275.951 $\pm$ 18.270} & \underline{23.932 $\pm$ 1.928} \\
EMA=0.5 & \textcolor{gray}{0.263 $\pm$ 0.032} & \textcolor{gray}{112.639 $\pm$ 4.923} & \underline{43.221 $\pm$ 6.720} & \underline{281.042 $\pm$ 27.016} & \textbf{23.932 $\pm$ 1.505} \\
\midrule
\multicolumn{6}{c}{\textbf{Preferences: 0.80-0.10-0.10}} \\
\midrule
PASTA (Ours) & \textcolor{gray}{0.260 $\pm$ 0.047} & \textbf{66.517 $\pm$ 5.864} & \underline{43.731 $\pm$ 7.953} & \textbf{293.086 $\pm$ 38.983} & \textcolor{gray}{22.240 $\pm$ 0.426} \\
EMA=0.2 & \underline{0.263 $\pm$ 0.040} & \textcolor{gray}{64.850 $\pm$ 4.018} & \textcolor{gray}{43.661 $\pm$ 3.552} & \underline{275.213 $\pm$ 12.398} & \underline{23.999 $\pm$ 1.819} \\
EMA=0.5 & \textbf{0.295 $\pm$ 0.048} & \underline{65.868 $\pm$ 3.836} & \textbf{44.683 $\pm$ 4.339} & \textcolor{gray}{274.919 $\pm$ 19.985} & \textbf{26.291 $\pm$ 1.553} \\
\bottomrule
\end{tabular}
\end{table*}

% PACKAGES REQUIRED: \usepackage{booktabs}, \usepackage{xcolor}
\begin{table*}[h]
\centering
\fontsize{6pt}{7pt}\selectfont
\caption{Results (Sensitivity Maintenance Rate $\rho$): MO-StealthVisualSearch. Metrics computed over 5 seeds.} \label{tab:appendix_sensitivity_rho}
\begin{tabular}{lcccccc}
\toprule
Method & Hypervolume $\uparrow$ & E. Utility $\uparrow$ & Obj 0 $\uparrow$ & Obj 1 $\uparrow$ & Obj 2 $\uparrow$ \\
\midrule
\multicolumn{6}{c}{\textbf{Preferences: 0.10-0.70-0.20}} \\
\midrule
PASTA (Ours) & 0.280 $\pm$ 0.015 & 195.274 $\pm$ 5.948 & \underline{49.537 $\pm$ 5.084} & 265.126 $\pm$ 8.716 & 23.660 $\pm$ 1.866 \\
MR=0.05 & \textcolor{gray}{0.244 $\pm$ 0.064} & \textcolor{gray}{190.684 $\pm$ 26.002} & 45.859 $\pm$ 4.495 & \textcolor{gray}{259.404 $\pm$ 36.899} & 22.577 $\pm$ 3.371 \\
MR=0.3 & 0.251 $\pm$ 0.032 & 194.787 $\pm$ 22.124 & 46.898 $\pm$ 4.246 & 265.157 $\pm$ 31.918 & \textcolor{gray}{22.439 $\pm$ 1.624} \\
MR=0.5 & \underline{0.301 $\pm$ 0.100} & \textbf{211.908 $\pm$ 22.890} & \textcolor{gray}{44.971 $\pm$ 9.317} & \textbf{289.217 $\pm$ 32.455} & \underline{24.797 $\pm$ 3.150} \\
MR=0.8 & \textbf{0.324 $\pm$ 0.052} & \underline{198.885 $\pm$ 18.681} & \textbf{51.496 $\pm$ 6.124} & \underline{269.399 $\pm$ 26.779} & \textbf{25.780 $\pm$ 1.697} \\
\midrule
\multicolumn{6}{c}{\textbf{Preferences: 0.20-0.20-0.60}} \\
\midrule
PASTA (Ours) & \textbf{0.297 $\pm$ 0.024} & 78.147 $\pm$ 3.245 & \underline{49.639 $\pm$ 4.819} & 266.512 $\pm$ 17.545 & \textbf{24.861 $\pm$ 1.108} \\
MR=0.05 & \textcolor{gray}{0.255 $\pm$ 0.046} & 76.532 $\pm$ 5.920 & \textcolor{gray}{46.204 $\pm$ 5.401} & 267.981 $\pm$ 32.662 & 22.824 $\pm$ 2.025 \\
MR=0.3 & 0.260 $\pm$ 0.043 & \textcolor{gray}{75.416 $\pm$ 4.726} & 47.005 $\pm$ 4.095 & \textcolor{gray}{258.764 $\pm$ 21.100} & \underline{23.770 $\pm$ 1.740} \\
MR=0.5 & \underline{0.291 $\pm$ 0.055} & \textbf{80.979 $\pm$ 4.765} & 48.601 $\pm$ 7.890 & \textbf{288.500 $\pm$ 28.533} & \textcolor{gray}{22.598 $\pm$ 1.679} \\
MR=0.8 & 0.288 $\pm$ 0.039 & \underline{79.441 $\pm$ 4.916} & \textbf{49.821 $\pm$ 5.296} & \underline{277.824 $\pm$ 36.805} & 23.187 $\pm$ 4.181 \\
\midrule
\multicolumn{6}{c}{\textbf{Preferences: 0.20-0.60-0.20}} \\
\midrule
PASTA (Ours) & \underline{0.315 $\pm$ 0.064} & 169.322 $\pm$ 21.980 & \textbf{56.074 $\pm$ 7.651} & \textcolor{gray}{255.315 $\pm$ 36.800} & 24.589 $\pm$ 1.513 \\
MR=0.05 & \textcolor{gray}{0.250 $\pm$ 0.043} & 177.194 $\pm$ 12.945 & \textcolor{gray}{42.929 $\pm$ 9.935} & 273.083 $\pm$ 23.252 & \textcolor{gray}{23.794 $\pm$ 1.207} \\
MR=0.3 & 0.287 $\pm$ 0.041 & \underline{182.753 $\pm$ 8.371} & 43.537 $\pm$ 3.520 & \underline{281.556 $\pm$ 14.012} & \textbf{25.559 $\pm$ 2.698} \\
MR=0.5 & 0.267 $\pm$ 0.066 & \textcolor{gray}{167.876 $\pm$ 14.280} & 47.369 $\pm$ 10.174 & 255.868 $\pm$ 22.883 & 24.409 $\pm$ 1.072 \\
MR=0.8 & \textbf{0.319 $\pm$ 0.036} & \textbf{185.517 $\pm$ 11.707} & \underline{49.085 $\pm$ 7.240} & \textbf{284.448 $\pm$ 20.317} & \underline{25.157 $\pm$ 2.527} \\
\midrule
\multicolumn{6}{c}{\textbf{Preferences: 0.33-0.33-0.33}} \\
\midrule
PASTA (Ours) & \textbf{0.320 $\pm$ 0.023} & \underline{118.530 $\pm$ 4.456} & \textbf{49.680 $\pm$ 5.938} & 280.646 $\pm$ 17.195 & \textbf{25.263 $\pm$ 2.799} \\
MR=0.05 & \underline{0.315 $\pm$ 0.023} & \textbf{123.052 $\pm$ 1.149} & \underline{47.405 $\pm$ 2.913} & \textbf{297.683 $\pm$ 3.377} & \underline{24.069 $\pm$ 1.589} \\
MR=0.3 & \textcolor{gray}{0.250 $\pm$ 0.048} & \textcolor{gray}{114.149 $\pm$ 7.507} & 44.671 $\pm$ 5.151 & \textcolor{gray}{275.467 $\pm$ 24.262} & \textcolor{gray}{22.309 $\pm$ 1.681} \\
MR=0.5 & 0.257 $\pm$ 0.038 & 118.134 $\pm$ 9.133 & \textcolor{gray}{41.309 $\pm$ 3.879} & \underline{289.439 $\pm$ 31.350} & 23.652 $\pm$ 3.790 \\
MR=0.8 & 0.278 $\pm$ 0.009 & 116.102 $\pm$ 3.136 & 45.953 $\pm$ 3.735 & 278.440 $\pm$ 12.255 & 23.913 $\pm$ 1.736 \\
\midrule
\multicolumn{6}{c}{\textbf{Preferences: 0.40-0.40-0.20}} \\
\midrule
PASTA (Ours) & 0.275 $\pm$ 0.065 & \textcolor{gray}{127.797 $\pm$ 10.253} & \textbf{51.372 $\pm$ 9.268} & \textcolor{gray}{256.518 $\pm$ 25.545} & 23.205 $\pm$ 1.841 \\
MR=0.05 & 0.271 $\pm$ 0.036 & 128.436 $\pm$ 5.064 & \underline{50.245 $\pm$ 6.019} & 259.275 $\pm$ 14.241 & 23.140 $\pm$ 1.424 \\
MR=0.3 & \underline{0.276 $\pm$ 0.043} & 130.192 $\pm$ 5.251 & \textcolor{gray}{46.762 $\pm$ 10.154} & 266.281 $\pm$ 23.310 & \textbf{24.874 $\pm$ 2.009} \\
MR=0.5 & \textbf{0.298 $\pm$ 0.072} & \textbf{136.020 $\pm$ 8.206} & 47.370 $\pm$ 11.232 & \textbf{280.363 $\pm$ 18.888} & \underline{24.635 $\pm$ 1.860} \\
MR=0.8 & \textcolor{gray}{0.263 $\pm$ 0.045} & \underline{131.059 $\pm$ 11.072} & 47.678 $\pm$ 4.968 & \underline{268.582 $\pm$ 31.053} & \textcolor{gray}{22.771 $\pm$ 3.008} \\
\midrule
\multicolumn{6}{c}{\textbf{Preferences: 0.50-0.30-0.20}} \\
\midrule
PASTA (Ours) & \underline{0.314 $\pm$ 0.046} & 110.091 $\pm$ 8.617 & \textbf{55.086 $\pm$ 5.878} & \textcolor{gray}{258.818 $\pm$ 31.954} & 24.517 $\pm$ 1.208 \\
MR=0.05 & \textcolor{gray}{0.267 $\pm$ 0.007} & 109.159 $\pm$ 10.730 & 49.342 $\pm$ 4.859 & 266.262 $\pm$ 35.762 & \textcolor{gray}{23.046 $\pm$ 3.894} \\
MR=0.3 & \textbf{0.316 $\pm$ 0.076} & \textbf{112.872 $\pm$ 9.340} & \underline{50.060 $\pm$ 9.436} & \textbf{276.062 $\pm$ 33.627} & \underline{25.119 $\pm$ 1.702} \\
MR=0.5 & 0.291 $\pm$ 0.050 & \underline{110.540 $\pm$ 7.549} & 47.977 $\pm$ 4.709 & \underline{272.094 $\pm$ 27.686} & 24.614 $\pm$ 3.239 \\
MR=0.8 & 0.281 $\pm$ 0.031 & \textcolor{gray}{108.840 $\pm$ 2.626} & \textcolor{gray}{46.340 $\pm$ 7.989} & 268.798 $\pm$ 17.858 & \textbf{25.152 $\pm$ 0.859} \\
\midrule
\multicolumn{6}{c}{\textbf{Preferences: 0.60-0.30-0.10}} \\
\midrule
PASTA (Ours) & 0.266 $\pm$ 0.043 & \underline{114.676 $\pm$ 5.266} & \textcolor{gray}{42.179 $\pm$ 4.961} & \textbf{289.966 $\pm$ 18.784} & 23.787 $\pm$ 3.410 \\
MR=0.05 & 0.251 $\pm$ 0.040 & 112.228 $\pm$ 8.194 & 44.751 $\pm$ 7.715 & \underline{277.113 $\pm$ 32.658} & \textcolor{gray}{22.440 $\pm$ 0.812} \\
MR=0.3 & \underline{0.303 $\pm$ 0.055} & 112.280 $\pm$ 7.929 & \textbf{50.309 $\pm$ 6.854} & 265.303 $\pm$ 17.479 & \underline{25.038 $\pm$ 2.183} \\
MR=0.5 & \textcolor{gray}{0.249 $\pm$ 0.082} & \textcolor{gray}{108.355 $\pm$ 11.365} & 45.929 $\pm$ 10.613 & \textcolor{gray}{261.688 $\pm$ 31.267} & 22.914 $\pm$ 2.443 \\
MR=0.8 & \textbf{0.309 $\pm$ 0.070} & \textbf{114.894 $\pm$ 9.876} & \underline{48.774 $\pm$ 7.395} & 277.056 $\pm$ 36.671 & \textbf{25.124 $\pm$ 2.027} \\
\midrule
\multicolumn{6}{c}{\textbf{Preferences: 0.80-0.10-0.10}} \\
\midrule
PASTA (Ours) & 0.260 $\pm$ 0.047 & \textbf{66.517 $\pm$ 5.864} & 43.731 $\pm$ 7.953 & \underline{293.086 $\pm$ 38.983} & \textcolor{gray}{22.240 $\pm$ 0.426} \\
MR=0.05 & \textcolor{gray}{0.235 $\pm$ 0.018} & \textcolor{gray}{62.159 $\pm$ 1.554} & 42.828 $\pm$ 5.296 & \textcolor{gray}{254.515 $\pm$ 27.772} & \underline{24.445 $\pm$ 1.504} \\
MR=0.3 & \underline{0.267 $\pm$ 0.099} & 64.850 $\pm$ 8.987 & \underline{44.785 $\pm$ 9.544} & 266.101 $\pm$ 32.265 & 24.115 $\pm$ 2.879 \\
MR=0.5 & \textbf{0.291 $\pm$ 0.047} & \underline{66.175 $\pm$ 2.740} & \textbf{45.134 $\pm$ 5.628} & 274.799 $\pm$ 28.227 & \textbf{25.885 $\pm$ 2.808} \\
MR=0.8 & 0.264 $\pm$ 0.044 & 65.970 $\pm$ 3.781 & \textcolor{gray}{42.442 $\pm$ 6.739} & \textbf{297.402 $\pm$ 21.649} & 22.756 $\pm$ 2.278 \\
\bottomrule
\end{tabular}
\end{table*}

% PACKAGES REQUIRED: \usepackage{booktabs}, \usepackage{xcolor}
\begin{table*}[h]
\centering
\fontsize{6pt}{7pt}\selectfont
\caption{Results (Sensitivity Utopia): MO-StealthVisualSearch. Metrics computed over 5 seeds.} \label{tab:ablation_utopia_all}
\begin{tabular}{lcccccc}
\toprule
Method & Hypervolume $\uparrow$ & E. Utility $\uparrow$ & Obj 0 $\uparrow$ & Obj 1 $\uparrow$ & Obj 2 $\uparrow$ \\
\midrule
\multicolumn{6}{c}{\textbf{Preferences: 0.10-0.70-0.20}} \\
\midrule
z=1.0 & \underline{0.284 $\pm$ 0.046} & \underline{204.019 $\pm$ 18.863} & 47.335 $\pm$ 4.444 & \underline{277.926 $\pm$ 26.653} & 23.689 $\pm$ 2.924 \\
z=1.01 & 0.265 $\pm$ 0.028 & 190.366 $\pm$ 19.853 & \textbf{50.559 $\pm$ 4.098} & 258.234 $\pm$ 28.725 & \textcolor{gray}{22.730 $\pm$ 1.216} \\
PASTA (Ours) & 0.280 $\pm$ 0.015 & 195.274 $\pm$ 5.948 & 49.537 $\pm$ 5.084 & 265.126 $\pm$ 8.716 & 23.660 $\pm$ 1.866 \\
z=1.1 & \textcolor{gray}{0.258 $\pm$ 0.039} & \textcolor{gray}{185.258 $\pm$ 13.442} & 47.997 $\pm$ 5.219 & \textcolor{gray}{250.952 $\pm$ 19.326} & \underline{23.962 $\pm$ 2.005} \\
z=1.5 & \textbf{0.323 $\pm$ 0.059} & \textbf{210.389 $\pm$ 11.206} & \underline{50.365 $\pm$ 6.057} & \textbf{286.442 $\pm$ 16.273} & \textbf{24.214 $\pm$ 1.820} \\
z=5.0 & 0.267 $\pm$ 0.027 & 201.783 $\pm$ 21.048 & \textcolor{gray}{47.187 $\pm$ 4.926} & 275.013 $\pm$ 30.858 & 22.777 $\pm$ 1.601 \\
\midrule
\multicolumn{6}{c}{\textbf{Preferences: 0.20-0.20-0.60}} \\
\midrule
z=1.0 & \textbf{0.310 $\pm$ 0.060} & 78.214 $\pm$ 7.449 & \textbf{52.129 $\pm$ 6.781} & \textcolor{gray}{263.375 $\pm$ 36.160} & \textbf{25.189 $\pm$ 2.593} \\
z=1.01 & 0.295 $\pm$ 0.031 & \underline{81.354 $\pm$ 5.859} & 45.668 $\pm$ 7.522 & \underline{286.114 $\pm$ 36.087} & \underline{24.996 $\pm$ 1.309} \\
PASTA (Ours) & \underline{0.297 $\pm$ 0.024} & 78.147 $\pm$ 3.245 & \underline{49.639 $\pm$ 4.819} & 266.512 $\pm$ 17.545 & 24.861 $\pm$ 1.108 \\
z=1.1 & \textcolor{gray}{0.245 $\pm$ 0.011} & 76.492 $\pm$ 2.125 & \textcolor{gray}{41.633 $\pm$ 4.606} & 267.225 $\pm$ 15.824 & 24.534 $\pm$ 1.885 \\
z=1.5 & 0.256 $\pm$ 0.073 & \textcolor{gray}{76.418 $\pm$ 7.751} & 45.718 $\pm$ 8.380 & 267.440 $\pm$ 31.572 & \textcolor{gray}{22.976 $\pm$ 2.909} \\
z=5.0 & 0.294 $\pm$ 0.056 & \textbf{81.539 $\pm$ 4.723} & 47.151 $\pm$ 9.058 & \textbf{290.077 $\pm$ 24.430} & 23.489 $\pm$ 1.198 \\
\midrule
\multicolumn{6}{c}{\textbf{Preferences: 0.20-0.60-0.20}} \\
\midrule
z=1.0 & 0.288 $\pm$ 0.057 & 177.026 $\pm$ 8.974 & 47.400 $\pm$ 8.727 & 271.020 $\pm$ 14.273 & \underline{24.669 $\pm$ 1.774} \\
z=1.01 & 0.271 $\pm$ 0.032 & \textbf{190.191 $\pm$ 16.243} & \textcolor{gray}{42.921 $\pm$ 2.233} & \textbf{294.917 $\pm$ 27.579} & \textcolor{gray}{23.285 $\pm$ 1.878} \\
PASTA (Ours) & \textbf{0.315 $\pm$ 0.064} & \textcolor{gray}{169.322 $\pm$ 21.980} & \textbf{56.074 $\pm$ 7.651} & \textcolor{gray}{255.315 $\pm$ 36.800} & 24.589 $\pm$ 1.513 \\
z=1.1 & \underline{0.311 $\pm$ 0.042} & \underline{187.639 $\pm$ 13.031} & 46.293 $\pm$ 6.267 & \underline{288.844 $\pm$ 21.750} & \textbf{25.370 $\pm$ 0.878} \\
z=1.5 & \textcolor{gray}{0.268 $\pm$ 0.039} & 170.161 $\pm$ 21.050 & \underline{48.878 $\pm$ 6.736} & 259.401 $\pm$ 36.377 & 23.724 $\pm$ 1.704 \\
z=5.0 & 0.291 $\pm$ 0.031 & 183.597 $\pm$ 20.573 & 47.006 $\pm$ 3.841 & 282.293 $\pm$ 35.332 & 24.099 $\pm$ 0.825 \\
\midrule
\multicolumn{6}{c}{\textbf{Preferences: 0.33-0.33-0.33}} \\
\midrule
z=1.0 & \textbf{0.326 $\pm$ 0.017} & \textbf{120.161 $\pm$ 7.428} & \underline{49.452 $\pm$ 4.116} & \textbf{285.771 $\pm$ 26.554} & \underline{25.259 $\pm$ 1.744} \\
z=1.01 & \textcolor{gray}{0.241 $\pm$ 0.048} & \textcolor{gray}{106.270 $\pm$ 4.801} & 47.513 $\pm$ 6.525 & \textcolor{gray}{248.385 $\pm$ 14.451} & 22.911 $\pm$ 2.601 \\
PASTA (Ours) & \underline{0.320 $\pm$ 0.023} & \underline{118.530 $\pm$ 4.456} & \textbf{49.680 $\pm$ 5.938} & \underline{280.646 $\pm$ 17.195} & \textbf{25.263 $\pm$ 2.799} \\
z=1.1 & 0.259 $\pm$ 0.069 & 112.257 $\pm$ 8.722 & 46.953 $\pm$ 10.328 & 266.991 $\pm$ 26.710 & \textcolor{gray}{22.828 $\pm$ 1.168} \\
z=1.5 & 0.272 $\pm$ 0.042 & 115.147 $\pm$ 6.966 & 47.697 $\pm$ 6.805 & 274.863 $\pm$ 23.262 & 22.881 $\pm$ 1.894 \\
z=5.0 & 0.254 $\pm$ 0.057 & 111.270 $\pm$ 12.451 & \textcolor{gray}{45.456 $\pm$ 7.282} & 264.877 $\pm$ 39.132 & 23.477 $\pm$ 2.037 \\
\midrule
\multicolumn{6}{c}{\textbf{Preferences: 0.40-0.40-0.20}} \\
\midrule
z=1.0 & \textcolor{gray}{0.274 $\pm$ 0.043} & \underline{130.761 $\pm$ 11.578} & \underline{49.914 $\pm$ 1.409} & \underline{265.576 $\pm$ 29.018} & \textcolor{gray}{22.827 $\pm$ 1.572} \\
z=1.01 & \textbf{0.290 $\pm$ 0.078} & 130.230 $\pm$ 9.698 & \textcolor{gray}{47.835 $\pm$ 8.080} & 265.167 $\pm$ 25.885 & \textbf{25.148 $\pm$ 2.326} \\
PASTA (Ours) & 0.275 $\pm$ 0.065 & \textcolor{gray}{127.797 $\pm$ 10.253} & \textbf{51.372 $\pm$ 9.268} & \textcolor{gray}{256.518 $\pm$ 25.545} & 23.205 $\pm$ 1.841 \\
z=1.1 & \underline{0.287 $\pm$ 0.053} & \textbf{134.373 $\pm$ 6.229} & 49.036 $\pm$ 7.684 & \textbf{275.225 $\pm$ 16.868} & 23.343 $\pm$ 1.787 \\
z=1.5 & 0.275 $\pm$ 0.052 & 128.978 $\pm$ 4.314 & 48.822 $\pm$ 8.497 & 261.713 $\pm$ 11.026 & 23.820 $\pm$ 1.165 \\
z=5.0 & 0.278 $\pm$ 0.030 & 128.194 $\pm$ 9.102 & 49.300 $\pm$ 9.474 & 258.809 $\pm$ 29.158 & \underline{24.754 $\pm$ 1.531} \\
\midrule
\multicolumn{6}{c}{\textbf{Preferences: 0.50-0.30-0.20}} \\
\midrule
z=1.0 & 0.266 $\pm$ 0.055 & 106.442 $\pm$ 8.045 & 45.401 $\pm$ 5.484 & 262.623 $\pm$ 30.352 & \underline{24.772 $\pm$ 2.760} \\
z=1.01 & \textcolor{gray}{0.253 $\pm$ 0.060} & \textcolor{gray}{105.255 $\pm$ 6.462} & \textcolor{gray}{42.591 $\pm$ 5.995} & 263.152 $\pm$ 21.964 & \textbf{25.067 $\pm$ 4.010} \\
PASTA (Ours) & \underline{0.314 $\pm$ 0.046} & 110.091 $\pm$ 8.617 & \textbf{55.086 $\pm$ 5.878} & 258.818 $\pm$ 31.954 & 24.517 $\pm$ 1.208 \\
z=1.1 & \textbf{0.334 $\pm$ 0.086} & \textbf{118.486 $\pm$ 4.551} & 48.849 $\pm$ 8.311 & \textbf{297.085 $\pm$ 18.178} & 24.678 $\pm$ 3.351 \\
z=1.5 & 0.269 $\pm$ 0.032 & \underline{110.419 $\pm$ 7.720} & 44.583 $\pm$ 4.772 & \underline{277.664 $\pm$ 33.635} & 24.140 $\pm$ 3.385 \\
z=5.0 & 0.265 $\pm$ 0.027 & 106.636 $\pm$ 3.730 & \underline{50.391 $\pm$ 7.901} & \textcolor{gray}{256.039 $\pm$ 18.371} & \textcolor{gray}{23.147 $\pm$ 1.800} \\
\midrule
\multicolumn{6}{c}{\textbf{Preferences: 0.60-0.30-0.10}} \\
\midrule
z=1.0 & \textbf{0.294 $\pm$ 0.057} & \underline{113.368 $\pm$ 4.312} & \underline{48.340 $\pm$ 3.799} & 273.061 $\pm$ 11.104 & \textbf{24.457 $\pm$ 3.822} \\
z=1.01 & 0.264 $\pm$ 0.022 & 112.831 $\pm$ 8.218 & \textbf{49.249 $\pm$ 6.922} & 270.158 $\pm$ 35.134 & 22.346 $\pm$ 3.256 \\
PASTA (Ours) & 0.266 $\pm$ 0.043 & \textbf{114.676 $\pm$ 5.266} & \textcolor{gray}{42.179 $\pm$ 4.961} & \textbf{289.966 $\pm$ 18.784} & \underline{23.787 $\pm$ 3.410} \\
z=1.1 & \underline{0.274 $\pm$ 0.049} & 111.051 $\pm$ 6.830 & 47.862 $\pm$ 5.758 & \textcolor{gray}{266.589 $\pm$ 16.483} & 23.573 $\pm$ 1.481 \\
z=1.5 & 0.256 $\pm$ 0.024 & \textcolor{gray}{110.304 $\pm$ 4.191} & 45.193 $\pm$ 5.736 & 269.482 $\pm$ 24.281 & 23.436 $\pm$ 2.291 \\
z=5.0 & \textcolor{gray}{0.248 $\pm$ 0.041} & 112.376 $\pm$ 4.704 & 45.490 $\pm$ 4.113 & \underline{276.430 $\pm$ 16.907} & \textcolor{gray}{21.531 $\pm$ 1.882} \\
\midrule
\multicolumn{6}{c}{\textbf{Preferences: 0.80-0.10-0.10}} \\
\midrule
z=1.0 & \underline{0.305 $\pm$ 0.044} & \underline{67.383 $\pm$ 4.514} & 45.194 $\pm$ 6.262 & \underline{286.501 $\pm$ 22.141} & \textbf{25.782 $\pm$ 2.370} \\
z=1.01 & \textbf{0.307 $\pm$ 0.065} & \textbf{69.212 $\pm$ 7.122} & \textbf{49.034 $\pm$ 8.968} & 274.958 $\pm$ 15.906 & 24.889 $\pm$ 1.175 \\
PASTA (Ours) & 0.260 $\pm$ 0.047 & 66.517 $\pm$ 5.864 & 43.731 $\pm$ 7.953 & \textbf{293.086 $\pm$ 38.983} & \textcolor{gray}{22.240 $\pm$ 0.426} \\
z=1.1 & 0.274 $\pm$ 0.041 & 67.130 $\pm$ 3.220 & \underline{47.371 $\pm$ 3.894} & \textcolor{gray}{268.622 $\pm$ 22.626} & 23.706 $\pm$ 1.745 \\
z=1.5 & 0.270 $\pm$ 0.057 & 64.619 $\pm$ 6.429 & 42.850 $\pm$ 9.080 & 278.431 $\pm$ 27.263 & \underline{24.956 $\pm$ 1.364} \\
z=5.0 & \textcolor{gray}{0.254 $\pm$ 0.040} & \textcolor{gray}{63.263 $\pm$ 5.715} & \textcolor{gray}{41.780 $\pm$ 8.915} & 273.542 $\pm$ 17.088 & 24.854 $\pm$ 3.236 \\
\bottomrule
\end{tabular}
\end{table*}

\subsection{Extended Results on MuJoCo Environments} \label{sec:appendix-mujoco-results}

We provide additional results for the four multi-objective MuJoCo environments shown in Fig. \ref{fig:mujoco-envs}.

\begin{figure*}[t]
    \centering
    
    % --- First Row ---
    \begin{subfigure}[b]{0.485\textwidth}
        \centering
        \includegraphics[width=\linewidth]{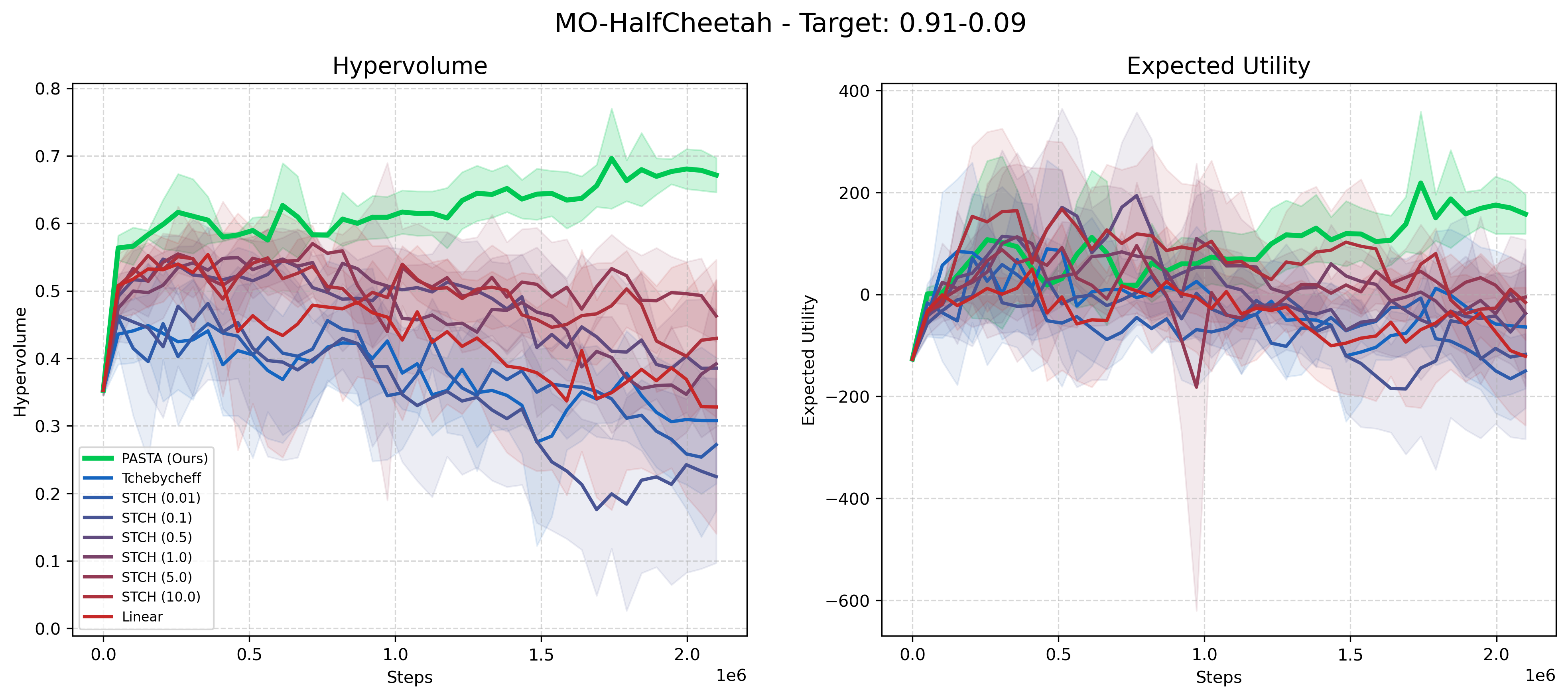} 
        \caption{\small Half Cheetah}
        \label{fig:top-left}
    \end{subfigure}
    \hfill 
    \begin{subfigure}[b]{0.485\textwidth}
        \centering
        \includegraphics[width=\linewidth]{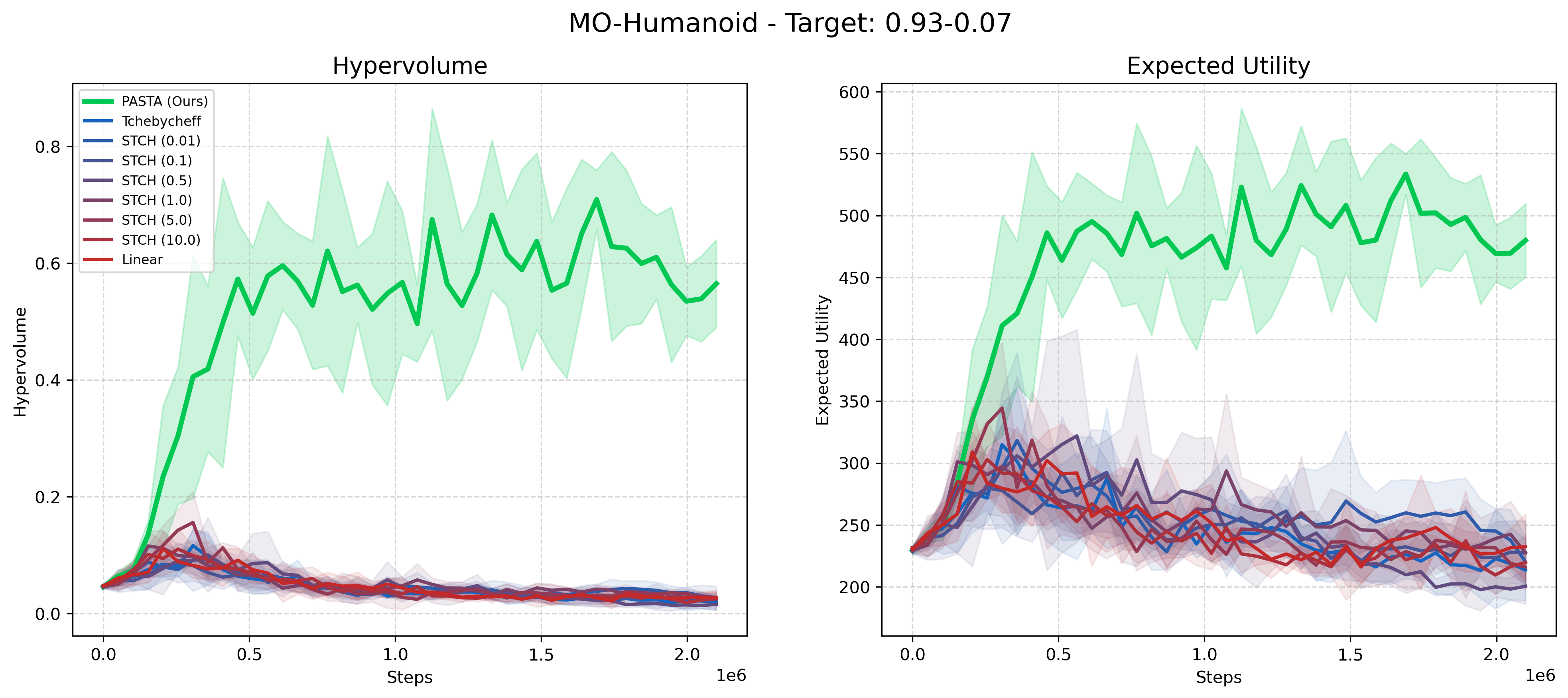}
        \caption{\small Humanoid}
        \label{fig:top-right}
    \end{subfigure}
    
    \vspace{0.2cm}
     
    % --- Second Row ---
    \begin{subfigure}[b]{0.485\textwidth}
        \centering
        \includegraphics[width=\linewidth]{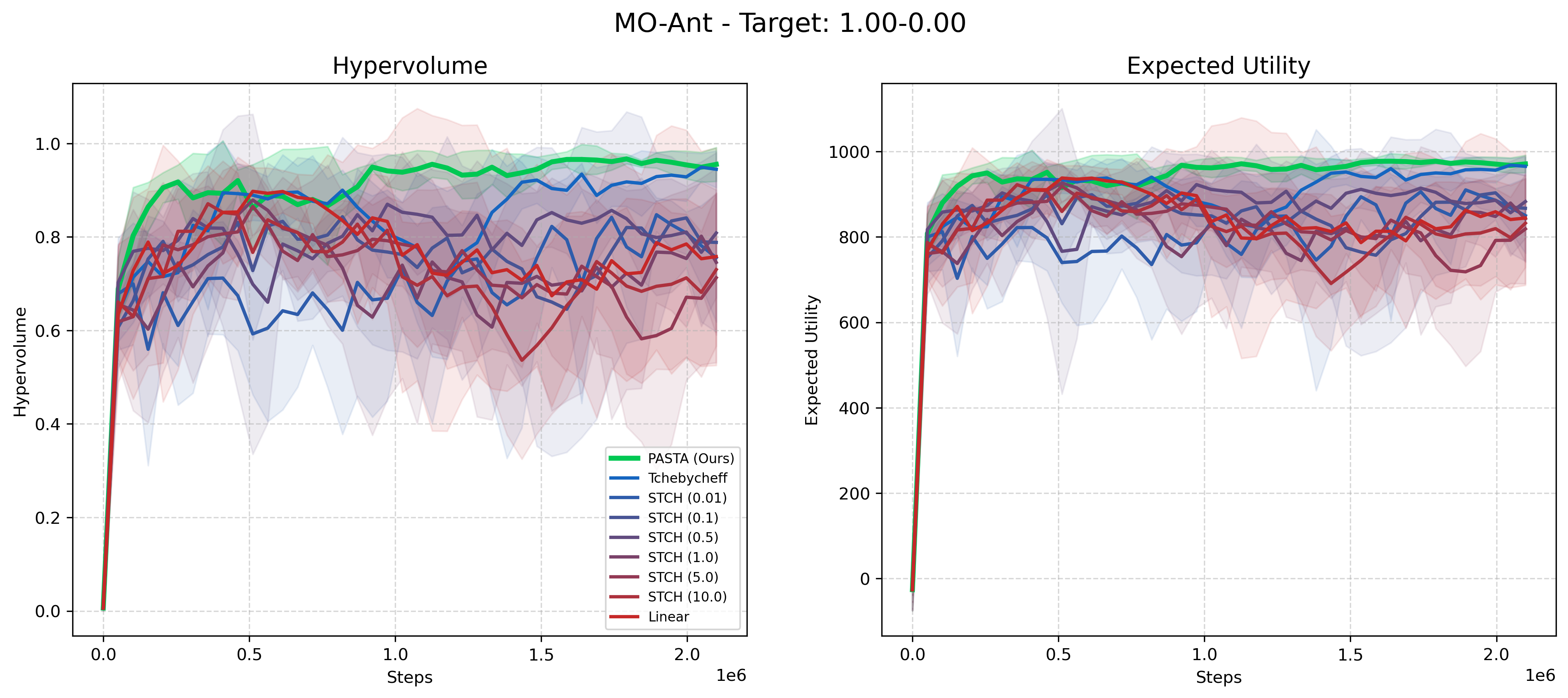}
        \caption{\small Ant}
        \label{fig:bottom-left}
    \end{subfigure}
    \hfill
    \begin{subfigure}[b]{0.485\textwidth}
        \centering
        \includegraphics[width=\linewidth]{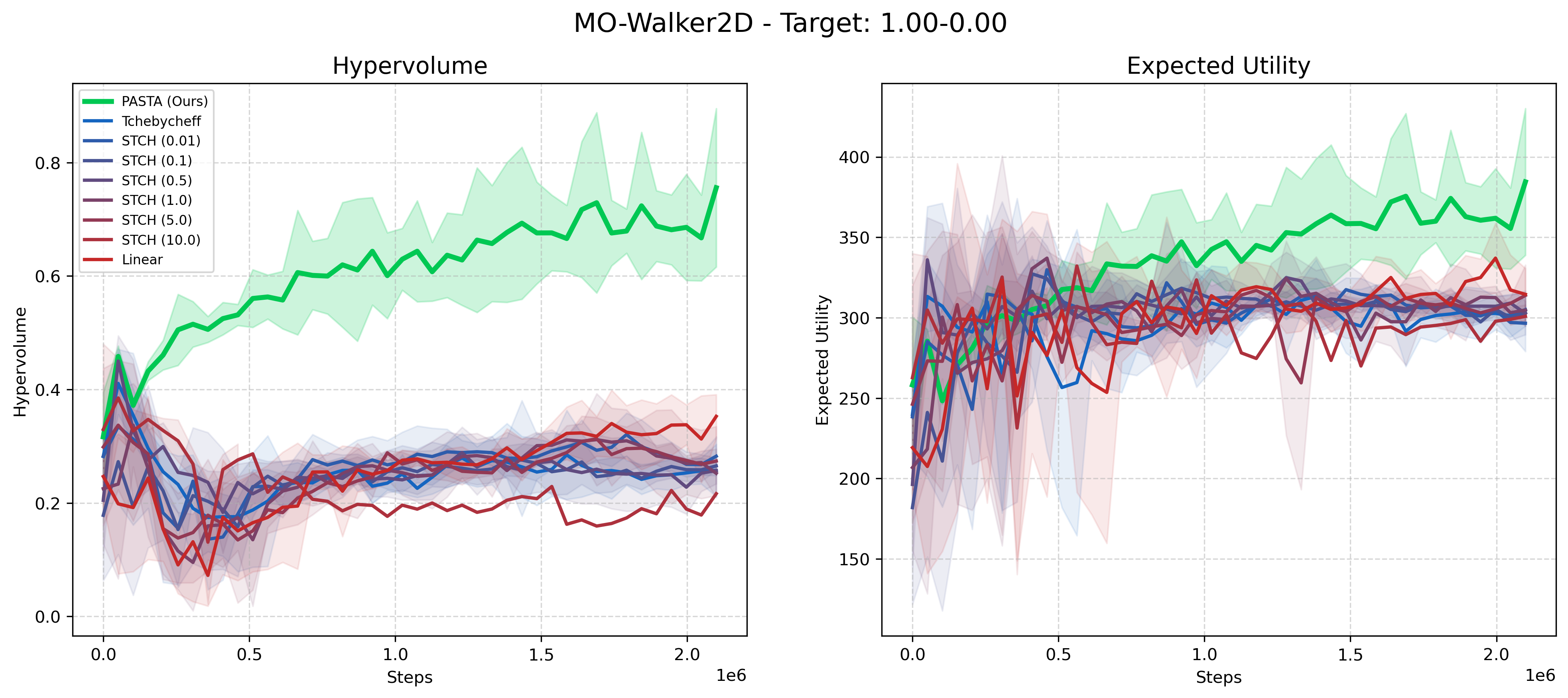}
        \caption{\small Walker 2D}
        \label{fig:bottom-right}
    \end{subfigure}
    
    \caption{\small Training statistics evaluated on validation episodes in the Multi-objective MuJoCo environments. Each subplot shows the results, as follows: \textit{(Left)} Evolution of the Hypervolume indicator, serving as a proxy for the quality and diversity of the approximated Pareto front. \textit{(Right)} The Expected Utility metric, measuring the alignment of the learned policies with specific preference weights. The solid lines represent the mean performance averaged over five seeds, while the shaded regions denote the standard deviation.}
    \label{fig:mujoco_performance_metrics}
\end{figure*}

Fig. \ref{fig:mujoco_performance_metrics} illustrates the training dynamics across validation episodes. In each subplot, we track Hypervolume (left) and Expected Utility (right), with shaded bands denoting the standard deviation over five random seeds. Overall, we can appreciate that PASTA achieves a better performance, with the largest difference in the MO-Humanoid environment and the smallest in the MO-Ant environment.

\begin{figure*}[t] 
    \centering
    
    % --- First Row ---
    \begin{subfigure}[b]{0.485\textwidth}
        \centering
        \includegraphics[width=\linewidth]{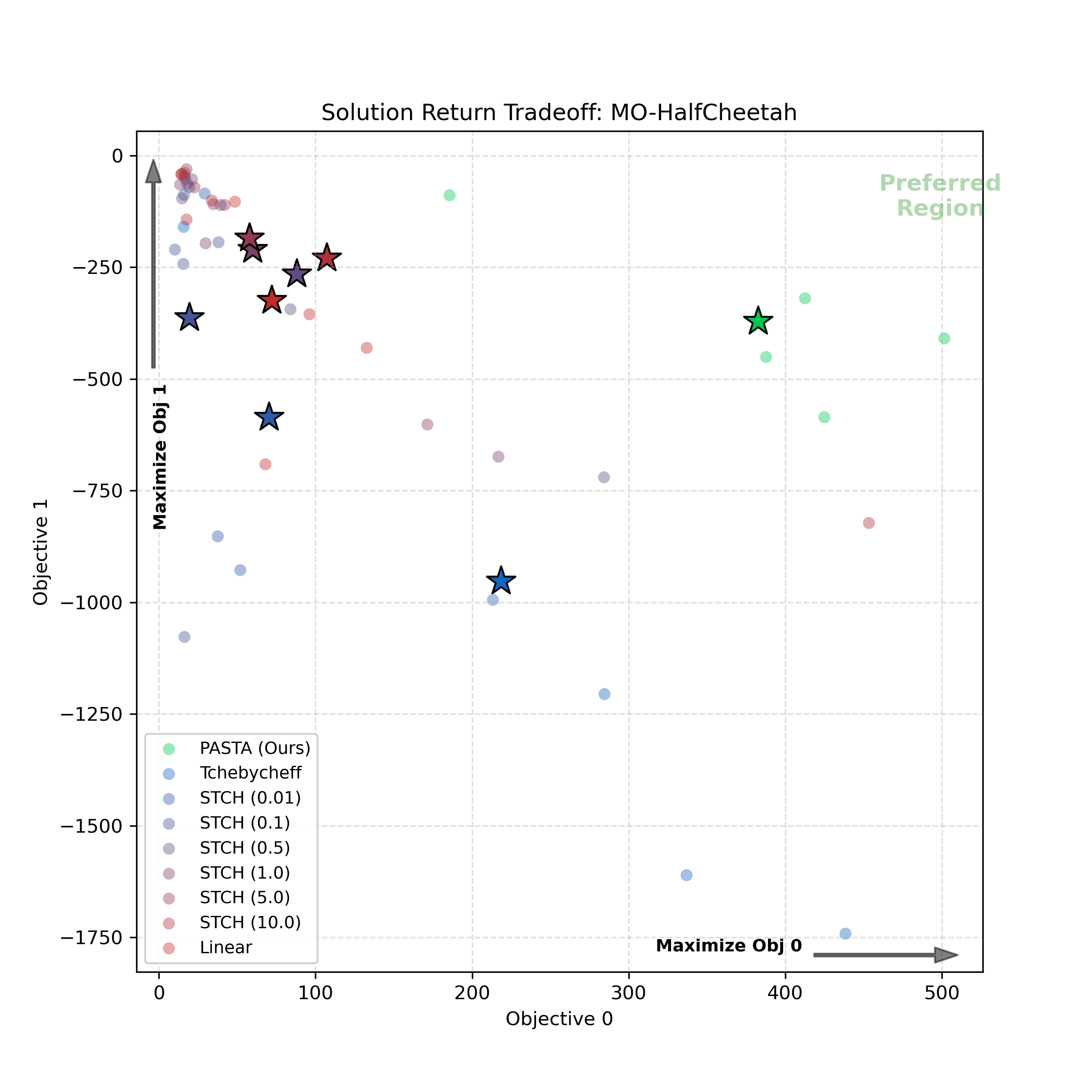} 
        \caption{\small Half Cheetah}
        \label{fig:top-left}
    \end{subfigure}
    \hfill 
    \begin{subfigure}[b]{0.485\textwidth}
        \centering
        \includegraphics[width=\linewidth]{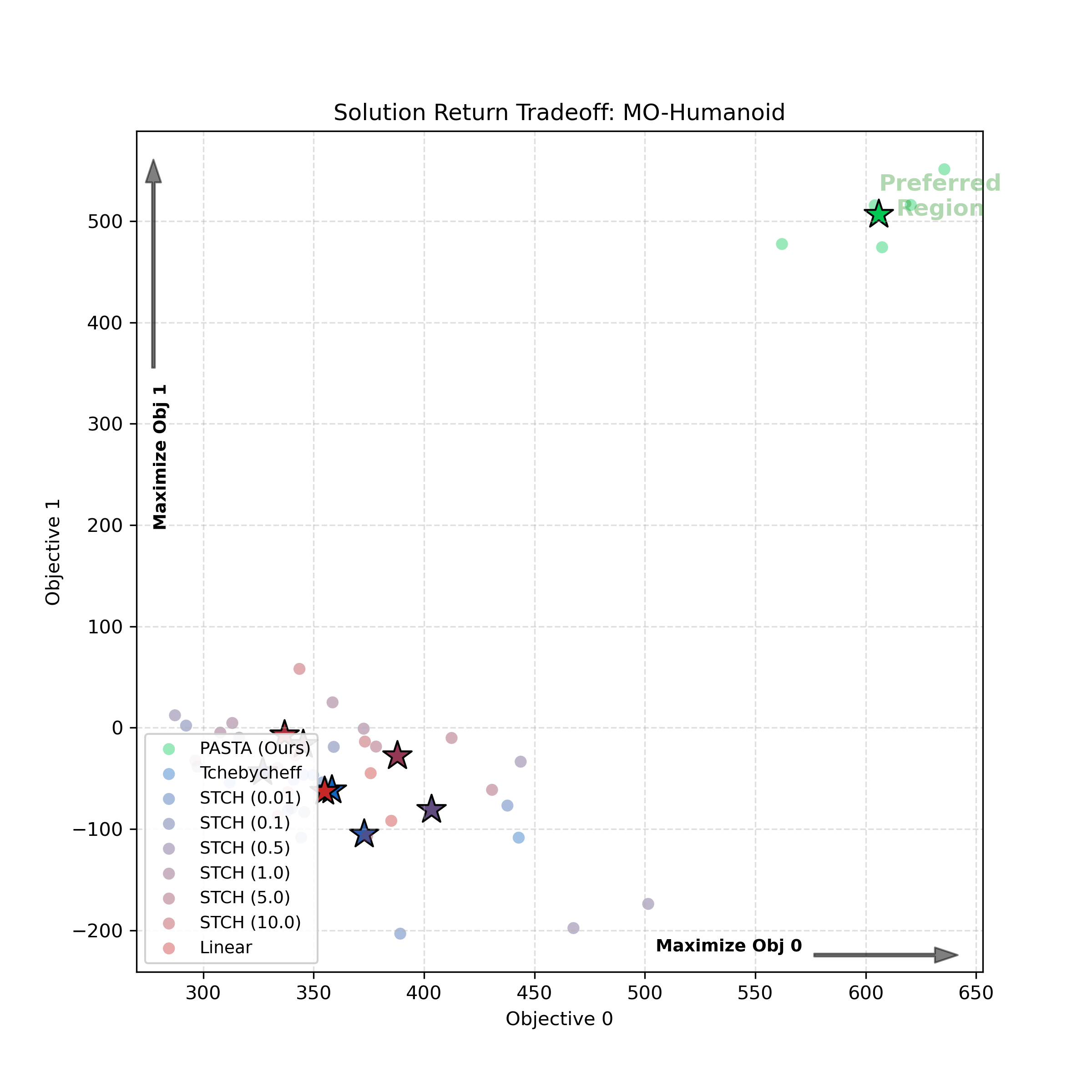}
        \caption{\small Humanoid}
        \label{fig:top-right}
    \end{subfigure}
    
    % --- Second Row ---
    \begin{subfigure}[b]{0.485\textwidth}
        \centering
        \includegraphics[width=\linewidth]{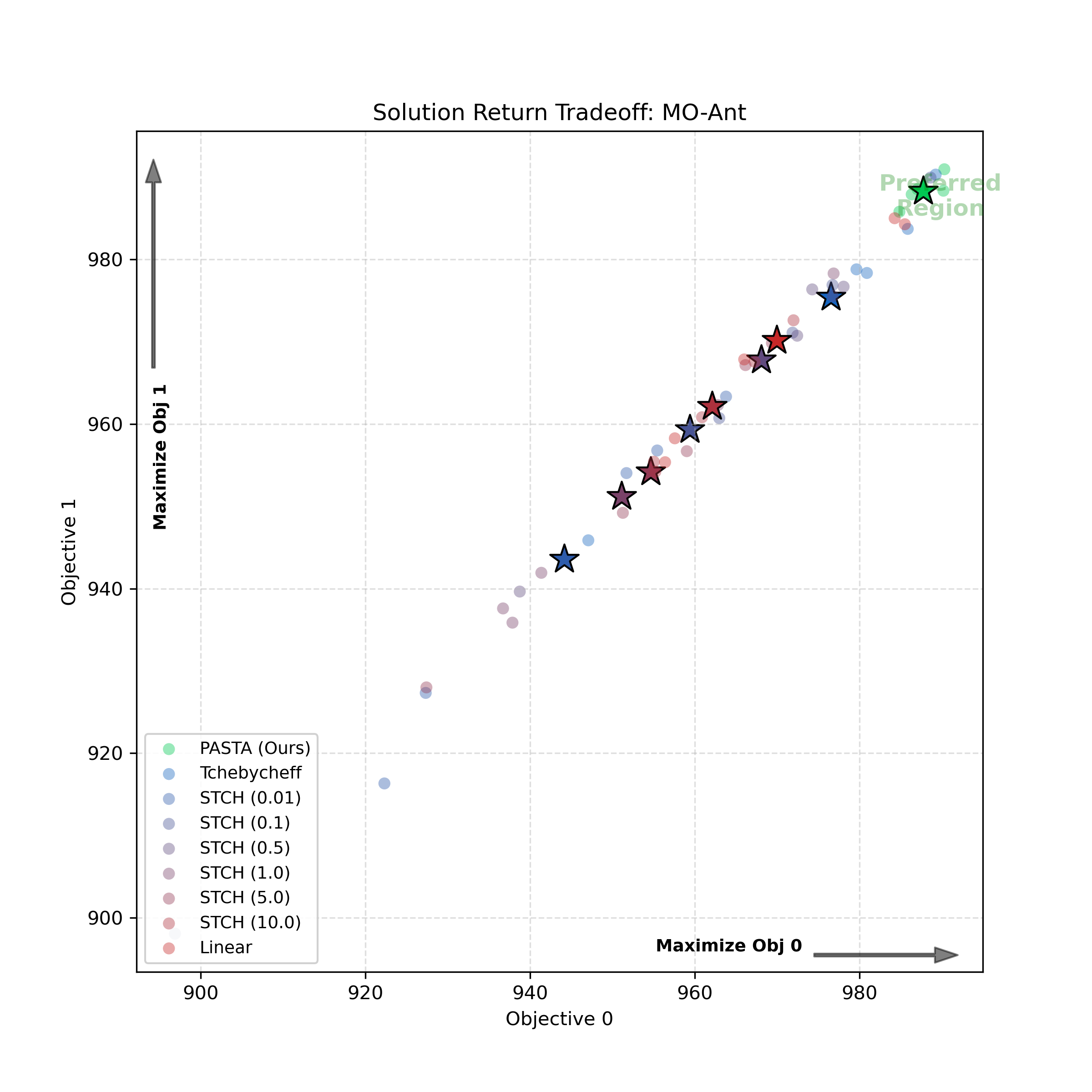}
        \caption{\small Ant}
        \label{fig:bottom-left}
    \end{subfigure}
    \hfill
    \begin{subfigure}[b]{0.485\textwidth}
        \centering
        \includegraphics[width=\linewidth]{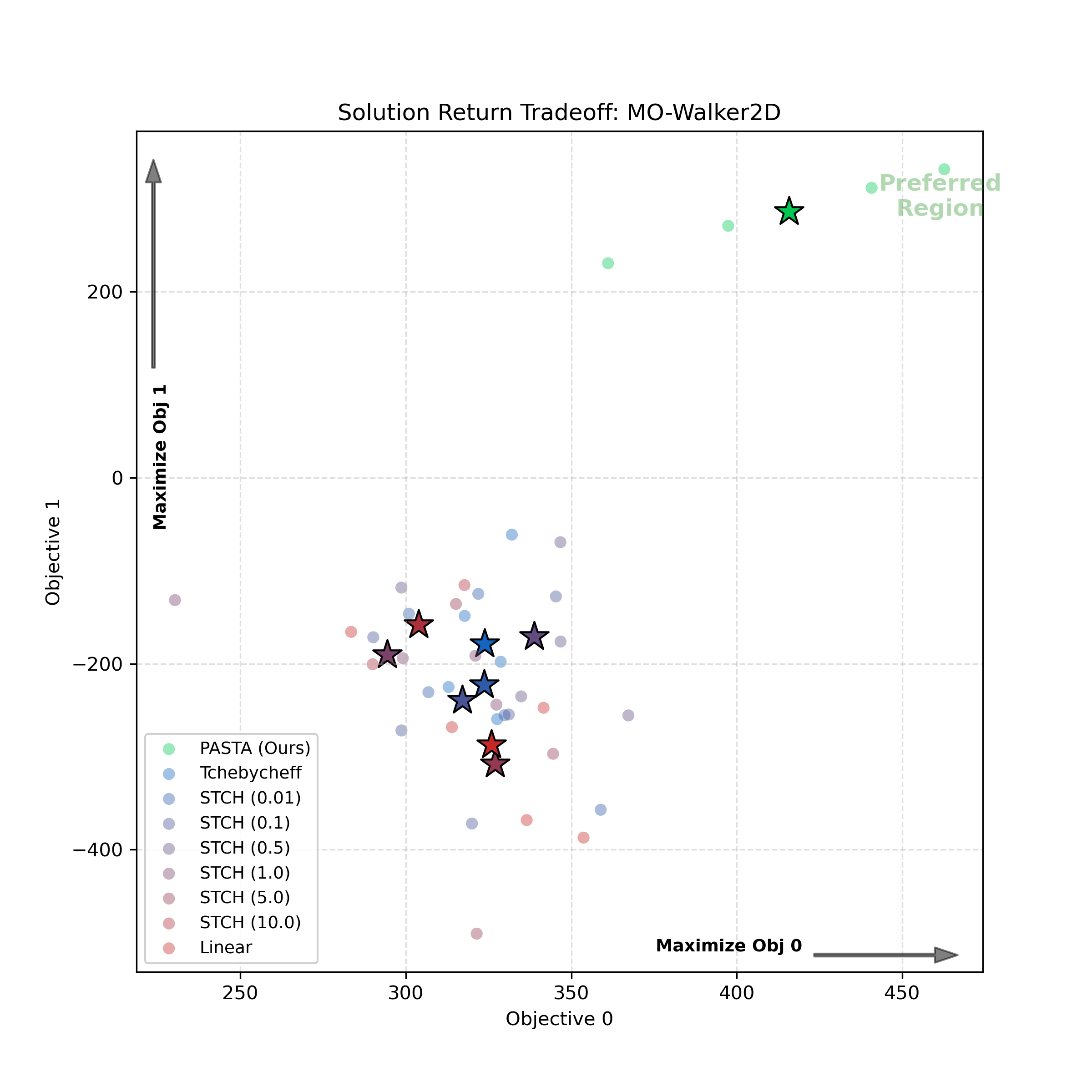}
        \caption{\small Walker 2D}
        \label{fig:bottom-right}
    \end{subfigure}
    
    \caption{\small Pareto front for each method on the Multi-objective MuJoCo environments.}
    \label{fig:mujoco_pareto_fronts}
\end{figure*}

Finally, Fig. \ref{fig:mujoco_pareto_fronts} visualizes the approximate Pareto fronts obtained by each method.
This comparison highlights the superior solution quality of PASTA, which consistently converges to a better-distributed and more expansive front than the baselines.

\subsection{Extended Results on Quadrotor Flights among Random Agents} \label{sec:appendix-quadrotor-demo}

\subsubsection{Environment Details}

We proceed to describe the multi-objective Frogger and Formation Crazyflie environments used for cross-platform evaluation of PASTA.
In both environments, opponents function as dynamic obstacles governed by a stochastic 1D patrolling policy. Each opponent is constrained to move along a fixed horizontal axis (e.g., $y=0$ for the Formation task, or $y \in \{-0.3, 0.3\}$ for the Frogger task) with a constant speed magnitude ($|\dot{x}| \in \{0.03, 0.04\}$ m/s). The motion follows a ``bounce" logic: velocity is inverted deterministically upon reaching the arena boundaries ($x_{\text{lim}} \approx \pm 0.95$). Additionally, to simulate unpredictable disturbances, there is a $5\%$ probability at every timestep that the opponent will spontaneously reverse its direction. This combination of boundary-constrained oscillation and stochastic direction switching requires the agents to learn robust collision avoidance policies that account for sudden adversarial movements.

\textbf{Frogger (Two-Line) Environment.}
This environment simulates a crossing task where a single Crazyflie must cross two lanes of dynamic traffic to reach a goal zone without colliding with patrolling opponents.

\noindent \underline{Objectives.}
The reward vector is: $\mathbf{r}_t = [r_{\text{goal}}, r_{\text{bounds}}, r_{\text{avoid}}]^T$.

\begin{itemize}
\item{Navigation ($r_{\text{goal}}$)}
Navigation is motivated by the change in distance to the goal $\mathbf{g}$:
\begin{equation}
    r_{\text{goal}} = \text{clip}\left( \|\mathbf{p}_{t-1} - \mathbf{g}\| - \|\mathbf{p}_{t} - \mathbf{g}\| , -1, 1 \right)
\end{equation}
A terminal success bonus of $+10$ is granted for reaching the goal. A penalty of $-15$ is applied for collisions or boundary violations.

\item{Boundary Adherence ($r_{\text{bounds}}$)}
To keep the agent within the viable flight area, a safety reward is calculated based on the distance to the nearest wall, $d_{\text{wall}}$:
\begin{equation}
    r_{\text{bounds}} = 0.1 \cdot \text{clip}\left( \frac{d_{\text{wall}}}{0.2}, 0, 1 \right)
\end{equation}
A penalty of $-25$ is applied if the agent violates the boundary.

\item{Traffic Avoidance ($r_{\text{avoid}}$)}
The agent must maintain a safe distance from all $M$ dynamic opponents. Let $d_{\text{opp}} = \min_{j \in M} \|\mathbf{p}_t - \mathbf{o}_{j,t}\|$. The avoidance reward is:
\begin{equation}
    r_{\text{avoid}} = 0.1 \cdot \text{clip}\left( \frac{d_{\text{opp}}}{0.3}, 0, 1 \right)
\end{equation}
A collision ($d_{\text{opp}} < 0.1$) results in a penalty of $-25$.
\end{itemize}

\textbf{Multi-Agent Formation Environment.}
In this environment, a team of $N=3$ Crazyflies must navigate from a starting zone to a target zone while maintaining an equilateral triangle formation, avoiding a dynamic obstacle, and staying within the flight arena bounds.

\noindent \underline{Objectives.}
The reward vector is: $\mathbf{r}_t = [r_{\text{goal}}, r_{\text{bounds}}, r_{\text{avoid}}, r_{\text{form}}]^T$.

\begin{itemize}
\item{Goal Reaching ($r_{\text{goal}}$)}
This objective guides the center of mass of the swarm, $\mathbf{p}_{\text{cm}}$, toward the goal position $\mathbf{g}$. It utilizes potential-based reward shaping and an energy penalty:
\begin{equation}
    r_{\text{goal}} = 5 \cdot (D_{t-1} - D_t) - 0.1 \cdot \bar{E}_t
\end{equation}
where $D_t = \|\mathbf{p}_{\text{cm}, t} - \mathbf{g}\|$ is the Euclidean distance to the goal, and $\bar{E}_t = \frac{1}{N} \sum_i \|\mathbf{a}_{i,t}\|$ is the average control effort. A sparse bonus of $+10$ is awarded upon convergence ($D_t < 0.1$), and a penalty of $-5$ is applied if a collision occurs.

\item{Boundary Safety ($r_{\text{bounds}}$)}
Agents are incentivized to stay away from the arena edges defined by $x_{\text{lim}}$ and $y_{\text{lim}}$. The reward is the aggregated safety of the most vulnerable agent:
\begin{equation}
    r_{\text{bounds}} = 0.1 \cdot \min_{i \in N} \left[ \text{clip}\left( \frac{\min(d_{x,i}, d_{y,i})}{0.2}, 0, 1 \right) \right]
\end{equation}
where $d_{x,i} = x_{\text{lim}} - |x_i|$ and $d_{y,i} = y_{\text{lim}} - |y_i|$. This creates a safety buffer of $0.2$ meters.

\item{Obstacle Avoidance ($r_{\text{avoid}}$)}
Agents must avoid a dynamic opponent moving with position $\mathbf{o}_t$. The reward scales with the distance of the closest agent to the obstacle:
\begin{equation}
    r_{\text{avoid}} = 0.2 \cdot \min_{i \in N} \left[ \text{clip}\left( \frac{\|\mathbf{p}_{i,t} - \mathbf{o}_t\|}{0.4}, 0, 1 \right) \right]
\end{equation}
If any agent collides with the obstacle ($\|\mathbf{p}_{i,t} - \mathbf{o}_t\| < 0.1$), a penalty of $-10$ is applied.

\item{Formation Integrity ($r_{\text{form}}$)}
The agents strive to maintain an equilateral triangle with side length $L_{\text{target}} = 0.45$m. Let $\mathcal{E}_t = \max_{j > i} |\|\mathbf{p}_i - \mathbf{p}_j\| - L_{\text{target}}|$ be the maximum deviation from the target link length. The reward combines an exponential precision bonus with a linear ``tether" penalty:
\begin{equation}
    r_{\text{form}} = \exp(-5 \cdot \mathcal{E}_t) - \text{clip}(\mathcal{E}_t, 0, 1)
\end{equation}
\end{itemize}

\subsubsection{System Specifications}

The flying task is validated using agile Crazyflie 2.1 nano-drones \cite{crazyswarm} within a flight arena equipped with a 12-camera Vicon motion capture system, which broadcasts ground truth poses for all agents at a frequency of 100~Hz. The task is complicated by the presence of \textit{guard drones} oscillating along the $x$-axis between $-2$~m and $2$~m. These guard drones follow a stochastic motion profile with a 5\% probability of changing its direction at any time step, forcing the agent(s) to reactively balance mission and safety constraints.

\subsubsection{Results}

% PACKAGES REQUIRED: \usepackage{booktabs}, \usepackage{xcolor}
\begin{table*}[h]
\centering
\fontsize{6pt}{7pt}\selectfont
\caption{Results (Baselines): CrazyflieFroggerEnv-v0. Metrics computed over 5 seeds.} \label{tab:crazy_frogger_results}
\begin{tabular}{lcccccc}
\toprule
Method & Hypervolume $\uparrow$ & E. Utility $\uparrow$ & Obj 0 $\uparrow$ & Obj 1 $\uparrow$ & Obj 2 $\uparrow$ \\
\midrule
\multicolumn{6}{c}{\textbf{Preferences: 0.40-0.05-0.55}} \\
\midrule
PASTA (Ours) & \textbf{0.796 $\pm$ 0.196} & \textbf{43.628 $\pm$ 5.542} & \textbf{9.068 $\pm$ 2.045} & \textbf{67.608 $\pm$ 6.271} & \textbf{66.582 $\pm$ 8.061} \\
Tchebycheff & 0.169 $\pm$ 0.156 & 16.741 $\pm$ 12.120 & -3.283 $\pm$ 4.559 & 35.154 $\pm$ 15.987 & 29.630 $\pm$ 17.462 \\
STCH (0.01) & 0.218 $\pm$ 0.227 & 18.697 $\pm$ 15.574 & -2.879 $\pm$ 6.026 & 38.986 $\pm$ 16.791 & 32.544 $\pm$ 22.498 \\
STCH (0.1) & 0.134 $\pm$ 0.179 & 13.864 $\pm$ 12.216 & \textcolor{gray}{-5.457 $\pm$ 5.099} & 34.743 $\pm$ 13.845 & 26.017 $\pm$ 17.579 \\
STCH (0.5) & \textcolor{gray}{0.129 $\pm$ 0.128} & \textcolor{gray}{13.225 $\pm$ 11.531} & -4.167 $\pm$ 3.472 & \textcolor{gray}{34.715 $\pm$ 9.893} & \textcolor{gray}{23.921 $\pm$ 17.603} \\
STCH (1.0) & 0.327 $\pm$ 0.385 & 24.627 $\pm$ 14.406 & 0.534 $\pm$ 6.764 & 41.854 $\pm$ 21.009 & \underline{40.583 $\pm$ 19.701} \\
STCH (5.0) & 0.238 $\pm$ 0.187 & 21.118 $\pm$ 11.103 & -0.663 $\pm$ 5.275 & 42.389 $\pm$ 10.284 & 35.025 $\pm$ 15.550 \\
STCH (10.0) & 0.216 $\pm$ 0.122 & 20.048 $\pm$ 9.415 & -0.519 $\pm$ 3.954 & 41.601 $\pm$ 9.024 & 33.046 $\pm$ 13.563 \\
Linear & \underline{0.360 $\pm$ 0.336} & \underline{24.867 $\pm$ 16.658} & \underline{1.128 $\pm$ 6.934} & \underline{47.537 $\pm$ 16.519} & 40.072 $\pm$ 23.771 \\
\bottomrule
\end{tabular}
\end{table*}

% PACKAGES REQUIRED: \usepackage{booktabs}, \usepackage{xcolor}
\begin{table*}[h]
\centering
\fontsize{6pt}{7pt}\selectfont
\caption{Results (Baselines): CrazyflieFormationEnv-v0. Metrics computed over 5 seeds.}  \label{tab:crazy_formation_results}
\begin{tabular}{lccccccc}
\toprule
Method & Hypervolume $\uparrow$ & E. Utility $\uparrow$ & Obj 0 $\uparrow$ & Obj 1 $\uparrow$ & Obj 2 $\uparrow$ & Obj 3 $\uparrow$ \\
\midrule
\multicolumn{7}{c}{\textbf{Preferences: 0.45-0.05-0.25-0.25}} \\
\midrule
PASTA (Ours) & \textbf{0.239 $\pm$ 0.144} & \textbf{146.507 $\pm$ 66.139} & \textcolor{gray}{-2.718 $\pm$ 5.913} & \textcolor{gray}{79.614 $\pm$ 16.634} & \textbf{142.665 $\pm$ 43.678} & \textbf{432.331 $\pm$ 228.230} \\
Tchebycheff & 0.168 $\pm$ 0.045 & \textcolor{gray}{53.297 $\pm$ 5.961} & 9.703 $\pm$ 1.363 & 93.011 $\pm$ 9.890 & 93.454 $\pm$ 10.776 & \textcolor{gray}{83.665 $\pm$ 8.909} \\
STCH (0.01) & \underline{0.209 $\pm$ 0.093} & 57.348 $\pm$ 10.506 & \textbf{10.476 $\pm$ 2.483} & 100.226 $\pm$ 17.715 & 101.238 $\pm$ 19.740 & 89.253 $\pm$ 14.522 \\
STCH (0.1) & 0.192 $\pm$ 0.051 & 58.255 $\pm$ 6.724 & 9.843 $\pm$ 1.909 & 97.997 $\pm$ 9.916 & 98.916 $\pm$ 10.819 & 96.785 $\pm$ 11.061 \\
STCH (0.5) & 0.193 $\pm$ 0.048 & 58.659 $\pm$ 6.200 & 9.946 $\pm$ 1.546 & 98.175 $\pm$ 9.880 & 98.827 $\pm$ 10.579 & 98.271 $\pm$ 9.629 \\
STCH (1.0) & 0.182 $\pm$ 0.053 & 56.996 $\pm$ 6.799 & 9.215 $\pm$ 1.372 & 96.028 $\pm$ 11.811 & 96.884 $\pm$ 12.862 & 95.306 $\pm$ 10.136 \\
STCH (5.0) & 0.205 $\pm$ 0.037 & \underline{60.721 $\pm$ 4.190} & \underline{10.295 $\pm$ 1.619} & \underline{100.241 $\pm$ 7.105} & 101.850 $\pm$ 6.733 & \underline{102.454 $\pm$ 6.496} \\
STCH (10.0) & \textcolor{gray}{0.159 $\pm$ 0.029} & 54.419 $\pm$ 4.058 & 8.882 $\pm$ 1.440 & 91.304 $\pm$ 6.487 & \textcolor{gray}{91.067 $\pm$ 7.060} & 92.362 $\pm$ 5.429 \\
Linear & 0.206 $\pm$ 0.066 & 60.270 $\pm$ 8.425 & 9.873 $\pm$ 1.037 & \textbf{100.349 $\pm$ 12.667} & \underline{101.958 $\pm$ 15.711} & 101.279 $\pm$ 13.836 \\
\bottomrule
\end{tabular}
\end{table*}

\begin{figure*}[t] 
    \centering
    
    % --- First Row ---
    \begin{subfigure}[b]{0.485\textwidth}
        \centering
        \includegraphics[width=\linewidth]{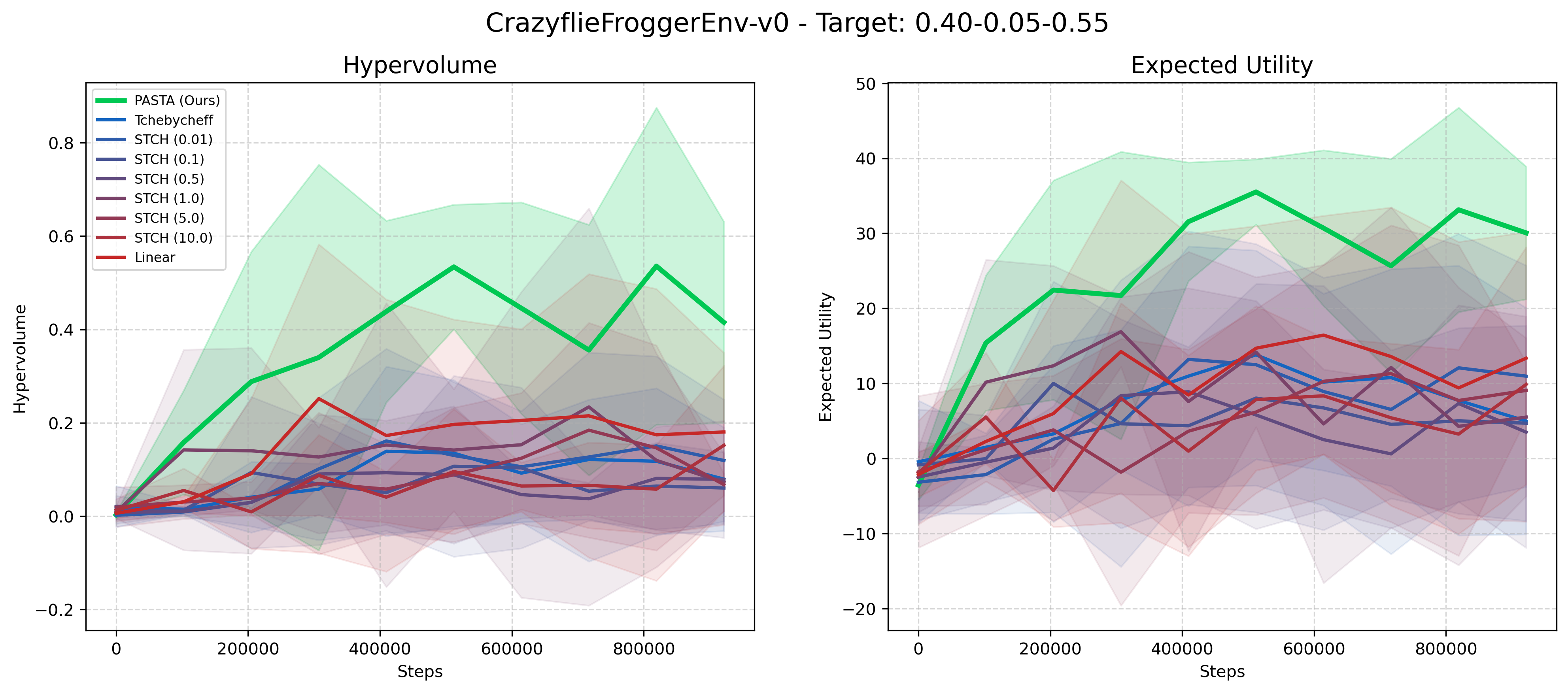} 
        \caption{\small Frogger}
        \label{fig:top-left}
    \end{subfigure}
    \hfill
    \begin{subfigure}[b]{0.485\textwidth}
        \centering
        \includegraphics[width=\linewidth]{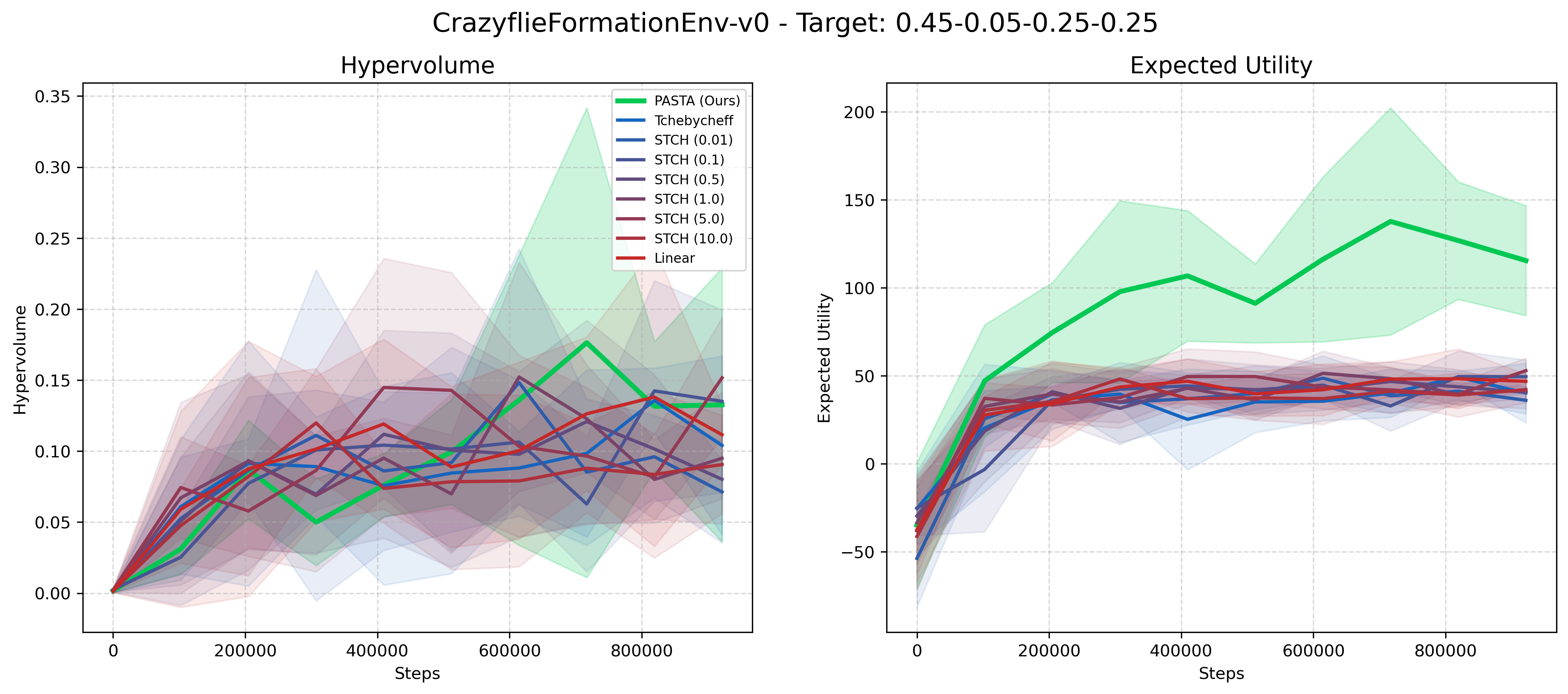}
        \caption{\small Formation}
        \label{fig:top-right}
    \end{subfigure}

    \caption{\small Training statistics evaluated on validation episodes in the Crazyflie environments. Each subplot shows the results, as follows: \textit{(Left)} Evolution of the Hypervolume indicator, serving as a proxy for the quality and diversity of the approximated Pareto front. \textit{(Right)} The Expected Utility metric, measuring the alignment of the learned policies with specific preference weights. The solid lines represent the mean performance averaged over five seeds, while the shaded regions denote the standard deviation.}
    \label{fig:crazy_performance_metrics}
\end{figure*}

Tables \ref{tab:crazy_frogger_results} and \ref{tab:crazy_formation_results} present the results for the Frogger and Formation Crazyflie environments, respectively. Fig. \ref{fig:crazy_performance_metrics} illustrates the training dynamics across validation episodes. In each subplot, we track Hypervolume (left) and Expected Utility (right), with shaded bands denoting the standard deviation over five random seeds. In general, PASTA achieves the best performance across tasks with no clear second-best.

\begin{figure*}[t] 
    \centering
    
    % --- First Row ---
    \begin{subfigure}[b]{0.485\textwidth}
        \centering
        \includegraphics[width=0.75\linewidth]{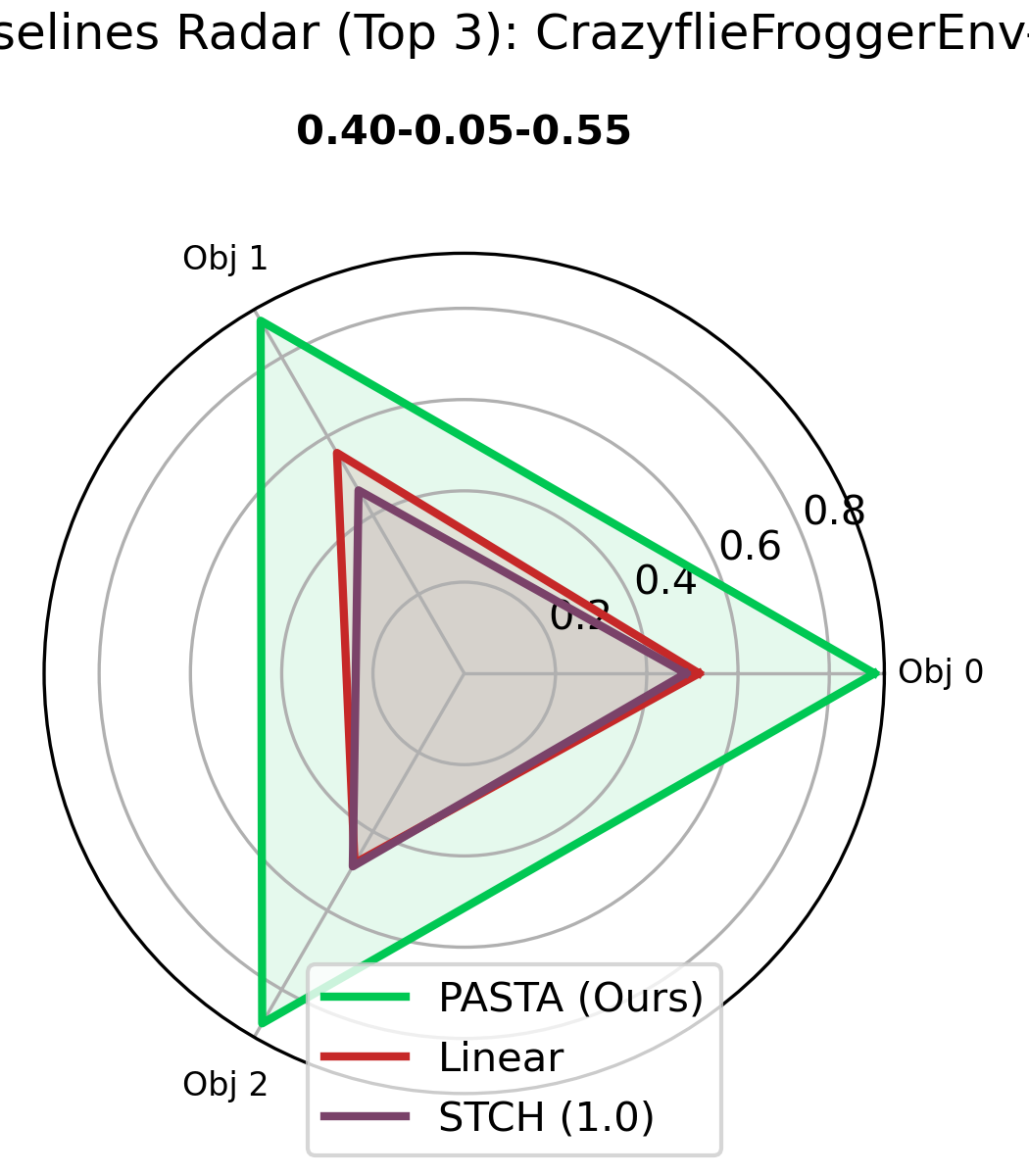} 
        \caption{\small Frogger}
        \label{fig:top-left}
    \end{subfigure}
    \hfill 
    \begin{subfigure}[b]{0.485\textwidth}
        \centering
        \includegraphics[width=0.75\linewidth]{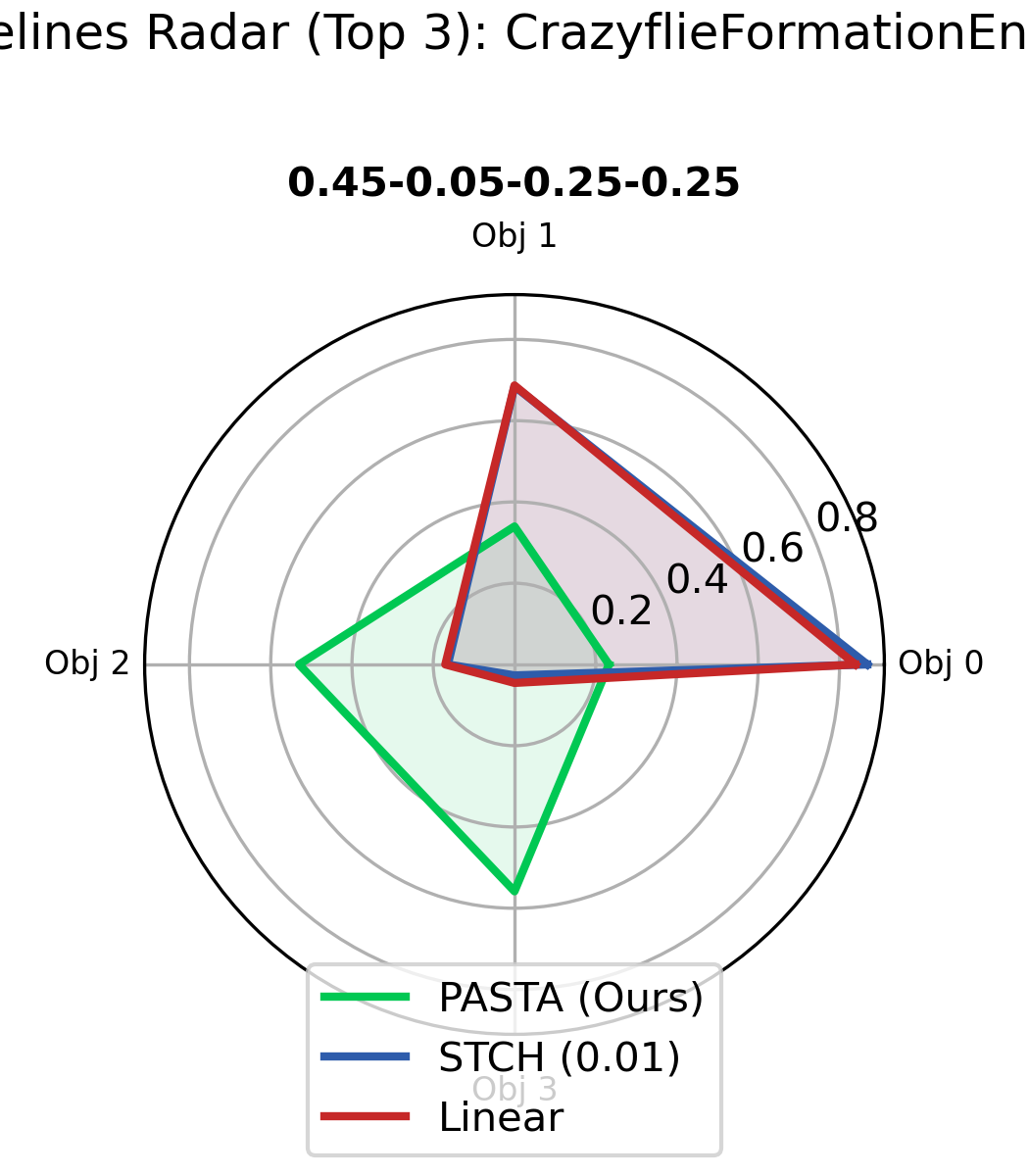}
        \caption{\small Formation}
        \label{fig:top-right}
    \end{subfigure}

    \caption{\small Normalized performance (utility) in Frogger \textit{(left)} and Formation \textit{(right)} Crazyflie environments. To reduce visual clutter, we show PASTA and the two baselines with better performance (in terms of hypervolume).}
    \label{fig:crazy_radar}
\end{figure*}

Fig. \ref{fig:crazy_radar} presents a radar plot comparing PASTA against the top two baselines in each environment. This comparison highlights the superior solution quality of PASTA in the Frogger environment. In the Formation environment, it demonstrates that PASTA correctly prioritized the higher-preference objectives (3 and 4) over objective 1, though this came at the expense of the goal-reaching objective. 

Upon closer inspection of the agents in the Formation environment, we observed that PASTA approached the goal while maintaining the formation's shape, but narrowly missed the exact target. Conversely, the baselines succeeded in reaching the goal but failed to enforce the shape constraint; this allowed them to secure the terminal goal-reaching bonus, artificially inflating the return for objective 1. This is also evidenced in Fig. \ref{fig:crazy_radar}, where their score in the formation's shape objective is minimal.

Finally, in the supplementary \textbf{Video 2} we showcase the real world experiments in the Frogger environment in two trials. The task involves reaching a set of goals in the corners of the environment while converging to a goal in the origin in between every other corner goal. This is repeated continuously. In our experiments, the trials ended whenever the quadrotors' battery was too low to keep flying. From this real world demonstrations, we saw our agent achieve $52$ out of $57$ objectives without crashing, meaning a task (goal-reaching among random dynamic obstacles) success rate of \textbf{91.2\%}. Note that we flew the ego and guard drones at different heights to protect the hardware and to enable continuous execution whenever a crash happened (we can count the number of crashes from the 2D view, also shown in the video), letting us collect a larger sample to properly estimate the policy's performance.

\end{document}